\documentclass[10pt,twocolumn,letterpaper]{article}
\usepackage{times}
\usepackage{epsfig}
\usepackage{graphicx}
\usepackage{amsmath}
\usepackage{amssymb}
\usepackage{multirow}
\usepackage{dsfont}
\usepackage{soul}
\usepackage{wrapfig}
\usepackage[pagebackref=true,breaklinks=true,colorlinks,bookmarks=false]{hyperref}

\newcommand{\todo}[1]{}
\renewcommand{\todo}[1]{{\color{red} [TODO: {#1}]}}
\newcommand{\tocheck}[1]{}      
\renewcommand{\tocheck}[1]{{\color{blue} [TOCHECK: {#1}]}}
\newcommand{\corrected}[1]{}
\renewcommand{\corrected}[1]{{\color{green} [CORR: {#1}]}}
\newcommand{\question}[1]{}
\renewcommand{\question}[1]{{\color{blue} [QUES: {#1}]}}
\newcommand{\bfff}[1]{}
\renewcommand{\bfff}[1]{{\color{blue} {#1}}}
\def\etal{\emph{et~al}.}

\begin{document}
\date{}
%%%%%%%%% TITLE
%\title{Anatomically structured graph CNNs for pose estimation of closely-interacting hands}

%\title{Parallel mesh reconstruction streams for pose estimation from a single RGB image}
\title{Parallel mesh reconstruction streams for pose estimation of interacting hands}

\author{Uri Wollner\\
Artificial Intelligence, GE Research\\
%Tel-Aviv, Israel\\
{\tt\small uri.wollner@ge.com}
\and
Guy Ben-Yosef\\
Artificial Intelligence, GE Research\\
%Tel-Aviv, Israel\\
{\tt\small guy.ben-yosef@ge.com}
}

\maketitle

%%%%%%%%% ABSTRACT
\begin{abstract}

% ICCV:
We present a new multi-stream 3D mesh reconstruction network (MSMR-Net) for hand pose estimation from a single RGB image. Our model consists of an image encoder followed by a mesh-convolution decoder composed of connected graph convolution layers. In contrast to previous models that form a single mesh decoding path, 
our decoder network incorporates multiple cross-resolution trajectories that are executed in parallel. Thus, global and local information are shared to form rich decoding representations at minor additional parameter cost compared to the single trajectory network. 
We demonstrate the effectiveness of our method in hand-hand and hand-object interaction scenarios at various levels of interaction. To evaluate the former scenario, we propose a method to generate RGB images of closely interacting hands. Moreoever, we suggest a metric to quantify the degree of interaction and show that close hand interactions are particularly challenging. Experimental results show that the MSMR-Net outperforms existing algorithms on the hand-object FreiHAND dataset as well as on our own hand-hand dataset. 
%Moreover, since MSMR-Net directly regresses image-to-3D mesh vertices, it can be easily extended to human pose estimation. We present SOTA-comparable results on the 3DPW dataset.

% \vspace{1cm}
% Keywords: Hand pose, GCN, RGB image, synthetic data, hand-hand interaction 

\end{abstract}

%%%%%%%%%%%%%%%%%%%%%%%%%%%%%%%%%%%
%%%%%%%%%%%%%%%%%%%%%%%%%%%%%%%%%%%
% INTRODUCTION
%%%%%%%%%%%%%%%%%%%%%%%%%%%%%%%%%%%
%%%%%%%%%%%%%%%%%%%%%%%%%%%%%%%%%%%
\section{Introduction}
\label{sec:introduction}

\begin{figure*}
	\begin{center}
		\tabcolsep 0.03cm
		\begin{tabular}{cccccccc}			
			%{  %            		left bottom right top
			\includegraphics[trim = 10mm 0mm 10mm 0mm, clip, height=1.9cm]{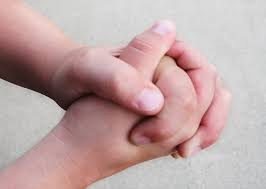} &
			\includegraphics[trim = 0mm 0mm 0mm 0mm, clip,  height=1.9cm]{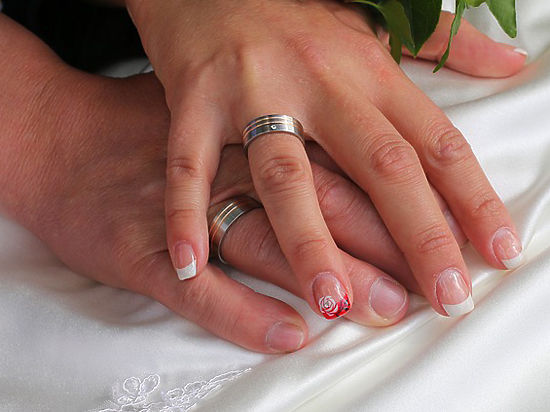} &
			\includegraphics[trim = 0mm 0mm 5mm 0mm, clip, height=1.9cm]{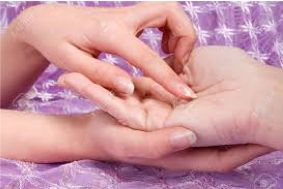} &
			\includegraphics[trim = 10mm 0mm 5mm 0mm, clip, height=1.9cm]{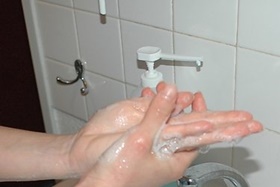} &
			\includegraphics[trim = 0mm 0mm 10mm 0mm, clip, height=1.9cm]{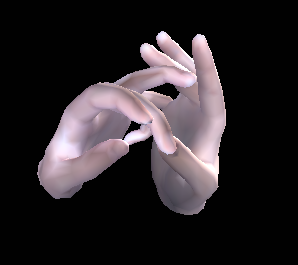} &
			\includegraphics[trim = 50mm 30mm 50mm 30mm, clip, height=1.9cm]{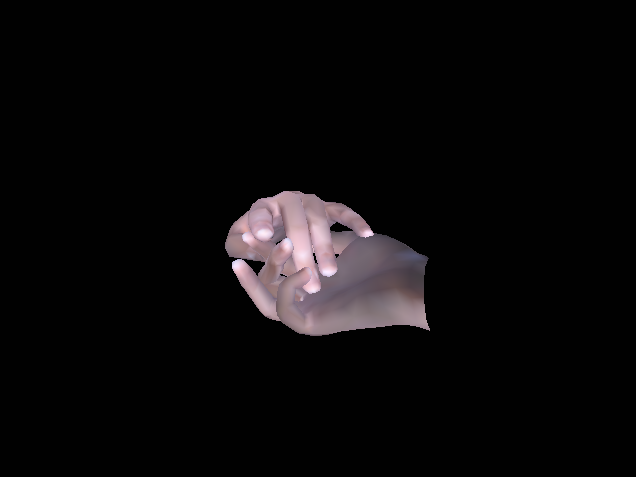} &
			\includegraphics[trim = 50mm 20mm 50mm 20mm, clip, height=1.9cm]{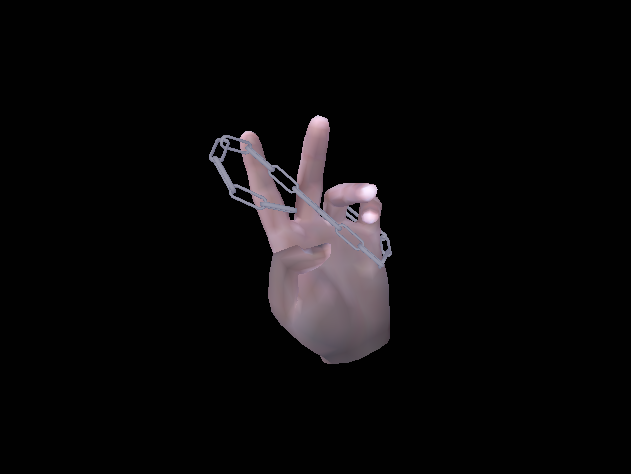} &
			\includegraphics[trim = 0mm 0mm 0mm 0mm, clip, height=1.9cm]{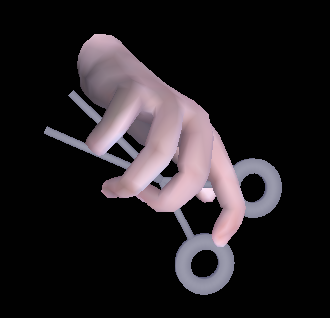}
			
% 			height=1.9cm]{figures/Hifiving_2.png} &
% 			\includegraphics[trim = 0mm 0mm 0mm 0mm, clip,
			
			\\
			(a) & (b) & (c) & (d) & (e) & (f) & (g) & (h) 
			%}
		\end{tabular}
	\end{center}
	\vspace{-0.2cm} 
	\caption{Closely interacting hands during (a) clasping, (b) stroking, (c) palm reading, and (d) washing. We emulate such scenarios using generated examples that include close hand-hand interactions (e-f), and close hand-object interactions (g-h). In such interactions, a hand's fingers cross another hand's fingers or pass through a perforated object. Thus, potentially challenging hand pose estimation algorithms.
	}
	\label{fig:intro}
\end{figure*}

% Describing the high-level motivation to study the problem, applications, use cases etc.: 
Human daily behavior include intensive interactions with surrounding objects and with other humans using hands. Capturing such interactions in natural RGB images and analyzing them is therefore important for a wide variety of applications such as monitoring human activity,   %~\cite{REF}, 
virtual/augmented reality,  %~\cite{JangEtAl2015, PiumsomboonEtAl2013}, 
and human-machine interactions %~\cite{SridharEtAl2015}, 
among others. In particular, analyzing close hand-hand or hand-object interaction scenarios (Fig.~\ref{fig:intro}) in terms of hand and object poses, will eventually allow better visual understanding of the interactions' type and tone.

% Particularly challenging scenarios for such applications include close hand-hand or hand-object interactions (Fig.~\ref{fig:intro}). Yet, analyzing these scenarios in term of hand and object poses will allow better visual understanding of the interaction type and tone, which are essential for fine-grained visual understanding of the  interaction types and tones.%(e.g., to assist experts such as technicians, surgeons, or pilots. 

% Describing the state-of-the-art in the field in high-level terms: 
Current hand pose estimation models from RGB images predict poses in either 2D, 3D, or both. While traditional approaches were focused on solely estimating coordinates of hand joints and fingertips (e.g., \cite{simon2017hand,ZimmermannBrox2017, Panteleris2018UsingAS,spurr2020weakly}), recent models tend to estimate a 3D dense hand mesh posed at the predicted configuration (e.g., \cite{boukhayma20193d, zhang2019,Hasson_2019_CVPR,ge20193d, dkulon2020cvpr,moon2020interhand26m,choi2020pose2mesh, lin2021end}). 
% Recently proposed hand pose estimation models specifically leverage the relative spatial positioning of the joints or mesh vertices in their algorithm.  These relations can be expressed via bio-mechanical constraints in the loss function during training of a neural network \cite{spurr2020weakly} or by employing Graph Neural Networks as a recent popular example~\cite{ge20193d, doosti2020hopenet,dkulon2020cvpr}. Still, most of these works tend to use local spatial relations, whereas long-range relations are yet under explored.
%
% In this work we aim to estimate the 3D location of hand joints and fingertips by means of leveraging the relative spatial positing of these joints as was previously performed on dense mesh vertices \cite{dkulon2020cvpr}. 
% Thus, considerably reducing the computations times of hand pose predictions while attaining state-of-the-art performance.
%
% Describing the problems that still exist and how we plan to overcome them: 
To develop and test hand pose estimation algorithms, multiple real~\cite{simon2017hand, dkulon2020cvpr, HernandoEtAl2018, Freihand2019, moon2020interhand26m} and synthetic ~\cite{MuellerEtAl2018, ZimmermannBrox2017, Hasson_2019_CVPR, Lin_2021_WACV} hand datasets were collected and generated in recent years.
% To estimate the accuracy and robustness of hand pose algorithms, numerous relatively large real \cite{SimonEtAl2017, dkulon2020cvpr, HernandoEtAl2018, JooEtAl2018, Freihand2019, moon2020interhand26m} and synthetic \cite{MuellerEtAl2018, ZimmermannBrox2017, Hasson_2019_CVPR} datasets were generated/collected in recent years. 
%
These datasets encompass various hand-object interactions \cite{Tzionas2016, dkulon2020cvpr, HampaliEtAl2019, Hasson_2019_CVPR, GRAB2020, Brahmbhatt_2020_ECCV, HernandoEtAl2018, Freihand2019}, single-hand gestures \cite{ZhangEtAl2016, ZimmermannBrox2017}, as well as hand-hand interactions \cite{Tzionas2016, moon2020interhand26m, Lin_2021_WACV}. 
% However, only few of them contain mesh ground truth annotations, which are suitable for developing mesh-based hand reconstruction models.

Despite the recent improvement in prediction accuracy of hand pose estimation models for hand-object scenarios, their prediction performance for intricate hand-hand and hand-object interactions remain unclear. 
% and the rapidly growing number of hand pose datasets, 
% these models fail to accurately predict hand poses from RGB images capturing intricate hand-hand and hand-object interactions. 
Furthermore, to the best of our knowledge, a metric to identify these intricate settings does not currently exist. To this end, we develop a novel mesh-based dataset generator for closely-interacting hand-hand and hand-object scenarios. We then investigate the performance of existing models on the generated data. In addition, we propose a visibility-ratio metric to identify and characterize hand interaction scenes. 
% CVPR version:
%Based on insights of our analysis, we develop a new hand pose estimation model consisting of a Graph Convolution neural Network that is structured according to anatomical hand parts. 
%Thus, promoting learning and propagation of long-range spatial relations. Experimental results on various external and internal datasets show that our model can better estimate joints of closely interacting hands examples than recently proposed models, while still providing high prediction accuracy for the non-interacting cases.

% New ICCV version:
Based on insights of our analysis, we develop a hand reconstruction model consisting of Graph Convolution neural Networks (GCNs). Our GCN decoder architecture is structured with multi-resolution mesh maps that are interconnected to create several mesh decoding trajectories. This structure allows high-resolution mesh layers to obtain global contextual information that exist at the low-resolution mesh layers. Experimental results on external and internal datasets show that our model outperforms recently proposed models on hand mesh reconstruction and pose estimation tasks. 
% In particular, we show that our model is useful for predicting pose of closely interacting hands.

%and while containing significantly less parameters than its competitors\tocheck{Is this true? If so, we will need to put a table with the number of parameters for all models including ours, and to show a significant difference.}

    %^^^^^^^^^^^^^^^^^^^^^^^^^^^^^^^^^^^^^^^^^^^^^^^^^^^^^^^^^^^^^^^^^^^^^^^^^^^^^^^^^^^^^^^^^^^^^^^^^^^^^^^^

% Describing a summary of key contributions and novelty: 
To conclude, our key contributions are as follows:
\begin{itemize}
    \item 
    We present a hand mesh pose estimation model that utilizes a strategy of multiple mesh decoding paths. 
    % The proposed method outperforms current state-of-the-art (e.g.,~\cite{lin2021end,dkulon2020cvpr}) on mesh reconstruction of single hand, hand-object, and hand-hand image examples.
    \item 
    We introduce a method to generate synthetic examples of close hand-hand and hand-object interactions. It extends previous %hand-object image generation 
    work (e.g.,~\cite{Hasson_2019_CVPR}) by allowing spatial configurations of two or more intertwined articulated objects. %It is suitable for developing mesh models.
    \item 
    We introduce a metric to assess the degree of hands' interactions in an image. We use it to form groups of hand-hand interaction scenarios and test the accuracy of hand pose estimation models on them. 
    %hand interacting hand-hand and hand-object images.
    % \item We release a synthetic dataset of \textbf{XX} hands closely-interacting with other hands or objects. An example result is shown in Fig. \textbf{XX}.
\end{itemize}

%%%%%%%%%%%%%%%%%%%%%%%%%%%%%%%%%%%%%%%%%%%%%%%%%%%%%%%%%%%%%%%%%%%%%%%%%%%
%%%%%%%%%%%%%%%%%%%%%%%%%%%%%%%%%%%%%%%%%%%%%%%%%%%%%%%%%%%%%%%%%%%%%%%%%%%
% RELATED WORK
%%%%%%%%%%%%%%%%%%%%%%%%%%%%%%%%%%%%%%%%%%%%%%%%%%%%%%%%%%%%%%%%%%%%%%%%%%%
%%%%%%%%%%%%%%%%%%%%%%%%%%%%%%%%%%%%%%%%%%%%%%%%%%%%%%%%%%%%%%%%%%%%%%%%%%%
\section{Related work}
\label{sec:related_work}

\textbf{Hand mesh reconstruction}. 
Many traditional 3D hand pose approaches have utilized neural networks to directly regress 3D hand keypoints' positions~\cite{ZimmermannBrox2017, iqbalEtAl2018, CaiEtAl2018, doosti2020hopenet, MuellerEtAl2018, spurr2018, YangYao2019,spurr2020weakly}. More recent approaches estimate a mesh of the 3D hand surface~\cite{dkulon2020cvpr,zhou2020,ge20193d,zhang2019,moon2020interhand26m,choi2020pose2mesh,lin2021end} by either (a) predicting parameters of a deformable 3D hand model, such as the MANO model~\cite{MANO2017}, or (b) estimating 3D coordinates of mesh vertices directly. In both cases, the 3D hand keypoints are obtained by multiplying the estimated mesh by a predefined regression matrix.

Lin~\etal~\cite{lin2021end} presented a multi-layer Transformer \cite{Vaswani2017} encoder with progressive dimensionality reduction to reconstruct a 3D mesh. They show that their model captures short- and long-range interactions among body joints and mesh vertices. This work has motivated us to incorporate a transformer block in our network as well to better capture these long-range relations. 

% based on several self-attention layers combined with powerful HRNet backbone \cite{XX} that to this date provided the SOTA results on FrieHand, the current central hand pose benchmark.

%
Our work is also inspired by recent papers suggesting the use of Graph Convolution Networks (GCNs)~\cite{kipf2017semi,xu2018how} for propagating contextual information along the graph structure. The purpose of GCNs in the context of hand pose estimation algorithms is to contribute to the learning of high-level relationships between hand keypoints or mesh vertices. Typically, the pose model begins with encoding an input image into a latent feature vector. Then, a GCN decodes the latent features into a 3D hand skeleton or mesh.

Ge~\etal~\cite{ge20193d} were the first to include GCNs in a hand pose estimation model. They proposed a model to reconstruct a hand mesh from a latent feature vector that is computed based on 2D hand joints heat-maps and feature-maps obtained from a stacked hourglass network~\cite{newell2016stacked}. The reconstruction model uses GCNs that represent locally connected mesh vertices at different refinement levels. The nodes of the GCNs are the mesh vertices and their edges are defined based on the mesh topology. The model starts with a low-resolution spatial map (a coarse mesh) that is upsampled gradually to a obtain a high-order one (a finer mesh).  
Kulon~\etal~\cite{dkulon2020cvpr} also suggested a GCN-based mesh reconstruction model. Yet, in contrast to~\cite{ge20193d} that used spectral graph convolution operations in their network, \cite{dkulon2020cvpr} used convolution operations applied to connected mesh nodes according to a local spiral patch operator. 
% The latent feature vector in~\cite{dkulon2020cvpr} is the ResNet output~\cite{HeResnet50}. They have shown experimental results indicating that the spiral structure is more efficient than the spectral structure.
Other models include Choi~\etal~\cite{choi2020pose2mesh} who proposed a GCN-based system to estimate 3D coordinates of hand mesh vertices from 2D hand pose. 
%Their model essentially disregards an image's contextual information. Nevertheless, their results outperform previously proposed models.
We note that \cite{dkulon2020cvpr, ge20193d, choi2020pose2mesh} models are structured with a single, sequential, mesh decoding path.  In Sec.~\ref{sec:model} we present a GCN-based architecture that uses spiral convolution as well, but incorporates multiple interconnected paths for mesh decoding.

%%%%%%%%%%%%%%%%%% Synthetic Data:
\textbf{Datasets of synthetic hand images}. 
Datasets and benchmarks with synthesized single hand images have been introduced as an alternative to collecting and annotating real-world data~\cite{ZimmermannBrox2017, MuellerEtAl2018}. Images in these datasets were typically generated by a computer graphics software that rendered a digitized hand model. To create a more realistic hand appearance, GANs were further employed on the rendered images% of the graphics software
~\cite{MuellerEtAl2018}. 

Further work on synthetic hand-object images was done by Hasson~\etal~\cite{Hasson_2019_CVPR} who introduced the ObMan (Object Manipulation) dataset. \cite{Hasson_2019_CVPR} adapted the MANO hand model~\cite{MANO2017} and combined it with the GraspIt software \cite{Graspit} to generate hand grasp poses interacting with 2.7K everyday object models. The GraspIt algorithm was used to move the hand model towards the object and increase the hand-object contact surface while performing a grasping motion. 
In its current implementation, the GraspIt algorithm is able to create hand-object interactions such that the object is static and the hand is dynamic, i.e. the hand joints can rotate. However, it cannot provide interaction examples where two articulated objects are dynamically moving; a requirement for creating hand-hand interactions. 
To alleviate this constraint, we present in Sec.~\ref{sec:synthetic_data} a new method to create interactions of two dynamically moving articulated objects. We then use this method to generate synthetic hand-hand dataset for training and evaluating hand pose estimation models.

%%%%%%%%%%%%%%%%%% Hands interactions:
\textbf{Hand-hand interactions}. 
Very few studies were dedicated for pose estimation of interacting hands, including proposed datasets and models~\cite{oikonomidis2012tracking,Tzionas2016,mueller2019real,moon2020interhand26m, Lin_2021_WACV}. 
~\cite{moon2020interhand26m} collected 2.6 million real-captured images of interacting hand poses, and proposed a model for hand pose estimation of a single or two interacting hands. Our current work supplements previously published datasets by creating a large synthetic dataset of closely interacting hands at various configurations. The generated dataset consists of non-intersecting hand meshes that are suitable for development of mesh reconstruction models. Moreover, our dataset extends current mesh-annotated datasets for hand-object interactions. The currently public hand-object interaction datasets contain scenes with objects having simple geometric shapes (e.g., cylindrical or rectangular). Our generator can also produce scenes of hands interacting with articulated objects, in which the hand and object are intertwined (e.g., Fig~\ref{fig:intro}g). 
% These are likely challenging for pose estimation models.  
In this work we consider them similar to examples of close hand-hand interactions.

%%%%%%%%%%%%%%%%%% Resolution-based representation:
\textbf{Multiple decoding paths}. 
Strategies for multiple encoding and decoding paths are widely studied for convolution networks operating on images ~\cite{tompson2015efficient,newell2016stacked,chen2018deeplab, xie2018interleaved,wang2020deep}. These approaches interconnect data maps at different resolution levels that mainly differ in the manner and extent that these connections are integrated. For example,~\cite{tompson2015efficient} use separated paths that are aggregated at the end of the network process, ~\cite{newell2016stacked,chen2018deeplab} use connections between encoding and decoding layers at the same resolution level, and ~\cite{xie2018interleaved,wang2020deep} use extensive integration between all resolution maps throughout the encoding (or decoding) stage. Similar to the latter approach, we also use multiple connections between differently-sized resolution maps, as shown in Fig.~\ref{fig:decoder_archl}. To the best of our knowledge, we present the first multi-resolution interconnections graph convolution network for mesh reconstruction. 

%%%%%%%%%%%%%%%%%%%%%%%%%%%%%%%%%%%%%%%%%%%%%%%er%%%%%%%%%%%%%%%%%%%%%%%%%%%%
%%%%%%%%%%%%%%%%%%%%%%%%%%%%%%%%%%%%%%%%%%%%%%%%%%%%%%%%%%%%%%%%%%%%%%%%%%%
% Generating synthetic image examples for closely interacting hands
%%%%%%%%%%%%%%%%%%%%%%%%%%%%%%%%%%%%%%%%%%%%%%%%%%%%%%%%%%%%%%%%%%%%%%%%%%%
%%%%%%%%%%%%%%%%%%%%%%%%%%%%%%%%%%%%%%%%%%%%%%%%%%%%%%%%%%%%%%%%%%%%%%%%%%%
\section{INTER-SH dataset}
\label{sec:synthetic_data}
In this section we describe a novel method for generating INTERacting Synthetic Hand (INTER-SH) examples of close hand-hand and hand-object interactions as captured by an RGB camera. We then use the readily available ground truth annotations in their various forms, such as hand/object masks, to develop and evaluate hand pose predictors.

We simulate hands in our synthetic dataset based on the left and right hands models of MANO \cite{MANO2017}. The objects in the hand-object interaction examples are partly taken from the YCB dataset \cite{YCB} and partly self-generated. 
%Overall, 29 objects were considered: 25 for training and 4 for test sets. 

%camera-captured scenes that result in degraded performance for a given hand pose predictor.  

%\textbf{Mesh definition.} 

%\begin{wrapfigure}{}{0.2\textwidth}
% \begin{figure}
%     \center
%     \includegraphics[width=0.3\linewidth]{./figures/Figure_3.png}
%   \caption{Hand keypoints locations numbered from 0 to 20. The kinematic chain of the hand model starts at the root joint and ends at the fingertips following the bone structure. 
%   % OPTIONAL:
% %   In Sec.~\ref{sec:synthetic_data} we propose a method to generate  interaction examples based on a mesh with local coordinate systems that we assign for joints along the kinematic chain.
% %   In Sec.~\ref{sec:model} we propose a Graph Convolution Network model where nodes are joints and edges form cliques of anatomical hand parts as described in Eq~\ref{eq:gcn_edges}.
%   }
% \label{fig:hand_keypoints}
% %\end{wrapfigure}
% \end{figure}

\begin{figure}[h!]
\begin{center}
    \includegraphics[width=0.95\linewidth]{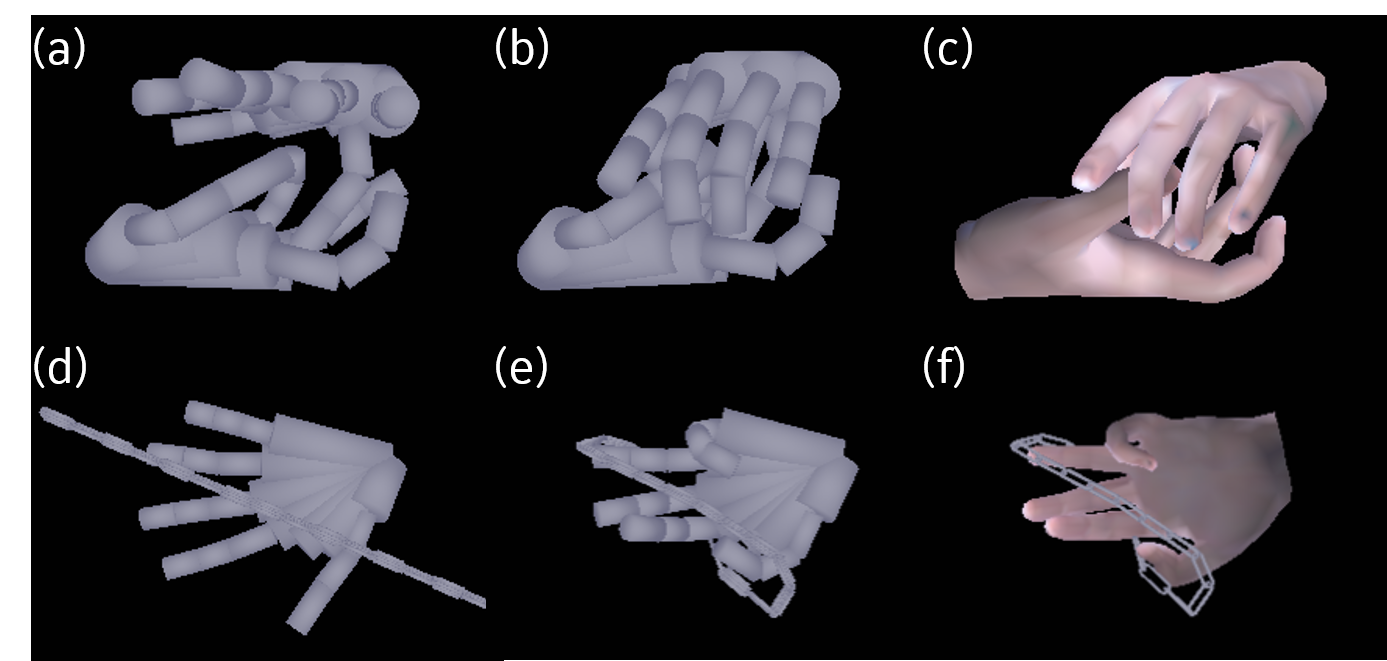}
\end{center}
   \caption{cylindrical presentation of MANO hand model \cite{MANO2017} (a-b) and a chain model (d-e) during data generation routine. The textured mesh scenes of (b) and (e) are shown in (c) and (f), respectively. In (a), the bottom hand is initialized with a random pose. The upper hand is in neutral pose. (b) Final pose of an interacting hands scene upon random motion of finger joints of both hand models. (d) The chain object is static and the hand is its neutral pose. (e) Final pose of hand interacting with chain. Validity of hand poses is based on collision detection of cylindrical rigid bodies. }
\label{fig:cyl_present}
\end{figure}

\subsection{Generating Mesh Interactions}
With the mesh models defined in Supplementary Sec.~\ref{sec:mesh_definition} and pose generating algorithm in Sec.~\ref{sec:generating_mesh_interactions}, we simulate interactions between two hand meshes, or a hand mesh and an object.

\textbf{Textures}. Similar to \cite{boukhayma20193d}, we assign the RGB values noted per 3D hand scan collected in \cite{MANO2017} to each vertex of the MANO mesh. The RGB assignment is performed based on the closest vertex in the original 3D scan to a vertex in the MANO mesh in neutral pose. The vertices' values are then interpolated along the mesh surface. The textures for the YCB objects \cite{YCB} are included in the dataset.

\textbf{Rendering.} We render the images using Pyrender \footnote{https://github.com/mmatl/pyrender}. For each hand-hand and hand-object configuration, we render object-only, hand-only, and complete scene images, as well as their corresponding segmentation maps. 

While our generated hand-object interaction scenes may resemble the output of previously proposed workflows, it is the ability to generate intricate hand-hand and hand-object interactions that contrasts our work from other generated datasets. %We compare our generated dataset's properties to others in Table \ref{tab:comparison_synth}.

%%%%%%%%%%%%%%%%%%%%%%%%%%%%%%%%%%%%%%%%%%%%%%%%%%%%%%%%%%%%%%%%%%%%%%%%%%%
%%%%%%%%%%%%%%%%%%%%%%%%%%%%%%%%%%%%%%%%%%%%%%%%%%%%%%%%%%%%%%%%%%%%%%%%%%%
% Hand pose model with anatomical GCN
%%%%%%%%%%%%%%%%%%%%%%%%%%%%%%%%%%%%%%%%%%%%%%%%%%%%%%%%%%%%%%%%%%%%%%%%%%%
%%%%%%%%%%%%%%%%%%%%%%%%%%%%%%%%%%%%%%%%%%%%%%%%%%%%%%%%%%%%%%%%%%%%%%%%%%%
%\section{Hand pose model with anatomically structured graph CNNs}
\section{Multi-path Mesh Decoder}
\label{sec:model}

\subsection{Architecture}
% % Motivation for the architecture:
% The generated images in Sec.~\ref{sec:synthetic_data} that capture close hand-object and hand-hand interactions may pose a particular challenge for hand pose estimation models (see experiments Section~\ref{sec:experiments} below). We conjecture that the reason is that they require to process high-level relationships and attributes between the hand parts, such as continuity, connectedness, or re-appearance relations. A model that accounts for such relations may provide more accurate localization of the hand joints in image-captured scenes where a hand penetrates a perforated object or interacts with another hand. To put focus on spatial relationships, we follow previous work in~\cite{ge20193d, doosti2020hopenet, dkulon2020cvpr} combining a Convolutional Neural Network for an image feature extractions stage, together with a Graph Convolution Network for learning high-level dependencies between joints. Different than previous works, we construct the GCN to maximize the relevant spatial joint structure learning, and remove potentially redundant structural information. This is achieved by forming GCN edges between joints and fingertips from the same anatomical part, namely the same finger. 

%---- Figure -----

\begin{figure*}[t]
\begin{center}
    \includegraphics[width=\textwidth]{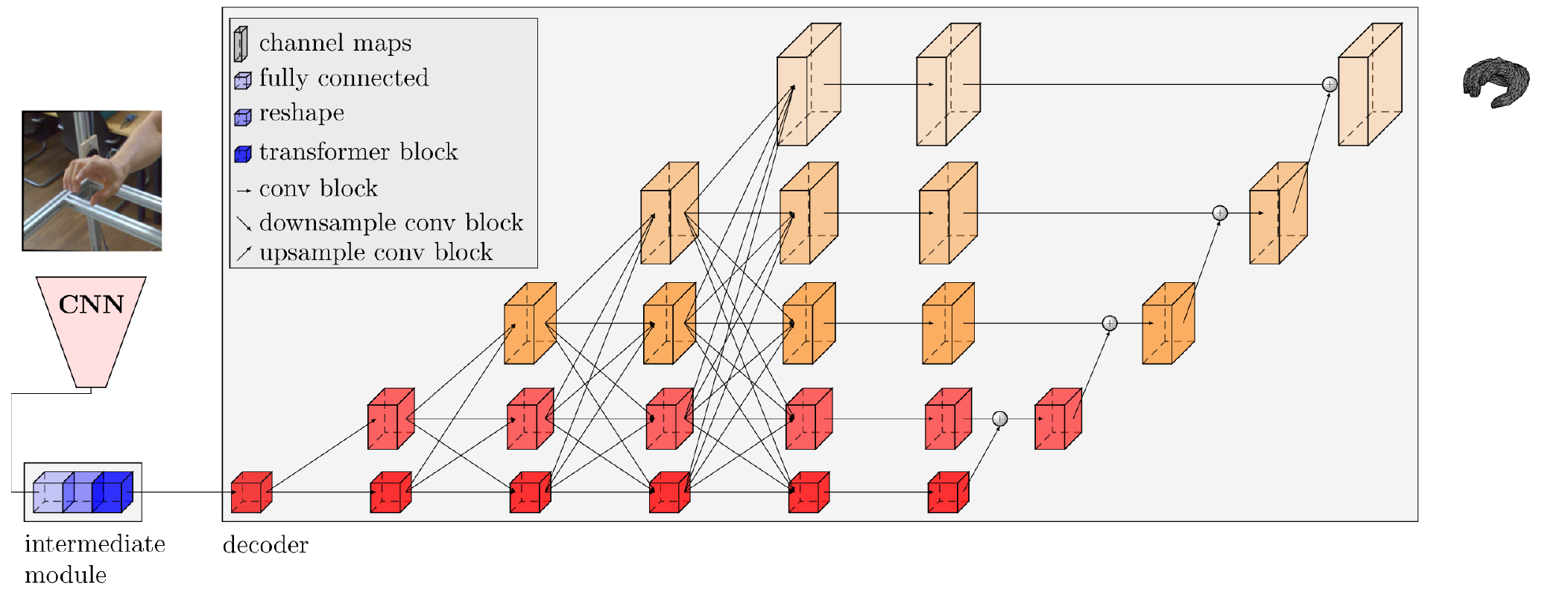}
\end{center}
   \caption{Architecture of the proposed hand pose estimation model. The input image is passed through a CNN encoder followed by an intermediate module and a GCN-based mesh decoder. Nodes in the GCN are hand vertices. In the proposed model, there are five stages. At each stage, a larger spatial resolution representation is added with respect to the previous stage. The convolutional stages comprise of multi-resolution blocks, consisting of a multi-resolution group convolution and a multi-resolution convolution.}
\label{fig:decoder_archl}
\end{figure*}

% Details for architecture:
%Figure~\ref{fig:decoder_archl} depicts our proposed framework. 
The input to our model is an image of size 224 $\times$ 224 centered around a single or interacting hand. The model's output is a prediction of 3D coordinates of hand mesh vertices $V$. Our network consists of 
(i) a Convolutional Neural Network encoder that provides a latent feature vector, 
(ii) an intermediate module comprising of a fully connected (FC) layer followed by a reshape layer and a Transformer \cite{Vaswani2017} block, 
and (iii) a multi-scale graph convolution mesh decoder. The purpose of the Transformer block is to promote short- and long-range interactions of the initial coarse mesh representation input to the decoder.

%MOVE TO IMPLEMENTATION:
% \textbf{CNN encoder}. The encoder's role is to distill relevant information from the input image for the mesh reconstruction task. \todo{[move to implmentation]The image encoder is pre-trained on the ImageNet classification task \textbf{[XX]}}. A feature vector $X$ of size 2048 is extracted from the last hidden layer of the encoder. This $X$ is then used as input to the intermediate module. 

% Based on empirical experiments, we compare the contribution of different image encoders to the mesh
% reconstruction task. Specifically, we experiment with Resnet50 \textbf{[X]}, HR-Net 18, and HR-Net 48 \textbf{[XX]}.

% \textbf{Intermediate module}. The feature vector $X$ is input to FC that outputs another feature vector $Y$ of size $C \times V_0$. $V_0$ denotes the number of vertices of the coarse mesh representation of the high resolution 3D model, whereas $C$ is the number of channels used at the decoder's first stage. $Y$ is then reshaped to an array $Z$ of size ($C, V_0$). The self-attention block takes $Z$ as input and outputs an array of similar size. The benefit of using a self-attention block has been previously shown in multiple tasks \textbf{[XX]}. We further corroborate these results in an ablation study shown herein.

\begin{table*}
    \begin{center}
    \begin{tabular}{|| c || l || c | c | c | c | }
    \hline
      Backbone & Model & PA-MPVPE & PA-MPJPE & F@5 mm & F@15 mm \\
    \hline \hline
        Resnet50 & Hasson \etal ~\cite{Hasson_2019_CVPR}  & 13.2 & - & 0.436 &  0.908  \\
        \hline
        Resnet50 & Kulon \etal ~\cite{dkulon2020cvpr}  & 8.6 & 8.3&  0.614 & 0.966 \\
        \hline
        Resnet50 & Doosti \etal ~\cite{doosti2020hopenet}$^*$  & 9.2  & -& - & -  \\
        \hline
        - & Pose2Mesh \cite{choi2020pose2mesh}  & 7.8 & 7.7 & 0.674 &  0.969 \\
        \hline
        - & I2LMeshNet \cite{moon2020i2lmeshnet}  & 7.6 & 7.4 & 0.681 &  0.973 \\
        \hline
        HRNet64 & METRO \cite{lin2021end}  & 6.7 & 6.8 & 0.717 &  0.981 \\
        \hline
        Resnet50 & Ours  & 7.6 & 7.4 & 0.674 &  0.97 \\
        \hline
        HRNet18 & Ours  & 7.0 & 7.2 & 0.701 & 0.977   \\
        \hline
        HRNet48 & Ours  & \textbf{6.6} & \textbf{6.7} & \textbf{0.719} & \textbf{0.981}   \\
    \hline \hline
    \end{tabular}
    \caption{Performance comparison with published methods, evaluated on FreiHAND online server. Bold faced text imply the best score.}
    \label{tab:external}
    \end{center}
\end{table*}

% \todo{Consider: move the CNN encoder and Intermediate module paragraphs to the implementation section. Then continue after the first introduction paragraph from here:}
\textbf{Multi-path decoder based on graph convolutions}. 

The implementation of the convolution function in our GCN is done by constructing a fully connected layer such that same linear operation is applied to all features of $e(v)$, for each $v$ ~\cite{lim2018simple,dkulon2020cvpr}. More formally, let $f(e(v))$ be the sequence of concatenated features for the nodes in $e(v)$. Let $g$ be a convolution kernel, and $f$ the features of all nodes. Then the convolution operation is:
\begin{equation}\label{eq:graph_conv}
    (f * g)_v = \sum_{i=1}^{|f(e(v))|} g_i f(e(v))_i.
\end{equation}
where $(f * g)_v$ is the output of the convolution window for $v$. The convolution is followed by the Exponential Linear Unit activation function that was selected based on empirical evidence. The spatial neighbourhood for the convolution operation were defined using a spiral patch operator similar to \cite{lim2018simple, dkulon2020cvpr}. 

Previously proposed GCN models use a single mesh decoding path. In contrast, we implement a strategy that allow for multiple and parallel mesh decoding trajectories, similar to previous 
% ideas presented for 
multi-scale fusion works on image encoding (e.g. ~\cite{xie2018interleaved,wang2020deep}). The structure of the graph convolution layers and their connections is shown in Fig.~\ref{fig:decoder_archl}. 

In the proposed model, there are five stages. At each stage, a larger spatial resolution representation is added with respect to the previous stage. Apart from the first stage, all stages comprise of multi-resolution blocks, consisting of a multi-resolution group convolution and a multi-resolution convolution \cite{wang2020deep}. For the latter convolution operator, a connection between input channels and output channels of differently-sized resolution data is achieved by utilizing precomputed upsampling and downsampling matrices.

To compute the upsampling and downsampling matrices, we construct five coarse mesh representations of the original mesh. At each coarsening stage, the number of vertices are reduced by a factor of 2 \cite{Ranjan2018}. The downsampling matrices of each stage are obtained by iteratively contracting vertex pairs based on quadric error metrics. The vertices of the downsampled mesh are a subset of the original mesh vertices. Vertices discarded during downsampling are projected into the closest triangle of the coarse mesh. Then, the barycentric coordinates of the projected vertex are used to define interpolation weights for the upsampling matrix.

The convolution blocks used within our decoder are structured as a variant of basic residual blocks \cite{HeResnet50}. Specifically, we replace the 2D convolution layers with graph convolutions defined in Equation \ref{eq:graph_conv}, and the batch normalization layers with layer normalizations. Data fusion layers of \cite{wang2020deep} were also incorporated and modified similar to our altered ResNet blocks.

% Loss
\subsection{Training}
As loss function we used the L1 norm between the ground truth mesh vertices $\mathcal{V} \in \mathbb{R}^{M \times 3}$ to the predicted vertices $\bar{\mathcal{V}}$. That is,
\begin{equation}
    \mathcal{L} = \frac{1}{M}\sum_{j=1}^{M} ||\mathcal{V}_{j} - \bar{\mathcal{V}}_{j}||_1.
\end{equation}

% %%%%%%%%%%%%%%%%%%%%%%%%%%%%%%%%%%%%%%%%%%%%%%%%%%
% % Implementation
% %%%%%%%%%%%%%%%%%%%%%%%%%%%%%%%%%%%%%%%%%%%%%%%%%%
% \subsection{Implementation}
To train our network we use the Adam solver \cite{adamSolver} with learning rate $10^{-4}$ for 200 epochs. Learning rate decay with factor 0.5 occurs after every 50 epochs. The images are normalized with the mean 0.5 and standard deviation of 1.0. We augment the data with random image crops and transformations. We use batch size of 32. Ground truth 3D hand vertices were translated such that the middle metacarpophalangeal joint 
% (i.e. joint number 9 in Figure \ref{fig:hand_keypoints})
is at the origin.

% \todo{
% More on the training process? Was there a mix of real and synthetic data? }

%%%%%%%%%%%%%%%%%%%%%%%%%%%%%%%%%%%%%%%%%%%%%%%%%%%%%%%%%%%%%%%%%%%%%%%%%%%
%%%%%%%%%%%%%%%%%%%%%%%%%%%%%%%%%%%%%%%%%%%%%%%%%%%%%%%%%%%%%%%%%%%%%%%%%%%
% Experiments and evaluation
%%%%%%%%%%%%%%%%%%%%%%%%%%%%%%%%%%%%%%%%%%%%%%%%%%%%%%%%%%%%%%%%%%%%%%%%%%%
%%%%%%%%%%%%%%%%%%%%%%%%%%%%%%%%%%%%%%%%%%%%%%%%%%%%%%%%%%%%%%%%%%%%%%%%%%%

\section{Experiments}
\label{sec:experiments}
In this section, we present the datasets and corresponding evaluation protocols used to compare our method to the state-of-the-art. In addition, we provide an analysis of our framework and suggest new metrics that help define a set of hard-cases for the hand-pose estimation problem given a RGB image.

\subsection{Datasets}
Our experiments were limited to detests that provide ground truth annotations for hand mesh reconstruction. We avoided datasets that only contains joints annotations, since techniques to infer mesh from joint keypoints (such as regressing a MANO model\cite{MANO2017}) tend to generate artifacts that harm the training procedure of the model. For hand mesh reconstruction this leaves us with the Friehand datasets~\cite{Freihand2019}, which is the main benchmark for hand pose models today, and our own INTER-SH dataset focusing on challenging hand interactions examples.

FreiHAND is a dataset with 130,240 training images \cite{Freihand2019}, including single hands and hand-object interactions, taken with a green screen as a background. The test set, consisting of 3,960 samples, was collected without the green screen in indoor and outdoor environments. The FreiHAND test set annotations is not available. The evaluation of the hand pose predictions is performed through submission of results to an online competition. 

% We also employ our INTERMESH dataset, as described in Section \ref{sec:synthetic_data}. We compare the prediction accuracy of different Using this dataset, we test the influence of different hand visibility ratios, described in the next subsection (Equation \ref{eq:vis_ratio}), on different hand prediction models. To this end, we divided our training data size equally among four different visibility ratio categories: 0.40-0.60, 0.60-0.80, 0.80-0.95, and 0.95-1.00. Each category contains 10,000 images. 

% For both FreiHAND and HO-3D datasets, their test set annotations are not available. The evaluation of the hand pose predictions is performed through submission of results to an online competition. The test set annotations for InterHand2.6M are included in the original dataset. 
%

% Finally, we also employ our generated dataset, as described in Section \ref{sec:synthetic_data}. Using this dataset, we test the influence of different hand visibility ratios, described in the next subsection (Equation \ref{eq:vis_ratio}), on different hand prediction models. To this end, we divided our training data size equally among four different visibility ratio categories: 0.40-0.60, 0.60-0.80, 0.80-0.95, and 0.95-1.00. Each category contains 10,000 images. 

\subsection{Evaluation Metrics}
\label{sec:interaction_metric}
%%%%%% Standard metrics
PA-MPJPE reconstruction error \cite{Zhou2019} performs a 3D alignment using Procrustes analysis (PA) \cite{Gower1975} followed by Mean-Per-Joint-Position-Error (MPJPE)\cite{Human36} computation. MPJPE measures the average Euclidean distances between the ground truth joints and the predicted joints.  In addition, we report the F-score at a given threshold d (F@d) defined as the harmonic mean of precision and recall \cite{Knapitsch2017}. Moreover, we compute the percentage of correct points (3D PCK) for different thresholds, and the Area Under Curve (AUC) for PCK.

\begin{table*}
    \begin{center}
    \begin{tabular}{|| c || c | c | c | c |}
    \hline
     &  \multicolumn{4}{c|}{$VR$ value range} \\
    \hline
      Model & [0.40, 0.60] & [0.60, 0.80] & [0.80, 0.95] & [0.95, 1.00] \\
    \hline \hline
        Hason \etal~\cite{Hasson_2019_CVPR}$^{\ddagger}$   & 0.43 (15.65) & 0.47 (14.91)& 0.48 (14.60)& 0.49 (13.64) \\
        \hline
        Kulon \etal~\cite{dkulon2020cvpr}$^{\dagger}$   & \underline{0.64 (10.24)}  & \textbf{0.65 (9.79)} & \underline{0.66 (9.58)}& \underline{0.71 (7.51)}  \\
        \hline
        Doosti \etal~\cite{doosti2020hopenet}$^*$  & 0.49 (13.86) & 0.50 (13.78) & 0.50 (13.65) & 0.52 (12.56)\\
        \hline
        Lin \etal~\cite{lin2021end}$^*$  & 0.59 (11.27) & 0.59 (11.59) & 0.60 (11.26) & 0.63 (9.67)\\
        \hline
        Ours  & \textbf{0.65 (9.69)}  & \underline{0.64 (10.28)}  & \textbf{0.66 (9.57)} & \textbf{0.71 (7.49)} \\
    % \hline \hline     \hline \hline
    %     Hason \etal, 2019~\cite{Hasson_2019_CVPR}$^{\ddagger}$&  15.65  & 14.91 & 14.60 & 13.64  \\
    %     \hline
    %     Kulon \etal, 2020~\cite{dkulon2020cvpr}$^{\dagger}$&  \textbf{10.24}  & \textbf{9.79} & \textbf{9.58} & \textbf{7.51} \\
    %     \hline
    %     Doosti \etal, 2020~\cite{doosti2020hopenet}$^*$& 13.86 & 13.78 & 13.65 & 12.56 \\
    %     \hline
    %     Ours  &  \bfff{10.55} & \bfff{10.12}  & \bfff{9.90} & \bfff{7.92}  \\
    \hline \hline
    \end{tabular}
    \caption{Results on INTER-SH using Resnet50 as backbone. Left hand pose AUC results in \%  on our own dataset of synthetic closely interacting hands images categorized by the hand's visibility ratio. Pose error in mm shown in parenthesis. AUC of 3D PCK is computed in an interval from 0 to 20 mm with 20 equally spaced thresholds. Bold faced and underlined text imply best and second-best score, respectively. The comparison is made against various methods that were either modified or recreated as code was unavailable.}
    \label{tab:internal}
    \end{center}
\end{table*}

%%%%%% The problem 
To further study the effect of the degree of hand-hand interactions on these metrics, we propose to group hand-hand interaction scenarios based on a visibility ratio metric. 

%%%%%% The novel proposed metric
{\bf Visibility ratio metric. }
% For a synthetically-generated image of a hand interaction scene, we additionally define the so called visibility ratio metric. 
Recall from Sec.~\ref{sec:synthetic_data} that in addition to generating a synthetic image, we also generate separate masks for the hands and objects that appear in the image and a mask image of the complete scene.  Let the mask of a single object or a hand be defined as $M^{\textrm{object only}}$ and that of complete scene as $M^{\textrm{scene}}$. The visibility ratio $VR$ of object $o$ is defined as:
\begin{equation} \label{eq:vis_ratio}
    VR_o = \sum_{i=1}^N  \mathds{1}_o(M^{\textrm{scene}}_i) / \sum_{i=1}^N \mathds{1}_o(M^{\textrm{object only}}_i),
\end{equation}
where $N$ is the number of pixels in a mask image, and $\mathds{1}_o(x)$ is the indicator function that returns one if pixel $x$ is an element of object $o$, and zero otherwise. If visibility ratio is one, then the additional object should minimally affect the prediction capability of a hand-pose predictive model. In Sec.~\ref{experimental_results}, we study the relation between the visibility ratio of hands in an image and the prediction score of various models for closely interacting hands.

\subsection{Results}
\label{experimental_results}
We report the pose estimation and mesh reconstruction accuracy of our model and compare it against other recent models. One of the models is the METRO model~\cite{lin2021end} that we have implemented ourselves, which produced state-of-the-art results on FreiHAND \cite{Freihand2019}. We also compared results to other models that produce mesh, keypoints, or both, namely  \cite{choi2020pose2mesh},~\cite{Moon_2020_ECCV_I2L-MeshNet},~\cite{dkulon2020cvpr},~\cite{doosti2020hopenet}, and~\cite{Hasson_2019_CVPR}. These last two models were originally designed to estimate both hand and object poses. For our experiments we modified them to predict positions of only hand mesh or joints. The modified methods are denoted as \cite{Hasson_2019_CVPR}$^{\ddagger}$ and \cite{doosti2020hopenet}$^*$. Also, since the code model for~\cite{dkulon2020cvpr} and ~\cite{lin2021end} was not released by the authors, we have reproduced their models to the best of our availability and denote it as \cite{dkulon2020cvpr}$^{\dagger}$ and \cite{lin2021end}$^{\dagger}$.

%%%%% Evaluation on Hand-Object datasets
\noindent\textbf{Evaluation on FreiHAND}. 
Table \ref{tab:external} compares the results of our model to others on the FreiHAND dataset. 
% which show that our model is currently SOTA on FreiHAND. In particular, 
Our model outperforms previous works. In particular, we outperform the recent METRO model \cite{lin2021end}  that uses a deeper backbone than us. Since we also use a Transformer layer, like METRO, we further explore its contribution in our ablation studies below. 
Our model also improves upon ~\cite{dkulon2020cvpr} that uses a single path mesh decoding approach. This raises the question regrading the contribution of our multi-path decoding strategy. This question is further explored in our ablation studies.  
We additionally compare the contribution of different image encoders to the mesh reconstruction task. Specifically, we experiment with Resnet50 \cite{he2015deep}, HR-Net with initial 18 and 48 channels \cite{wang2020deep}. Four output results using HR-Net 48 channels as backbone are shown in Figure \ref{fig:friehand_model_results}(a).

\begin{table*}
    \tabcolsep 0.05cm
    \begin{center}
    \begin{tabular}{| c | c | c | c | c | c | c | }
    \hline
      Single path & Multi-path & Attention & PA-MPVPE & PA-MPJPE & F@5 mm & F@15 mm   \\
    \hline \hline
        v &   &  & 7.4 & 7.5 & 0.686 & 0.974 \\
        \hline
        & v  &  & 7.3 & 7.4  & 0.69 & 0.975\\
        \hline
        & v  & v  & 7.0 & 7.2  & 0.701 & 0.977\\
    \hline \hline
    \end{tabular}
    \caption{Results for ablation studies evaluated using online Freihand server. HRNet18 is used here as backbone for all experiments.}
    \label{tab:ablation}
    \end{center}
\end{table*}

\begin{figure*}
\begin{center}
	\includegraphics[width=\textwidth]{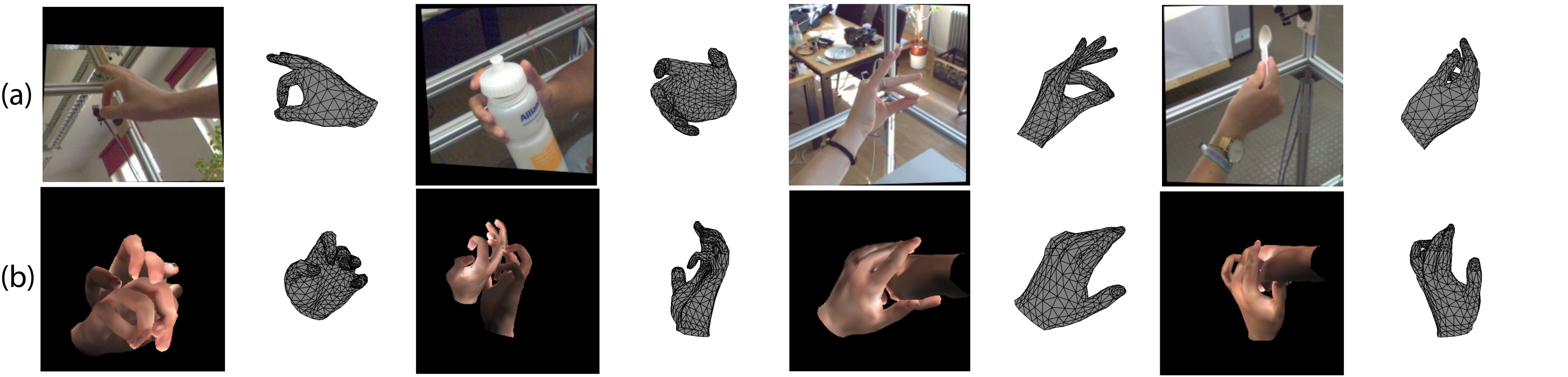}
	\end{center}
	\vspace{-0.2cm} 
	\caption{Mesh reconstruction results for our proposed model on the (a) FrieHAND dataset and (b) INTER-SH dataset. Left column is the original input and the right column is the predicted mesh,
	}
	\label{fig:friehand_model_results}
\end{figure*}

%%%%% Evaluation on generated Hand-Hand dataset
\noindent\textbf{Evaluation on INTER-SH}. 
To further assess the performance of existing models on closely interacting hands, we train and test them using our generated hand-hand interacting images in INTER-SH. Since here we had to run and test other models, we either used a modified published original version, or a self-implemented version. Tab.~\ref{tab:internal} show the models' prediction evaluation metrics on the test set computed only for the left hand in the scenes. We additionally categorize the metrics based on the visibility ratio, defined in Section \ref{sec:interaction_metric}. Our method outperforms recently published algorithms.
% Note that \cite{Hasson_2019_CVPR}$^{\ddagger}$ model underperforms compared to others despite the fact the generated data utilizes the MANO hand model which \cite{Hasson_2019_CVPR}$^{\ddagger}$ model is based on. This is perhaps due to the relative high angular rotation of the joints compared to the mean hand pose of the MANO model used in \cite{Hasson_2019_CVPR}$^{\ddagger}$. 
Four output examples of our model for this data are presented in Fig. \ref{fig:friehand_model_results}(b). Further outputs are shown in the supplementary section. 

It is not surprising to observe that all hand pose models improve their prediction accuracy when the visibility ratio of the hand increases. However, it is interesting to note that there is a relative large improvement at the $VR$ range of 0.95-1.00 compared to $VR$ range of 0.80-0.95. Thus, we learn that it is imperative to balance the dataset used for model training and testing to encompass different $VR$ ranges in order to obtain a more robust hand pose predictor. It is also interesting to notice that results on the low visibility INTER-SH examples are lower than the results obtained on FrieHAND dataset, suggesting INTER-SH has complementary examples to FreiHAND and poses a new interesting challenge to the hand mesh reconstruction community. 

\noindent\textbf{Ablation studies}.
To investigate the contribution of the different components to the performance of our model, we conducted ablation studies in which we compare versions of our model with removed modules and stages. The versions were tested on the FrieHAND test dataset. The ablations contained removal of the attention block and removal of the multi-path component (reducing the model to a single path decoding). The results in Tab.~\ref{tab:ablation} show that both the attention and multi-path components contribute to performance (notice that the attention block also adds depth). Specifically, comparing the single and multi path strategies that contain the exact same backbone and extract same network depth, we conclude that multi-path reconstruction is useful for the task of interacting hand mesh reconstruction. 
\section{Conclusions}
We have proposed a new multi-stream graph convolution network that provides state-of-the-art results for hand mesh reconstruction from singular RGB image capturing two core scenarios: hand-object and hand-hand interactions. Scenarios were tested using both real and synthetic data. We additionally present a new method to generate hand-hand and hand-object interactions data that may benefit the hand pose estimation community.

{\small
\bibliographystyle{ieee_fullname}
\bibliography{egbib}

\begin{thebibliography}{10}\itemsep=-1pt

\bibitem{boukhayma20193d}
Adnane Boukhayma, Rodrigo~de Bem, and Philip~HS Torr.
\newblock 3d hand shape and pose from images in the wild.
\newblock In {\em Proceedings of the IEEE Conference on Computer Vision and
  Pattern Recognition}, pages 10843--10852, 2019.

\bibitem{Brahmbhatt_2020_ECCV}
Samarth Brahmbhatt, Chengcheng Tang, Christopher~D. Twigg, Charles~C. Kemp, and
  James Hays.
\newblock {ContactPose}: A dataset of grasps with object contact and hand pose.
\newblock In {\em The European Conference on Computer Vision (ECCV)}, August
  2020.

\bibitem{CaiEtAl2018}
Yujun Cai, Liuhao Ge, Jianfei Cai, and Junsong Yuan.
\newblock Weakly-supervised 3d hand pose estimation from monocular rgb images.
\newblock In Vittorio Ferrari, Martial Hebert, Cristian Sminchisescu, and Yair
  Weiss, editors, {\em Computer Vision -- ECCV 2018}, pages 678--694, Cham,
  2018. Springer International Publishing.

\bibitem{YCB}
B. {Calli}, A. {Singh}, A. {Walsman}, S. {Srinivasa}, P. {Abbeel}, and A.~M.
  {Dollar}.
\newblock The ycb object and model set: Towards common benchmarks for
  manipulation research.
\newblock In {\em 2015 International Conference on Advanced Robotics (ICAR)},
  pages 510--517, 2015.

\bibitem{chen2018deeplab}
Liang-Chieh Chen, George Papandreou, Iasonas Kokkinos, Kevin Murphy, and Alan~L
  Yuille.
\newblock Deeplab: Semantic image segmentation with deep convolutional nets,
  atrous convolution, and fully connected crfs.
\newblock {\em IEEE transactions on pattern analysis and machine intelligence},
  40(4):834--848, 2018.

\bibitem{choi2020pose2mesh}
Hongsuk Choi, Gyeongsik Moon, and Kyoung~Mu Lee.
\newblock Pose2mesh: Graph convolutional network for 3d human pose and mesh
  recovery from a 2d human pose.
\newblock In {\em European Conference on Computer Vision}, pages 769--787.
  Springer, 2020.

\bibitem{doosti2020hopenet}
Bardia Doosti, Shujon Naha, Majid Mirbagheri, and David Crandall.
\newblock Hope-net: A graph-based model for hand-object pose estimation, 2020.

\bibitem{HernandoEtAl2018}
Guillermo Garcia-Hernando, Shanxin Yuan, Seungryul Baek, and Tae-Kyun Kim.
\newblock First-person hand action benchmark with rgb-d videos and 3d hand pose
  annotations.
\newblock In {\em In Proc. of the IEEE Conf. on Computer Vision and Pattern
  Recognition (CVPR)}, pages 409--419, 2018.

\bibitem{ge20193d}
Liuhao Ge, Zhou Ren, Yuncheng Li, Zehao Xue, Yingying Wang, Jianfei Cai, and
  Junsong Yuan.
\newblock 3d hand shape and pose estimation from a single rgb image, 2019.

\bibitem{Gower1975}
J.~C. Gower.
\newblock Generalized procrustes analysis.
\newblock {\em Psychometrika}, 1975.

\bibitem{HampaliEtAl2019}
Shreyas Hampali, Markus Oberweger, Mahdi Rad, and Vincent Lepetit.
\newblock {HO-3D:} {A} multi-user, multi-object dataset for joint 3d
  hand-object pose estimation.
\newblock {\em CoRR}, abs/1907.01481, 2019.

\bibitem{Hasson_2019_CVPR}
Yana Hasson, Gul Varol, Dimitrios Tzionas, Igor Kalevatykh, Michael~J. Black,
  Ivan Laptev, and Cordelia Schmid.
\newblock Learning joint reconstruction of hands and manipulated objects.
\newblock In {\em Proceedings of the IEEE/CVF Conference on Computer Vision and
  Pattern Recognition (CVPR)}, June 2019.

\bibitem{he2015deep}
Kaiming He, Xiangyu Zhang, Shaoqing Ren, and Jian Sun.
\newblock Deep residual learning for image recognition, 2015.

\bibitem{HeResnet50}
K. {He}, X. {Zhang}, S. {Ren}, and J. {Sun}.
\newblock Deep residual learning for image recognition.
\newblock In {\em 2016 IEEE Conference on Computer Vision and Pattern
  Recognition (CVPR)}, pages 770--778, 2016.

\bibitem{Human36}
C. {Ionescu}, D. {Papava}, V. {Olaru}, and C. {Sminchisescu}.
\newblock Human3.6m: Large scale datasets and predictive methods for 3d human
  sensing in natural environments.
\newblock {\em IEEE Transactions on Pattern Analysis and Machine Intelligence},
  36(7):1325--1339, 2014.

\bibitem{iqbalEtAl2018}
Umar Iqbal, Pavlo Molchanov, Thomas Breuel, Juergen Gall, and Jan Kautz.
\newblock Hand pose estimation via latent 2.5d heatmap regression, 2018.

\bibitem{Ketchel2005}
J. Ketchel and P. Larochelle.
\newblock Collision detection of cylindrical rigid bodies using line geometry.
\newblock In {\em Proceedings of IDETC/CIE}, 2005.

\bibitem{adamSolver}
Diederik~P. Kingma and Jimmy Ba.
\newblock Adam: {A} method for stochastic optimization.
\newblock In Yoshua Bengio and Yann LeCun, editors, {\em 3rd International
  Conference on Learning Representations, {ICLR} 2015, San Diego, CA, USA, May
  7-9, 2015, Conference Track Proceedings}, 2015.

\bibitem{kipf2017semi}
Thomas~N Kipf and Max Welling.
\newblock Semi-supervised classification with graph convolutional networks.
\newblock In {\em International Conference on Learning Representations (ICLR)},
  2017.

\bibitem{Knapitsch2017}
Arno Knapitsch, Jaesik Park, Qian-Yi Zhou, and Vladlen Koltun.
\newblock Tanks and temples: Benchmarking large-scale scene reconstruction.
\newblock {\em ACM Trans. Graph.}, 36(4), 2017.

\bibitem{dkulon2020cvpr}
Dominik Kulon, Riza~Alp Guler, Iasonas Kokkinos, Michael~M. Bronstein, and
  Stefanos Zafeiriou.
\newblock Weakly-supervised mesh-convolutional hand reconstruction in the wild.
\newblock In {\em The IEEE Conference on Computer Vision and Pattern
  Recognition (CVPR)}, 2020.

\bibitem{lim2018simple}
Isaak Lim, Alexander Dielen, Marcel Campen, and Leif Kobbelt.
\newblock A simple approach to intrinsic correspondence learning on
  unstructured 3d meshes.
\newblock In {\em Proceedings of the European Conference on Computer Vision
  (ECCV)}, pages 0--0, 2018.

\bibitem{Lin_2021_WACV}
Fanqing Lin, Connor Wilhelm, and Tony Martinez.
\newblock Two-hand global 3d pose estimation using monocular rgb.
\newblock In {\em Proceedings of the IEEE/CVF Winter Conference on Applications
  of Computer Vision (WACV)}, pages 2373--2381, January 2021.

\bibitem{lin2021end}
Kevin Lin, Lijuan Wang, and Zicheng Liu.
\newblock End-to-end human pose and mesh reconstruction with transformers.
\newblock 2021.

\bibitem{Graspit}
A.~T. {Miller} and P.~K. {Allen}.
\newblock Graspit! a versatile simulator for robotic grasping.
\newblock {\em IEEE Robotics Automation Magazine}, 11(4):110--122, 2004.

\bibitem{moon2020interhand26m}
Gyeongsik Moon, Shoou i Yu, He Wen, Takaaki Shiratori, and Kyoung~Mu Lee.
\newblock Interhand2.6m: A dataset and baseline for 3d interacting hand pose
  estimation from a single rgb image, 2020.

\bibitem{moon2020i2lmeshnet}
Gyeongsik Moon and Kyoung~Mu Lee.
\newblock I2l-meshnet: Image-to-lixel prediction network for accurate 3d human
  pose and mesh estimation from a single rgb image, 2020.

\bibitem{Moon_2020_ECCV_I2L-MeshNet}
Gyeongsik Moon and Kyoung~Mu Lee.
\newblock I2l-meshnet: Image-to-lixel prediction network for accurate 3d human
  pose and mesh estimation from a single rgb image.
\newblock In {\em European Conference on Computer Vision (ECCV)}, 2020.

\bibitem{MuellerEtAl2018}
Franziska Mueller, Florian Bernard, Oleksandr Sotnychenko, Srinath~Sridhar
  Dushyant~Mehta, Dan Casas, and Christian Theobalt.
\newblock {GAN}erated hands for real-time 3d hand tracking from monocular rgb.
\newblock In {\em In Proc. of the IEEE Conf. on Computer Vision and Pattern
  Recognition (CVPR)}, pages 49--59, 2018.

\bibitem{mueller2019real}
Franziska Mueller, Micah Davis, Florian Bernard, Oleksandr Sotnychenko, Mickeal
  Verschoor, Miguel~A Otaduy, Dan Casas, and Christian Theobalt.
\newblock Real-time pose and shape reconstruction of two interacting hands with
  a single depth camera.
\newblock {\em ACM Transactions on Graphics (TOG)}, 38(4):1--13, 2019.

\bibitem{newell2016stacked}
Alejandro Newell, Kaiyu Yang, and Jia Deng.
\newblock Stacked hourglass networks for human pose estimation.
\newblock In {\em European conference on computer vision}, pages 483--499.
  Springer, 2016.

\bibitem{oikonomidis2012tracking}
Iasonas Oikonomidis, Nikolaos Kyriazis, and Antonis~A Argyros.
\newblock Tracking the articulated motion of two strongly interacting hands.
\newblock In {\em 2012 IEEE Conference on Computer Vision and Pattern
  Recognition}, pages 1862--1869. IEEE, 2012.

\bibitem{Panteleris2018UsingAS}
P. Panteleris, I. Oikonomidis, and A. Argyros.
\newblock Using a single rgb frame for real time 3d hand pose estimation in the
  wild.
\newblock {\em 2018 IEEE Winter Conference on Applications of Computer Vision
  (WACV)}, pages 436--445, 2018.

\bibitem{Ranjan2018}
Anurag Ranjan, Timo Bolkart, Soubhik Sanyal, and Michael~J. Black.
\newblock Generating 3d faces using convolutional mesh autoencoders.
\newblock In Vittorio Ferrari, Martial Hebert, Cristian Sminchisescu, and Yair
  Weiss, editors, {\em Computer Vision -- ECCV 2018}, pages 725--741, Cham,
  2018. Springer International Publishing.

\bibitem{MANO2017}
Javier Romero, Dimitrios Tzionas, and Michael~J. Black.
\newblock Embodied hands: Modeling and capturing hands and bodies together.
\newblock {\em ACM Transactions on Graphics, (Proc. SIGGRAPH Asia)}, Nov. 2017.

\bibitem{simon2017hand}
Tomas Simon, Hanbyul Joo, Iain Matthews, and Yaser Sheikh.
\newblock Hand keypoint detection in single images using multiview
  bootstrapping.
\newblock In {\em Proceedings of the IEEE conference on Computer Vision and
  Pattern Recognition}, pages 1145--1153, 2017.

\bibitem{spurr2020weakly}
Adrian Spurr, Umar Iqbal, Pavlo Molchanov, Otmar Hilliges, and Jan Kautz.
\newblock Weakly supervised 3d hand pose estimation via biomechanical
  constraints, 2020.

\bibitem{spurr2018}
Adrian Spurr, Jie Song, Seonwook Park, and Otmar Hilliges.
\newblock Cross-modal deep variational hand pose estimation, 2018.

\bibitem{GRAB2020}
Omid Taheri, Nima Ghorbani, Michael~J. Black, and Dimitrios Tzionas.
\newblock {GRAB}: A dataset of whole-body human grasping of objects.
\newblock In {\em European Conference on Computer Vision (ECCV)}, 2020.

\bibitem{tompson2015efficient}
Jonathan Tompson, Ross Goroshin, Arjun Jain, Yann LeCun, and Christoph Bregler.
\newblock Efficient object localization using convolutional networks.
\newblock In {\em Proceedings of the IEEE conference on computer vision and
  pattern recognition}, pages 648--656, 2015.

\bibitem{Tzionas2016}
Dimitrios Tzionas, Luca Ballan, Abhilash Srikantha, Pablo Aponte, Marc
  Pollefeys, and Juergen Gall.
\newblock Capturing hands in action using discriminative salient points and
  physics simulation.
\newblock {\em International Journal of Computer Vision (IJCV)}, 2016.

\bibitem{Vaswani2017}
Ashish Vaswani, Noam Shazeer, Niki Parmar, Jakob Uszkoreit, Llion Jones,
  Aidan~N Gomez, \L~ukasz Kaiser, and Illia Polosukhin.
\newblock Attention is all you need.
\newblock In I. Guyon, U.~V. Luxburg, S. Bengio, H. Wallach, R. Fergus, S.
  Vishwanathan, and R. Garnett, editors, {\em Advances in Neural Information
  Processing Systems}, volume~30. Curran Associates, Inc., 2017.

\bibitem{wang2020deep}
Jingdong Wang, Ke Sun, Tianheng Cheng, Borui Jiang, Chaorui Deng, Yang Zhao,
  Dong Liu, Yadong Mu, Mingkui Tan, Xinggang Wang, et~al.
\newblock Deep high-resolution representation learning for visual recognition.
\newblock {\em IEEE transactions on pattern analysis and machine intelligence},
  2020.

\bibitem{xie2018interleaved}
Guotian Xie, Jingdong Wang, Ting Zhang, Jianhuang Lai, Richang Hong, and
  Guo-Jun Qi.
\newblock Interleaved structured sparse convolutional neural networks.
\newblock In {\em Proceedings of the IEEE Conference on Computer Vision and
  Pattern Recognition}, pages 8847--8856, 2018.

\bibitem{xu2018how}
Keyulu Xu, Weihua Hu, Jure Leskovec, and Stefanie Jegelka.
\newblock How powerful are graph neural networks?
\newblock In {\em International Conference on Learning Representations}, 2019.

\bibitem{YangYao2019}
L. {Yang} and A. {Yao}.
\newblock Disentangling latent hands for image synthesis and pose estimation.
\newblock In {\em 2019 IEEE/CVF Conference on Computer Vision and Pattern
  Recognition (CVPR)}, pages 9869--9878, 2019.

\bibitem{ZhangEtAl2016}
Jiawei Zhang, Jianbo Jiao, Mingliang Chen, Liangqiong Qu, Xiaobin Xu, and
  Qingxiong Yang.
\newblock 3d hand pose tracking and estimation using stereo matching.
\newblock {\em CoRR}, 2016.

\bibitem{zhang2019}
Xiong Zhang, Qiang Li, Hong Mo, Wenbo Zhang, and Wen Zheng.
\newblock End-to-end hand mesh recovery from a monocular rgb image, 2019.

\bibitem{Zhou2019}
X. {Zhou}, M. {Zhu}, G. {Pavlakos}, S. {Leonardos}, K.~G. {Derpanis}, and K.
  {Daniilidis}.
\newblock Monocap: Monocular human motion capture using a cnn coupled with a
  geometric prior.
\newblock {\em IEEE Transactions on Pattern Analysis and Machine Intelligence},
  41(4):901--914, 2019.

\bibitem{zhou2020}
Yuxiao Zhou, Marc Habermann, Weipeng Xu, Ikhsanul Habibie, Christian Theobalt,
  and Feng Xu.
\newblock Monocular real-time hand shape and motion capture using multi-modal
  data, 2020.

\bibitem{ZimmermannBrox2017}
Christian Zimmermann and Thomas Brox.
\newblock Learning to estimate 3d hand pose from single rgb images.
\newblock In {\em IEEE International Conference on Computer Vision (ICCV)},
  2017.

\bibitem{Freihand2019}
Christian Zimmermann, Duygu Ceylan, Jimei Yang, Bryan Russel, Max Argus, and
  Thomas Brox.
\newblock Freihand: A dataset for markerless capture of hand pose and shape
  from single rgb images.
\newblock In {\em IEEE International Conference on Computer Vision (ICCV)},
  2019.

\end{thebibliography}
}

%%%%%%%%%%%%%%%%%%%%%%%%%%%%%%%%%%%%%%%%%%%%%%%%%%
% Supplementary
%%%%%%%%%%%%%%%%%%%%%%%%%%%%%%%%%%%%%%%%%%%%%%%%%%
\newpage
\clearpage
\section{Supplementary}
In the following sections we explain in more detail our synthetic hand-hand and hand-object synthetic generator. In addition we present additional qualitative results of our proposed model on both Freihand and the Inter-SH dataset.

\subsection{Mesh Movement via Local Coordinate Systems}
\label{sec:mesh_definition}
Previous use of the MANO model includes randomizing the MANO parameters to generate random single hand pose examples, or using its mesh structure to generate examples of a hand grasping a static object. The grasp is generated by minimizing an energy function based on the hand-object distance. To generate close interactions of two dynamic meshes, we conceptually follow that latter method. The core part of our method is to define a local coordinate system (LCS) for each joint along a mesh's kinematic chain (namely the index order of joints in the mesh).
% we extend the use of the MANO model and and articulated objects to inter-penetrate one another. For ease of generating a more "penetrating" dynamic of meshes, . 

\textbf{LCS for MANO model} is defined for each joint along the kinematic chain of the MANO hand model in its neutral pose. % (Figure \ref{fig:hand_keypoints}). 
More formally, let the 16 joints of the hand plus 5 finger tips $[\textbf{j}^{3D}_0, \hdots,\textbf{j}^{3D}_{20}] = \textbf{J}^{3D} \in \mathbb{R}^{21 \times 3},$
% \begin{equation*}
% [\textbf{j}^{3D}_0, \hdots,\textbf{j}^{3D}_{20}] = \textbf{J}^{3D} \in \mathbb{R}^{21 \times 3},
% \end{equation*}
define a kinematic chain of the hand, where $\textbf{j}^{3D}_0$ is the root joint. The kinematic chain starts at the wrist (root joint) and ends at the finger tips.
For each non-root joint we establish the directions of the LCS as follows. First axis is defined as the direction of a joint to its immediate neighbour along the kinematic chain, $\overrightarrow{\textbf{z}}_i = \textbf{j}^{3D}_{i+1} - \textbf{j}^{3D}_{i}$.
% \begin{equation*}
% \overrightarrow{\textbf{z}}_i = \textbf{j}^{3D}_{i+1} - \textbf{j}^{3D}_{i}.
% \end{equation*}
To  define the second axis $\overrightarrow{\textbf{x}}_i$, we first locate a vertex $\textbf{v}_i$ on the MANO mesh model such that the direction $\hat{\textbf{x}}_i=\textbf{v}_i - \textbf{j}^{3D}_{i}$,
% \begin{equation*}
% \hat{\textbf{x}}_i=\textbf{v}_i - \textbf{j}^{3D}_{i}
% \end{equation*}
approximately follows the flexion direction of the joint. 
$\overrightarrow{\textbf{x}}_i$ is then the projection of $\hat{\textbf{x}}_i$ onto the perpendicular plane to $\overrightarrow{\textbf{z}}_i$.
 %The projection of point $\textbf{v}_i$ onto $\overrightarrow{\textbf{x}}_i$ defines the positive direction of $\overrightarrow{\textbf{x}}_i$ with respect to $\textbf{j}^{3D}_{i}$. 
% $\overrightarrow{\textbf{x}}_i$ is then estimated as the vector that minimizes the projection distance of $\hat{\textbf{x}}_i$ onto an orthogonal unit vector to $\overrightarrow{\textbf{z}}_i$.  
% The projection of point $\textbf{v}_i$ onto $\overrightarrow{\textbf{x}}_i$ defines the positive direction of $\overrightarrow{\textbf{x}}_i$ with respect to $\textbf{j}^{3D}_{i}$. 
Finally, the remaining axis is computed as $\overrightarrow{\textbf{y}}_i = \overrightarrow{\textbf{z}}_i \times \overrightarrow{\textbf{x}}_i$,
% \begin{equation*}
% \overrightarrow{\textbf{y}}_i = \overrightarrow{\textbf{z}}_i \times \overrightarrow{\textbf{x}}_i
% \end{equation*}
where $\times$ is the cross product. We normalize all axes vectors to a unit size.

\textbf{LCS for dynamic objects}. A similar process as performed on the MANO hand model to establish LCS for joints is done for dynamic mesh models, such as the chain model shown in Figure \ref{fig:cyl_present}e-f.

\subsection{Generating Mesh Interactions}
\label{sec:generating_mesh_interactions}
With the mesh models defined in Sec.~\ref{sec:mesh_definition}, we simulate interactions between two hand meshes, or a hand mesh and an object mesh using the following algorithm: 
\begin{enumerate}
  \item Place object or a hand at the center of the scene at a random pose and rotation.
  \item Insert an additional hand model to the scene at a distance from the scene’s center such that the objects do not intersect. The added hand model is at a neutral pose such that the front of the hand is facing the object located at scene’s center.
  \item Incrementally shift the hand model in (2) towards the scene's center until a collision occurs between it and the object/hand at scene's center.
  \item Randomly move each joint of the dynamic models in the scene until collision with another object/hand occurs or maximum angle of rotation is reached for each joint. In our case, all angles of rotation for the hand models are limited to 60 degrees. 
  \item Repeat 4 till the dynamic models can no longer perform a valid move. 
  \item Render final scene
\end{enumerate}
Figure \ref{fig:cyl_present} show an example of the workflow. The validity of hands' location in 3D space is performed based on mesh intersection. Specifically, for fast collision check, we approximate the hand and static objects as a collection of cylinders and annuli. Then, we check whether a cylinder of one hand intersects a cylinder of another hand/object. To this end, we have implemented the algorithm of \cite{Ketchel2005} that determines whether two cylinders intersect in a given configuration. We check in a similar manner the possible intersection of fingers located on the same hand given a hand pose.   

\subsection{Additional qualitative results}
\begin{figure*}
\begin{center}
		\tabcolsep 0.2cm
 		\renewcommand\arraystretch{0.1}
		%\resizebox{\textwidth}{!}{
		\begin{tabular}{cc}			
			%			\multicolumn{2}{c}{A}  & B \\

			\includegraphics[trim = 00mm 30mm 0mm 30mm, clip, height=3.0cm]{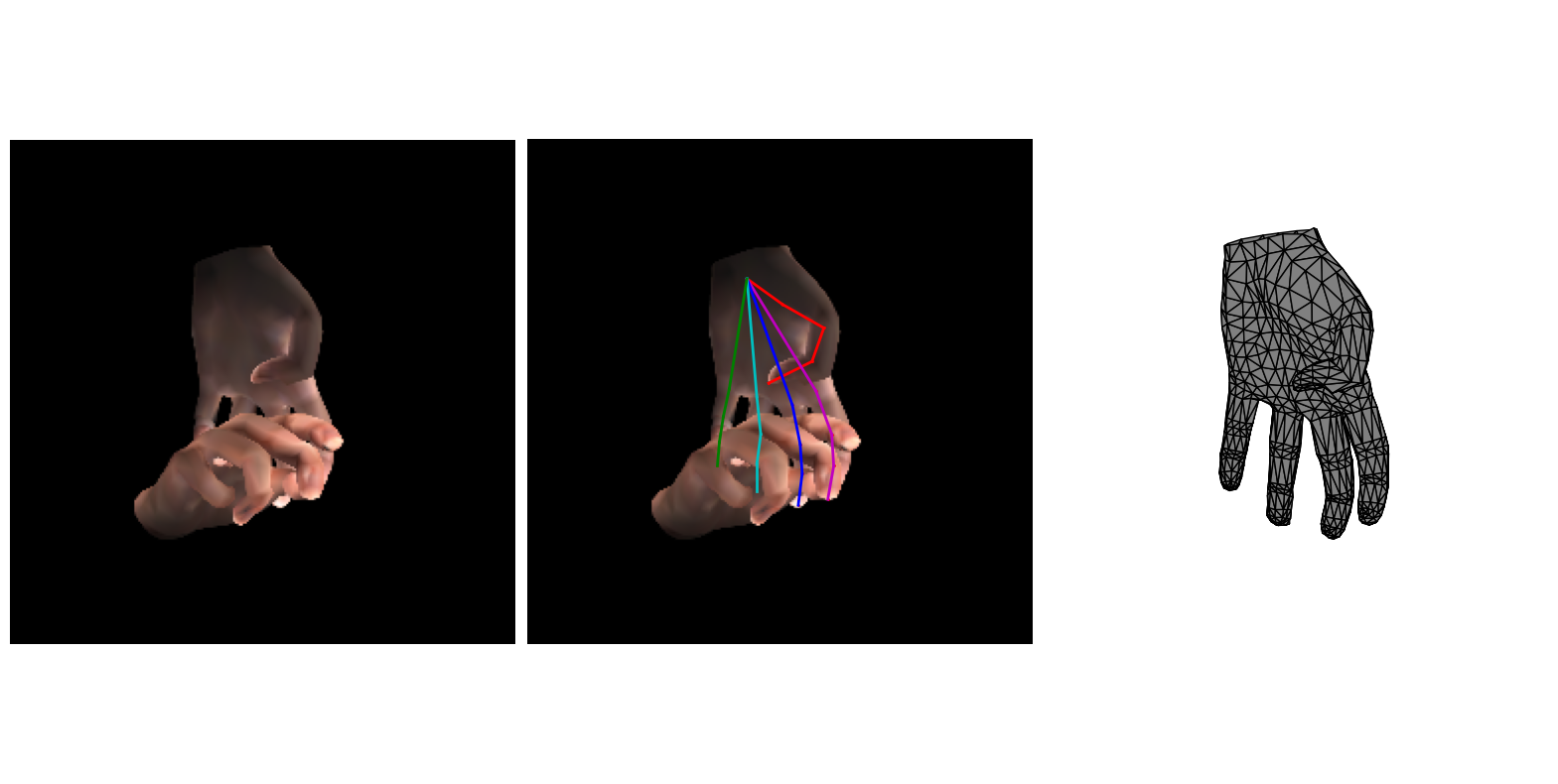}&
			\includegraphics[trim = 00mm 30mm 0mm 30mm, clip, height=3.0cm]{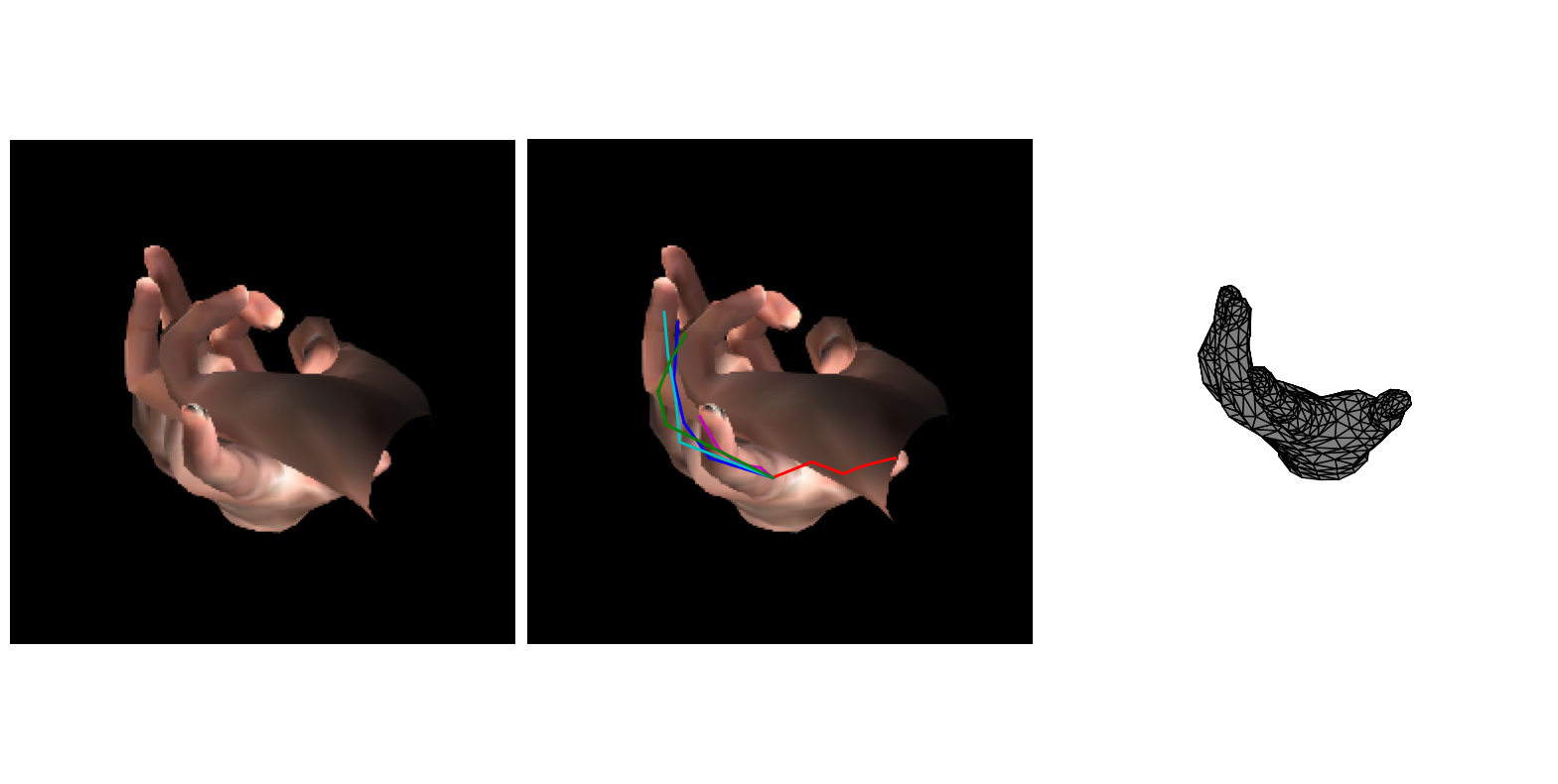} \\
			\includegraphics[trim = 00mm 30mm 0mm 30mm, clip, height=3.0cm]{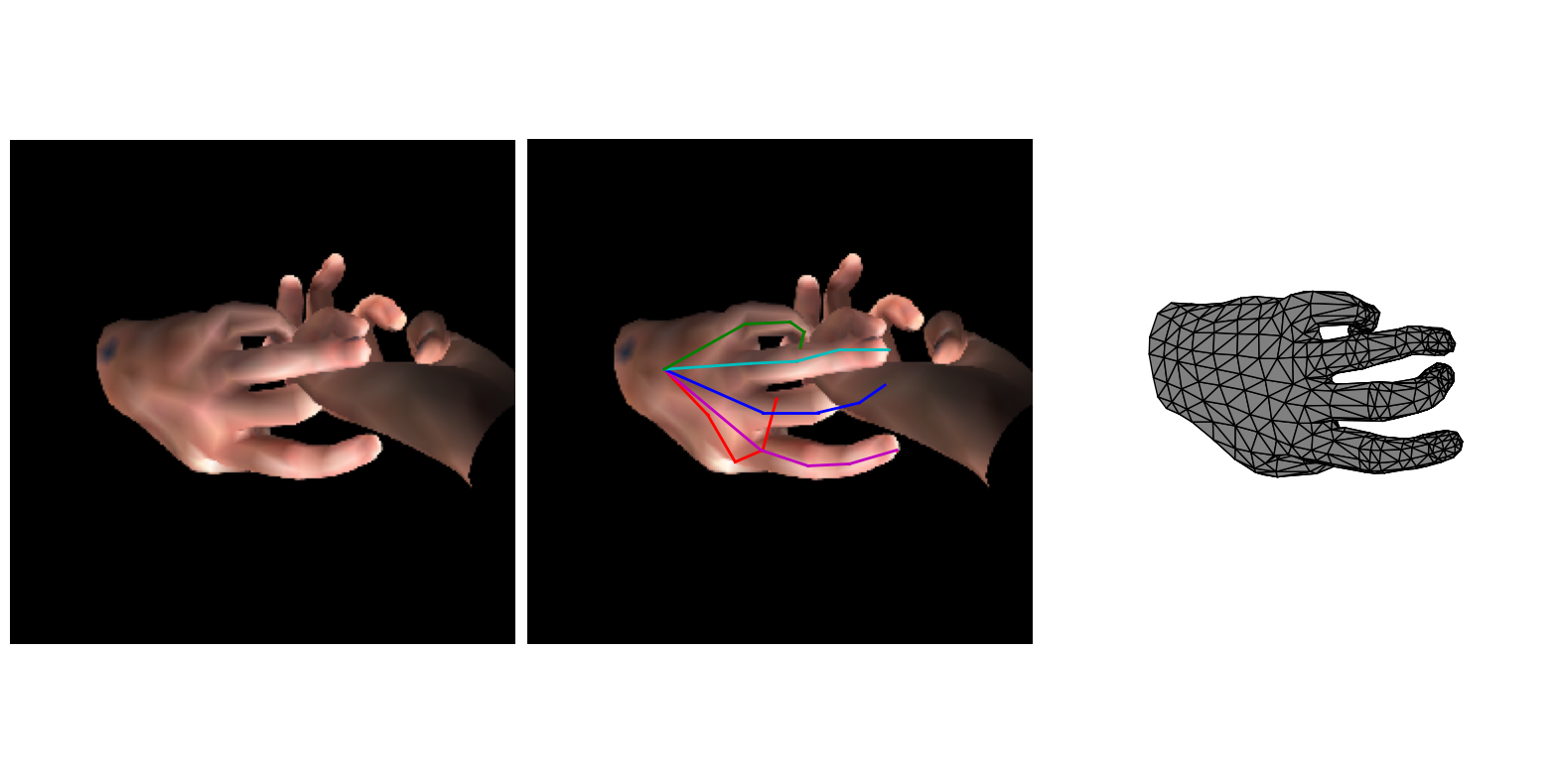} &
			\includegraphics[trim = 00mm 30mm 0mm 30mm, clip, height=3.0cm]{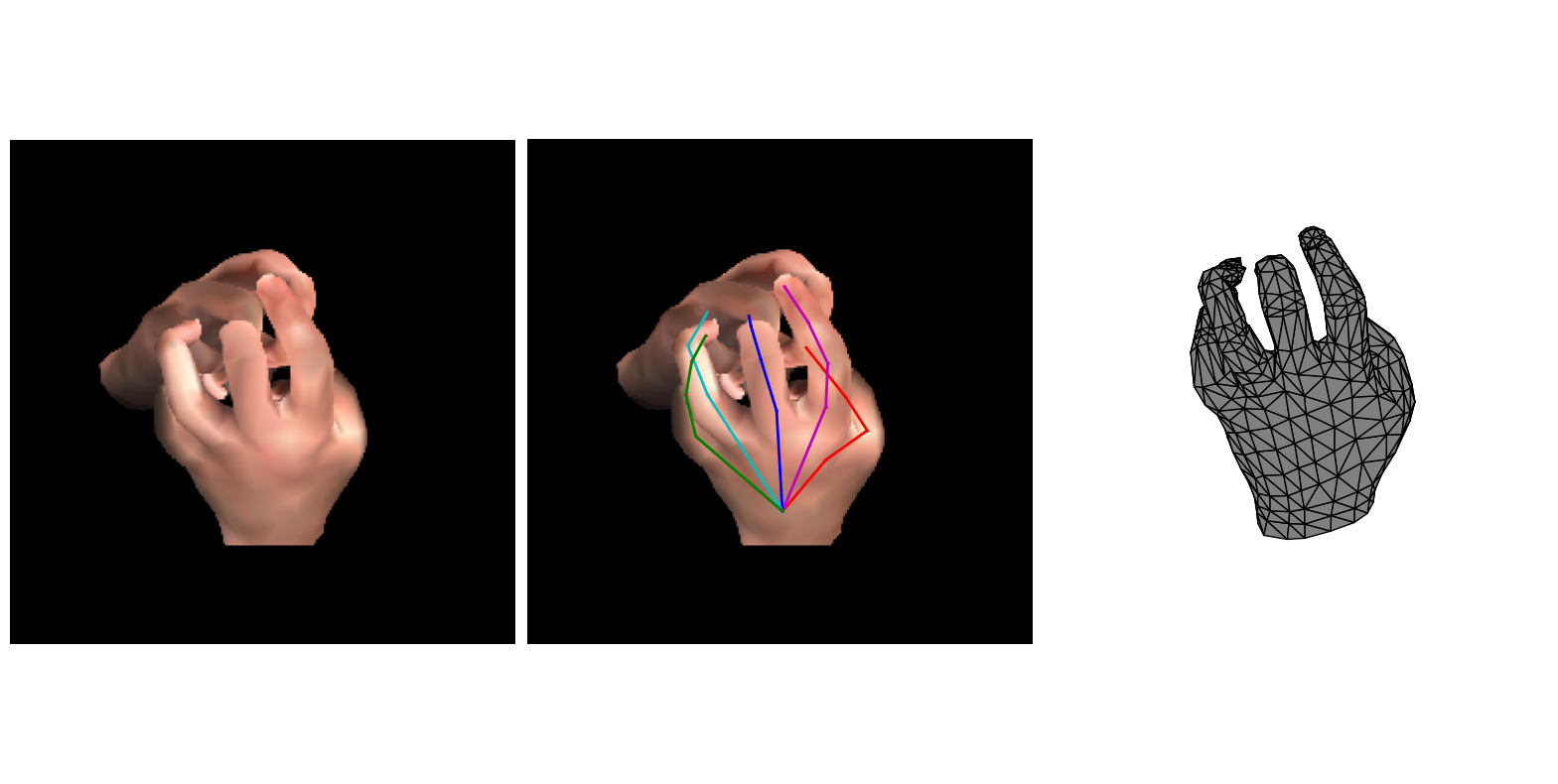} \\ 
			\includegraphics[trim = 0mm 30mm 00mm 30mm, clip, height=3.0cm]{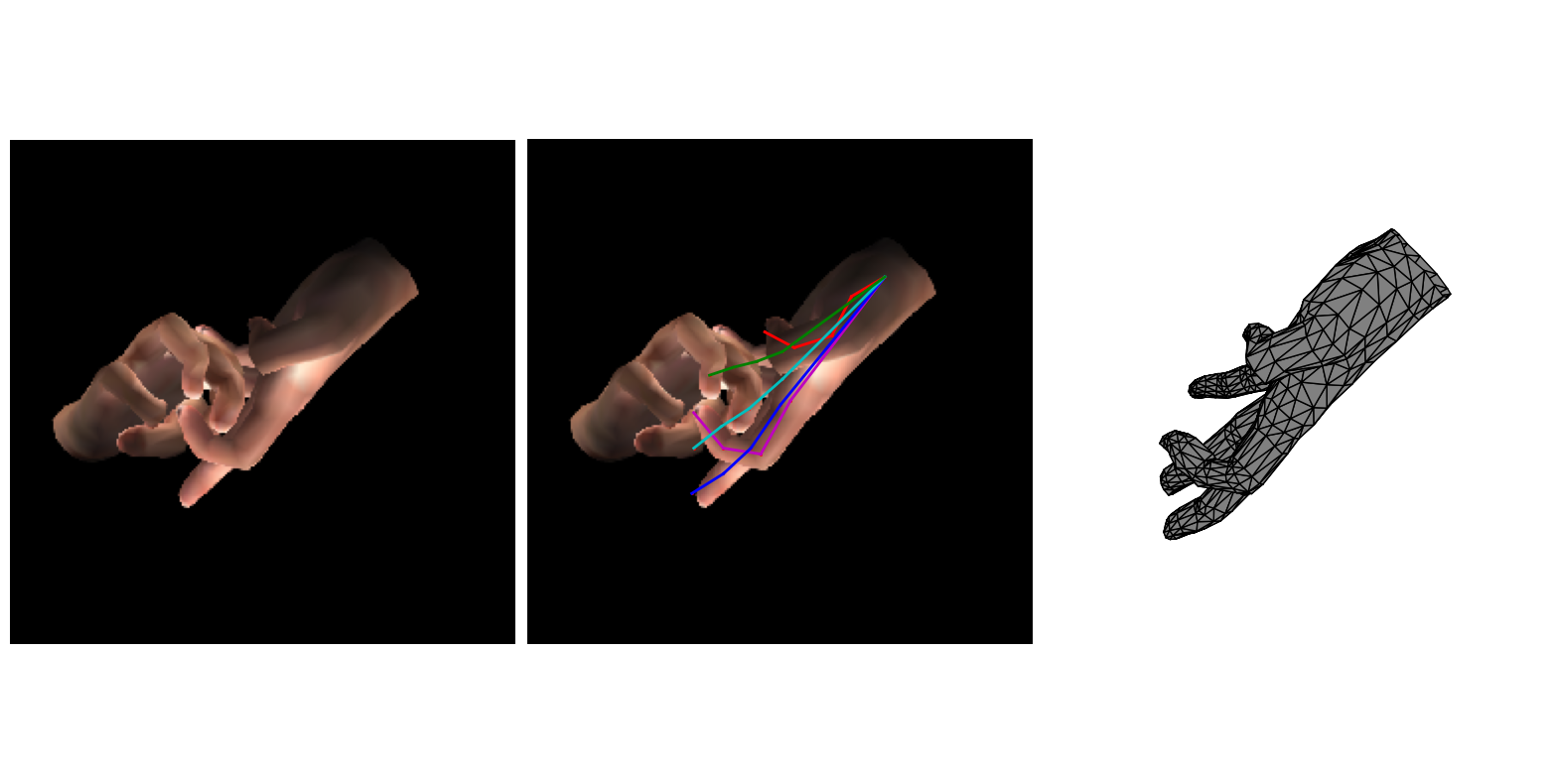}&
			\includegraphics[trim = 0mm 30mm 0mm 30mm, clip, height=3.0cm]{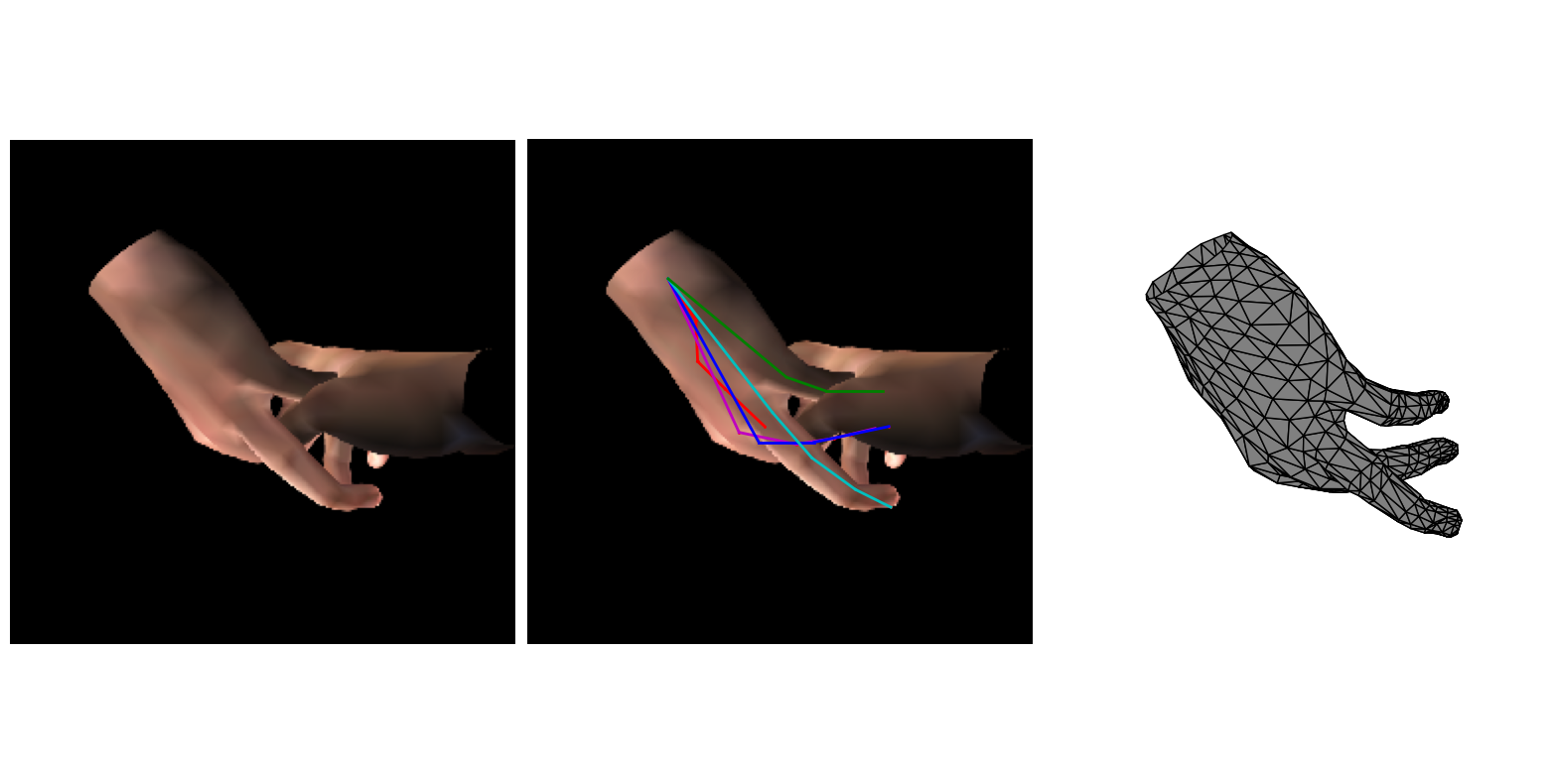} \\
			\includegraphics[trim = 0mm 30mm 0mm 30mm, clip, height=3.0cm]{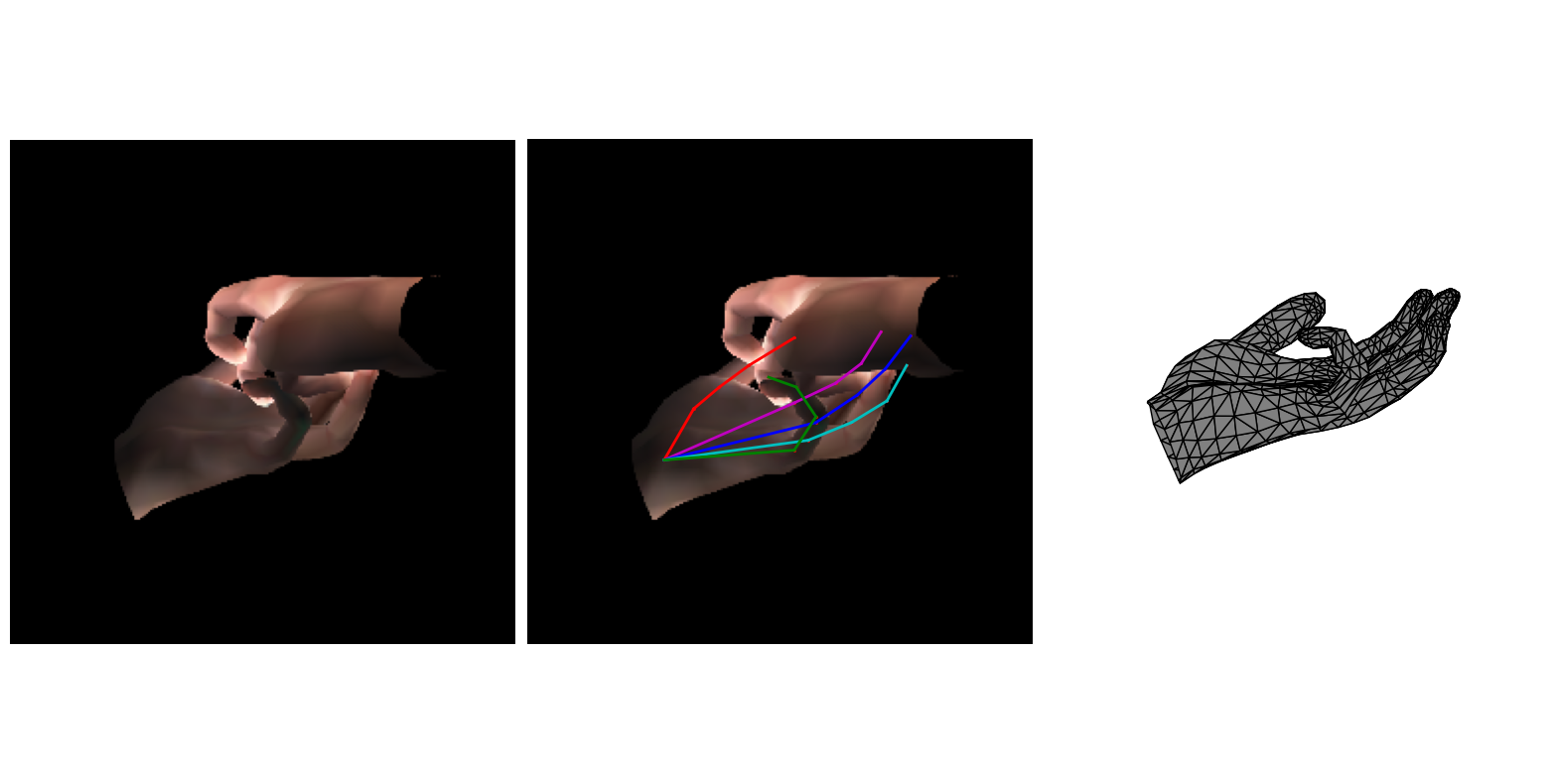} &
			\includegraphics[trim = 0mm 30mm 0mm 30mm, clip, height=3.0cm]{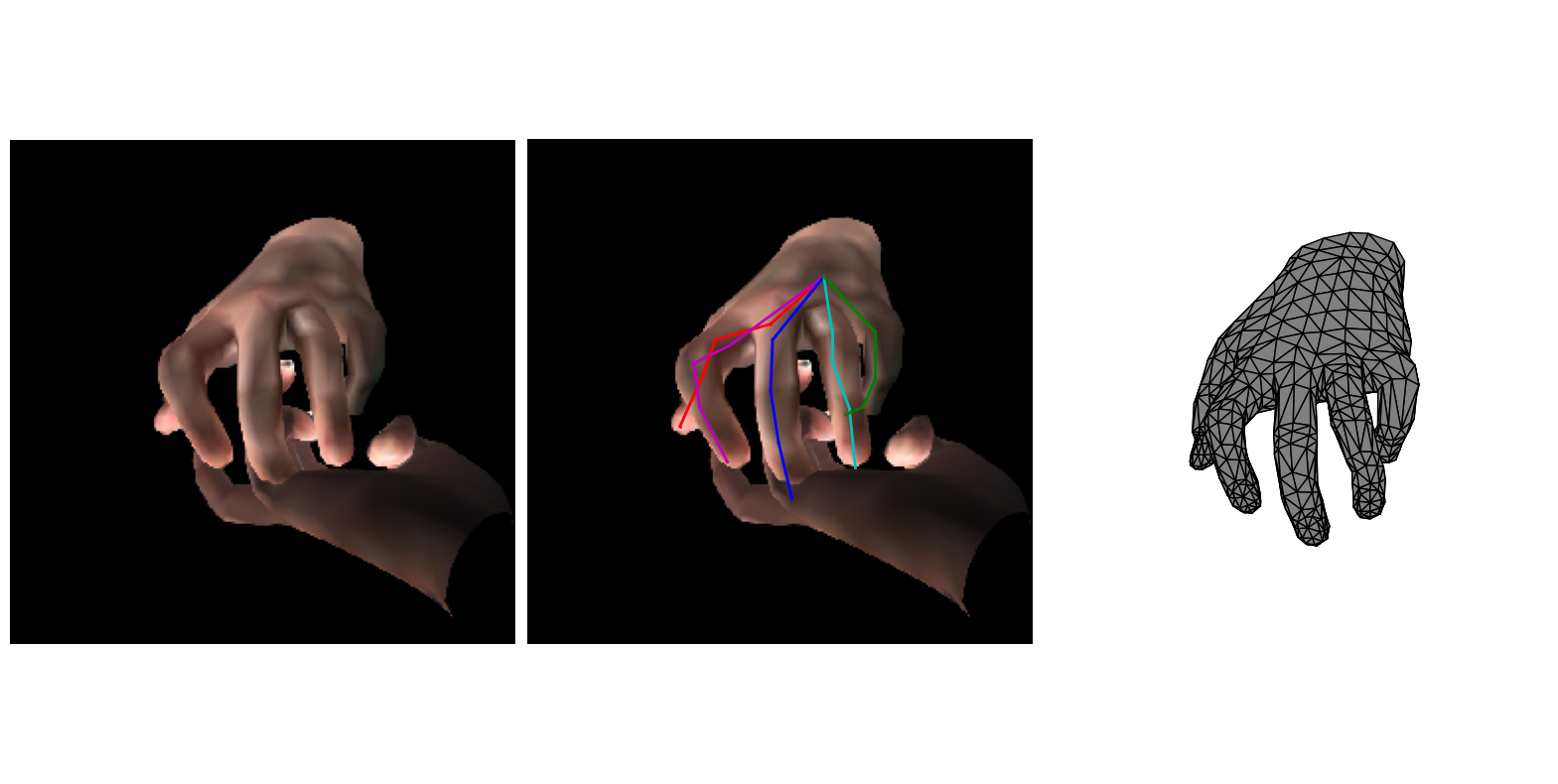} \\ 
			\includegraphics[trim = 0mm 30mm 00mm 30mm, clip, height=3.0cm]{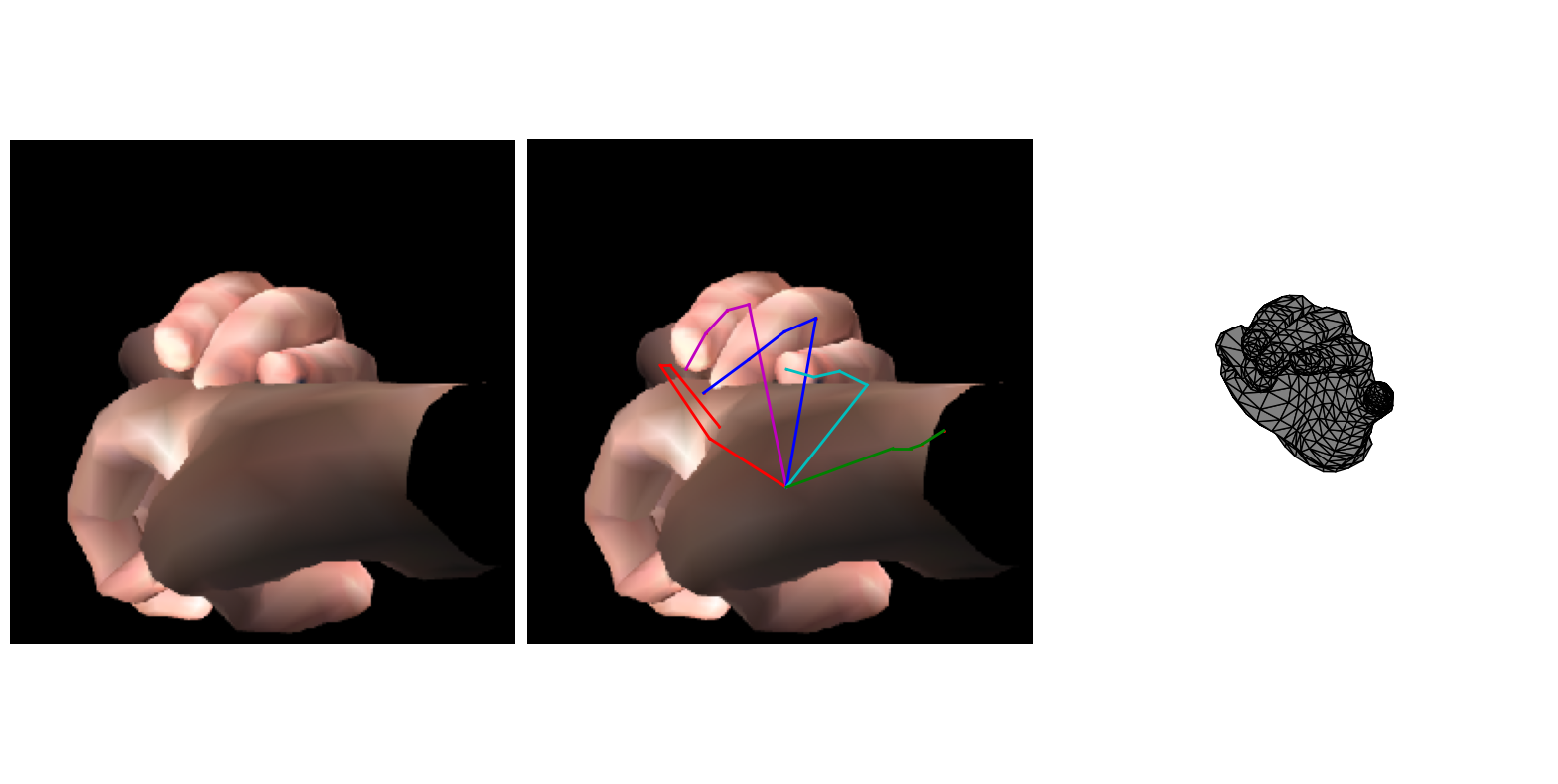}&
			\includegraphics[trim = 0mm 30mm 0mm 30mm, clip, height=3.0cm]{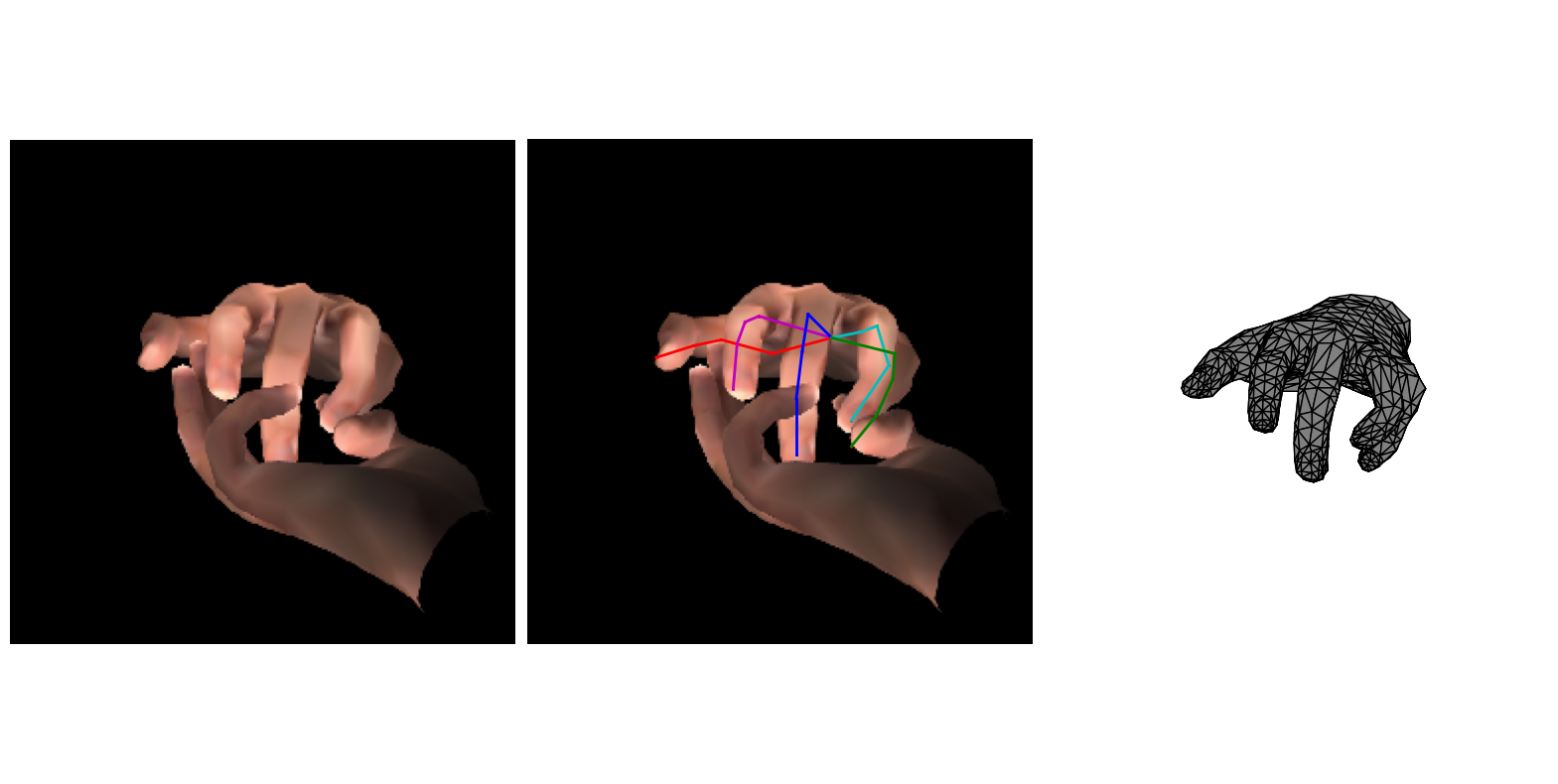} \\
			\includegraphics[trim = 0mm 30mm 0mm 30mm, clip, height=3.0cm]{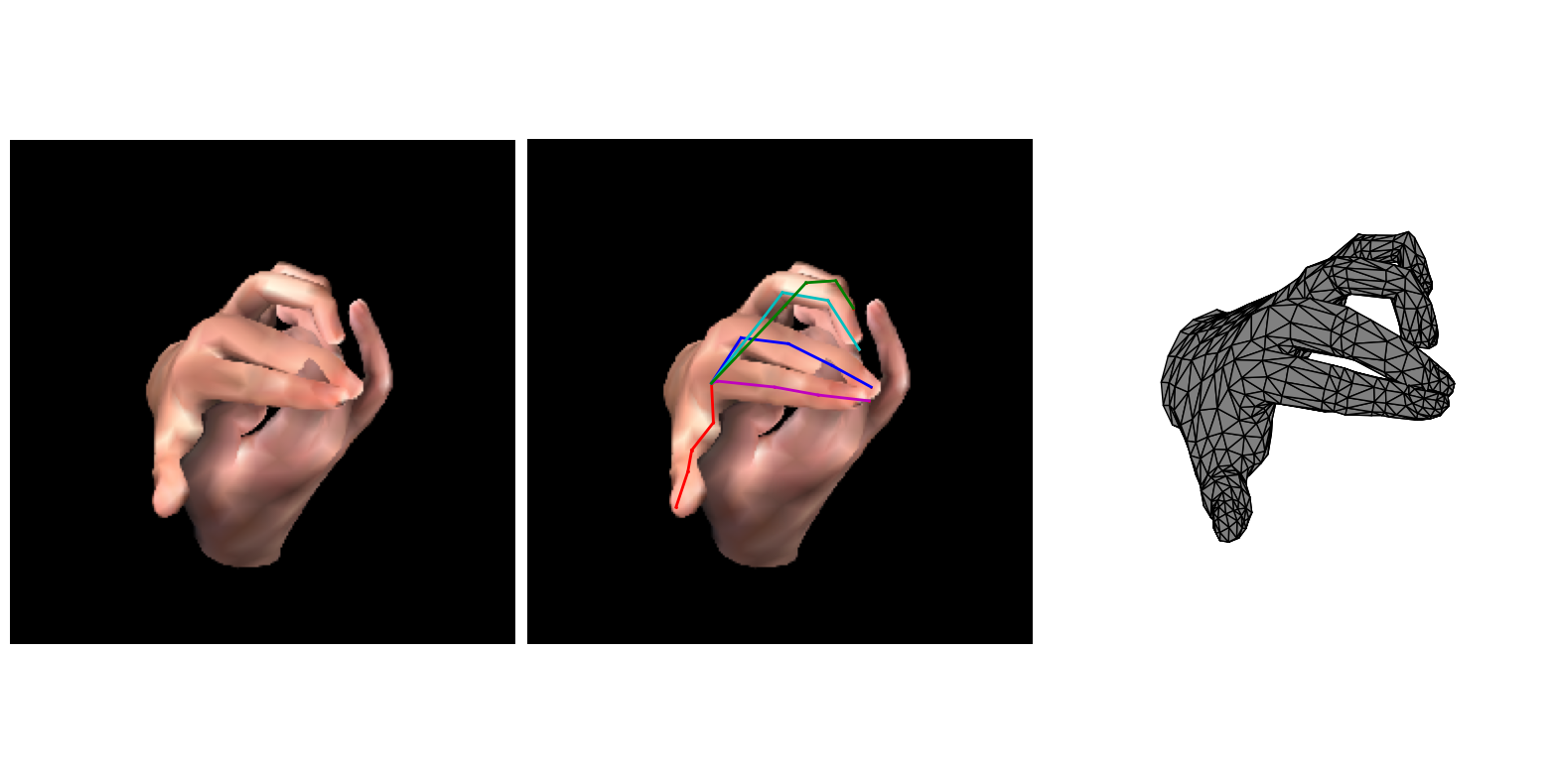} &
			\includegraphics[trim = 0mm 30mm 0mm 30mm, clip, height=3.0cm]{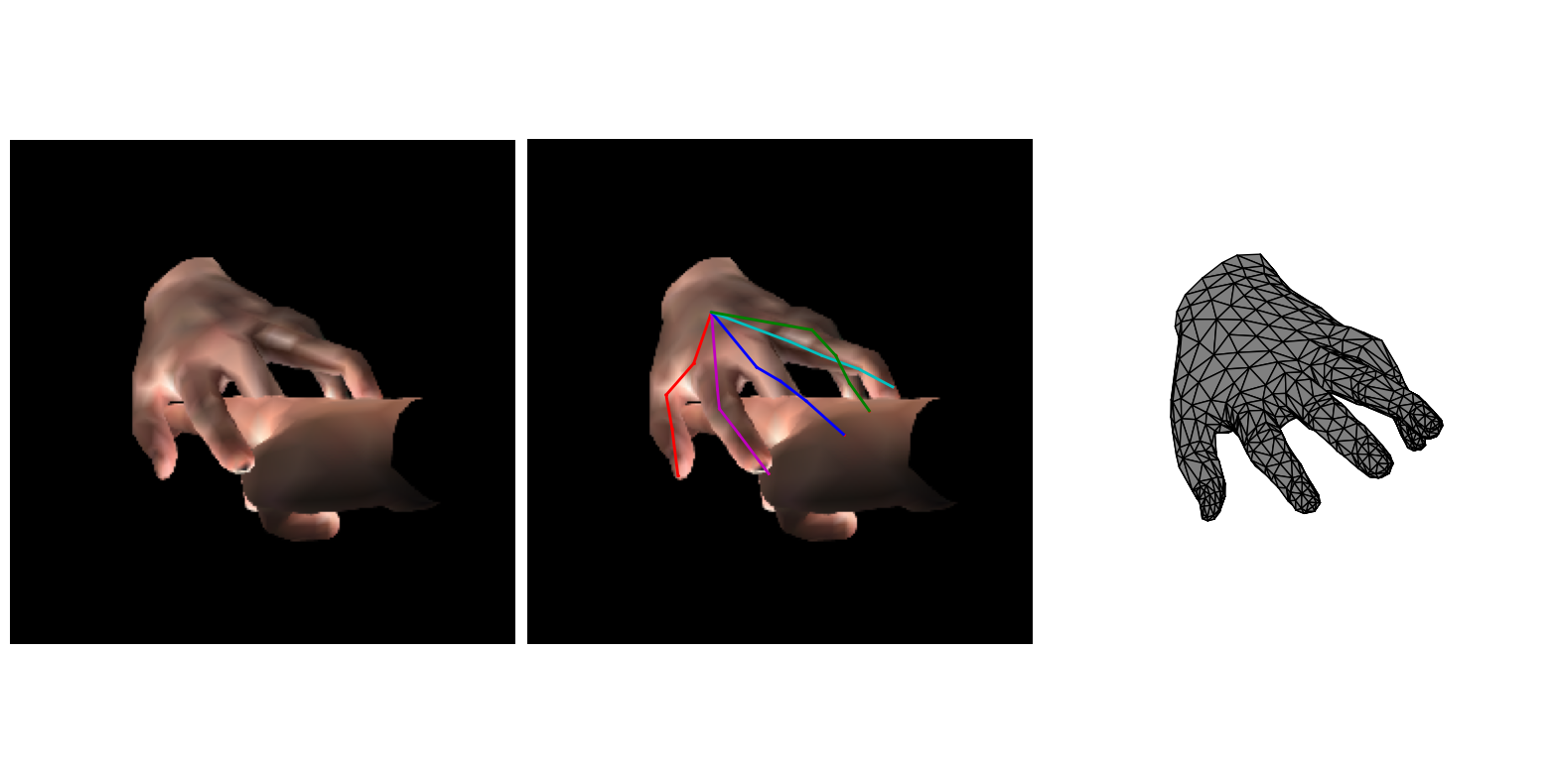} \\ 
		\end{tabular}
		%}
	\end{center}
	\vspace{-0.2cm} 
	\caption{
		Mesh reconstruction results for our proposed model on the INTER-SH dataset, Hand-Hand interaction images. Right column in the original input, central column shows projected 2D keypoints of the estimated 3D model on top of the original input frame, and the third column in the predicted 3D mesh.
	}
	\label{fig:inter_model_results_hand_hand}
\end{figure*}

\begin{figure*}
\begin{center}
		\tabcolsep 0.2cm
 		\renewcommand\arraystretch{0.1}
		%\resizebox{\textwidth}{!}{
		\begin{tabular}{cc}			
			%			\multicolumn{2}{c}{A}  & B \\

			\includegraphics[trim = 0mm 30mm 00mm 30mm, clip, height=3.0cm]{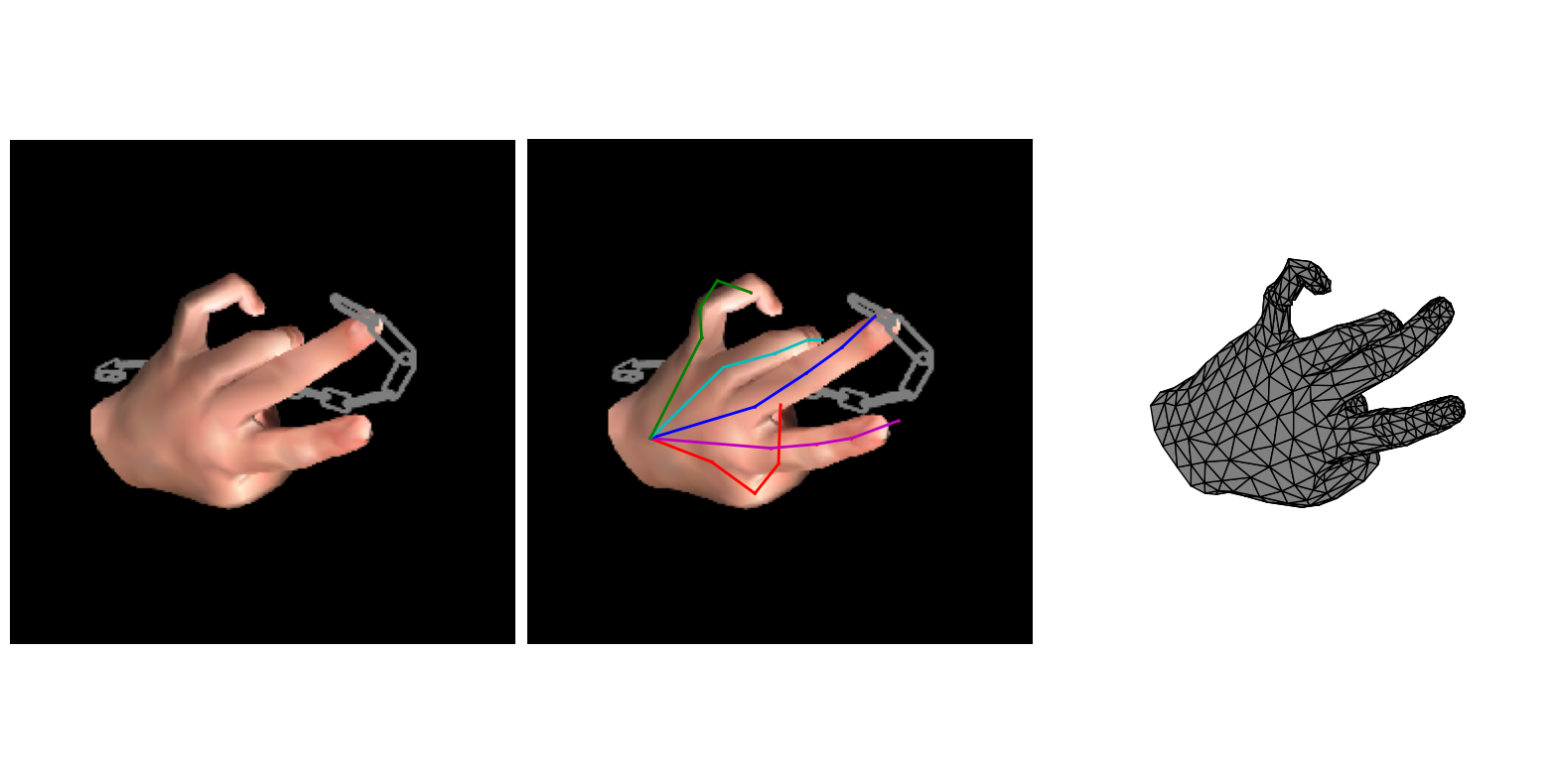}&
			\includegraphics[trim = 0mm 30mm 0mm 30mm, clip, height=3.0cm]{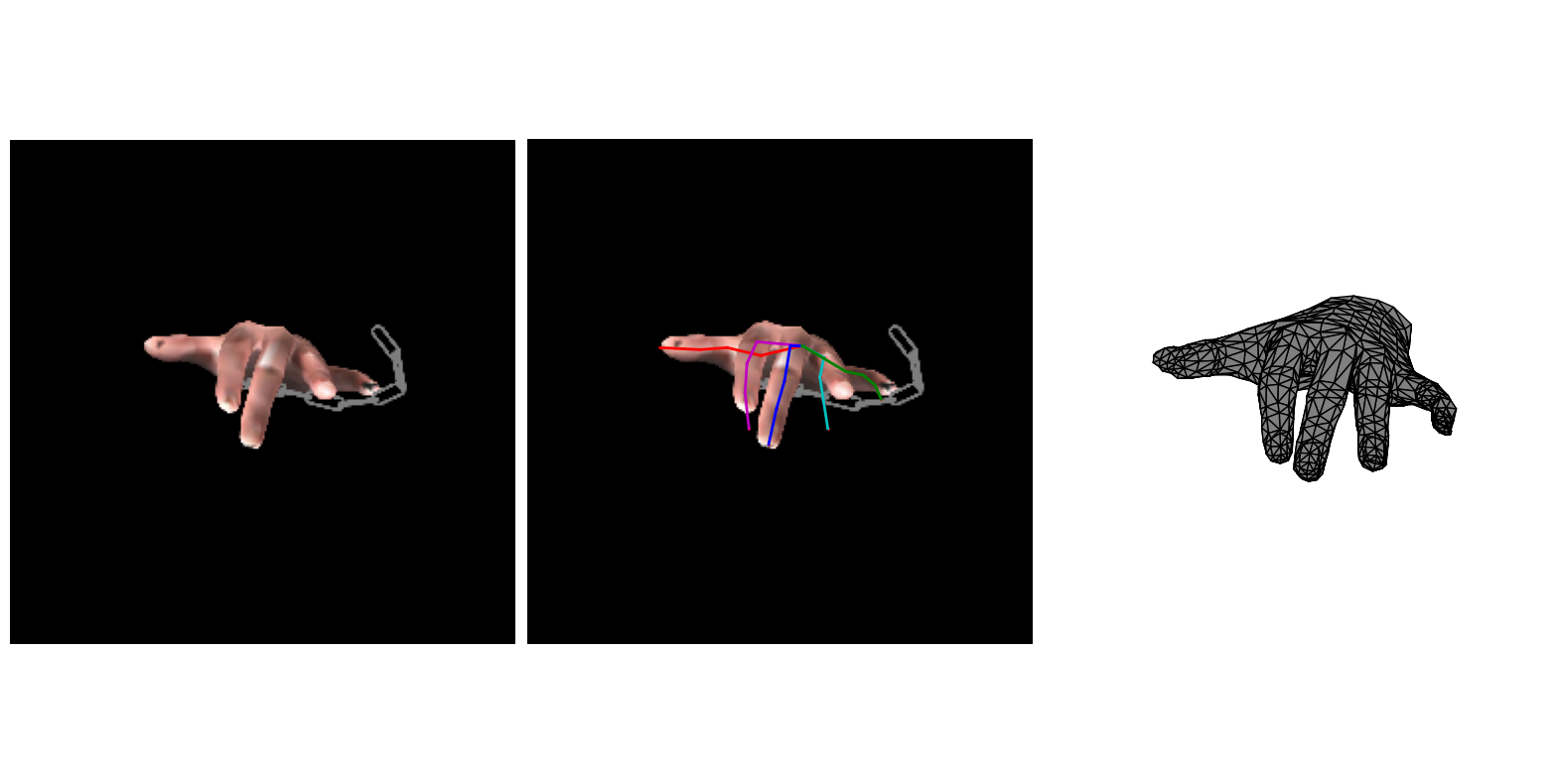} \\
			\includegraphics[trim = 0mm 30mm 0mm 30mm, clip, height=3.0cm]{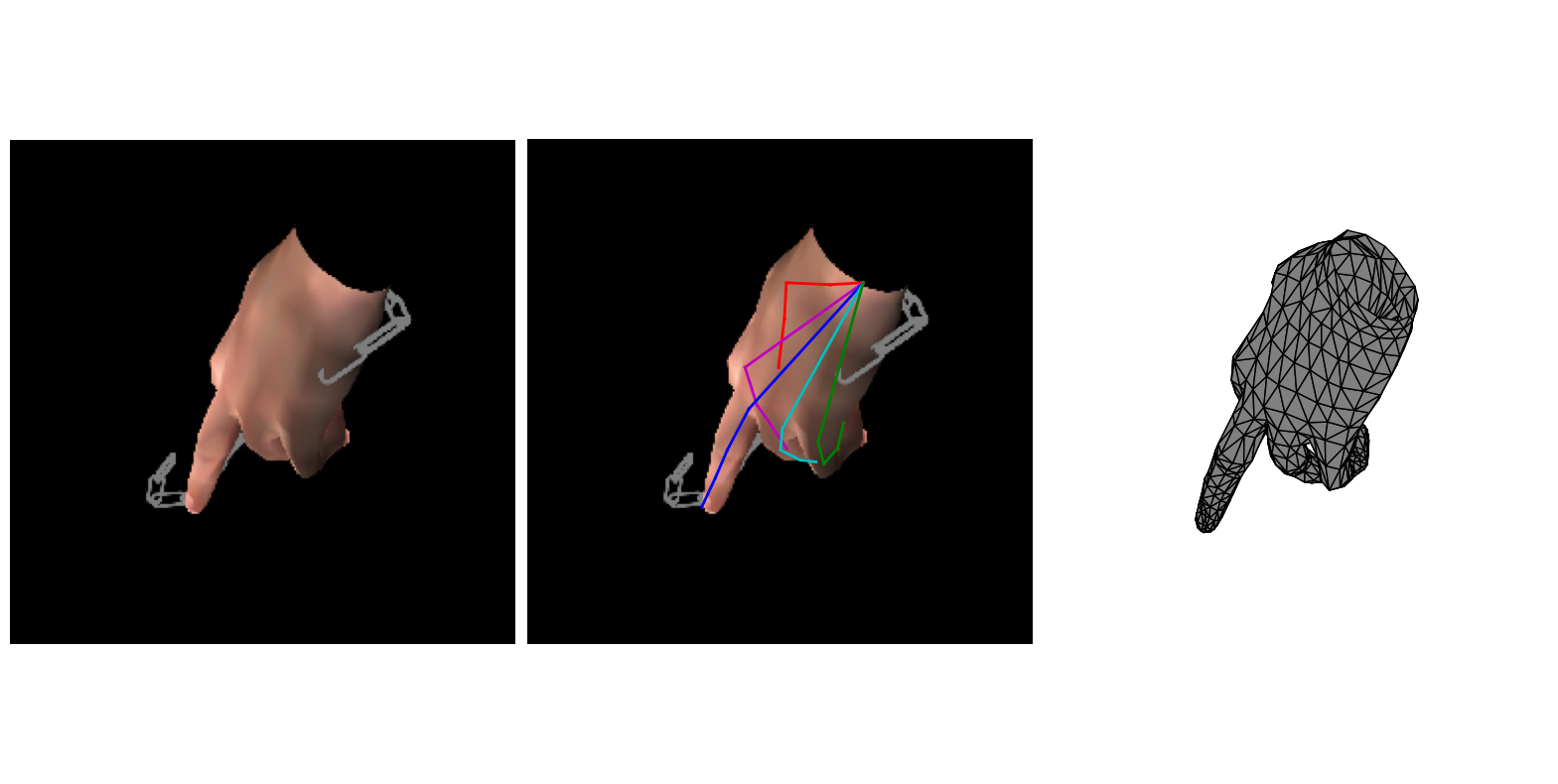} &
			\includegraphics[trim = 0mm 30mm 0mm 30mm, clip, height=3.0cm]{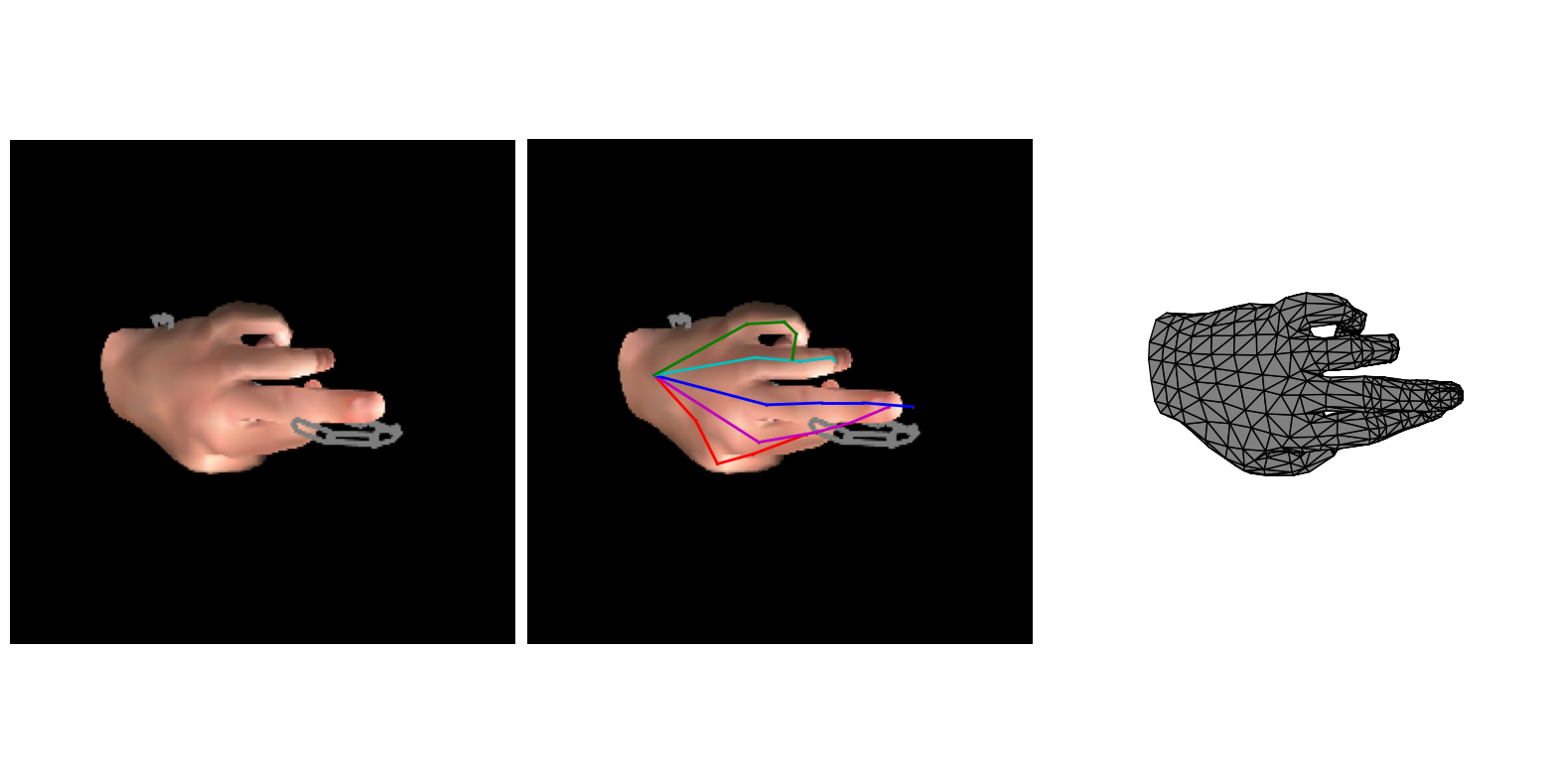} \\ 
			\includegraphics[trim = 0mm 30mm 00mm 30mm, clip, height=3.0cm]{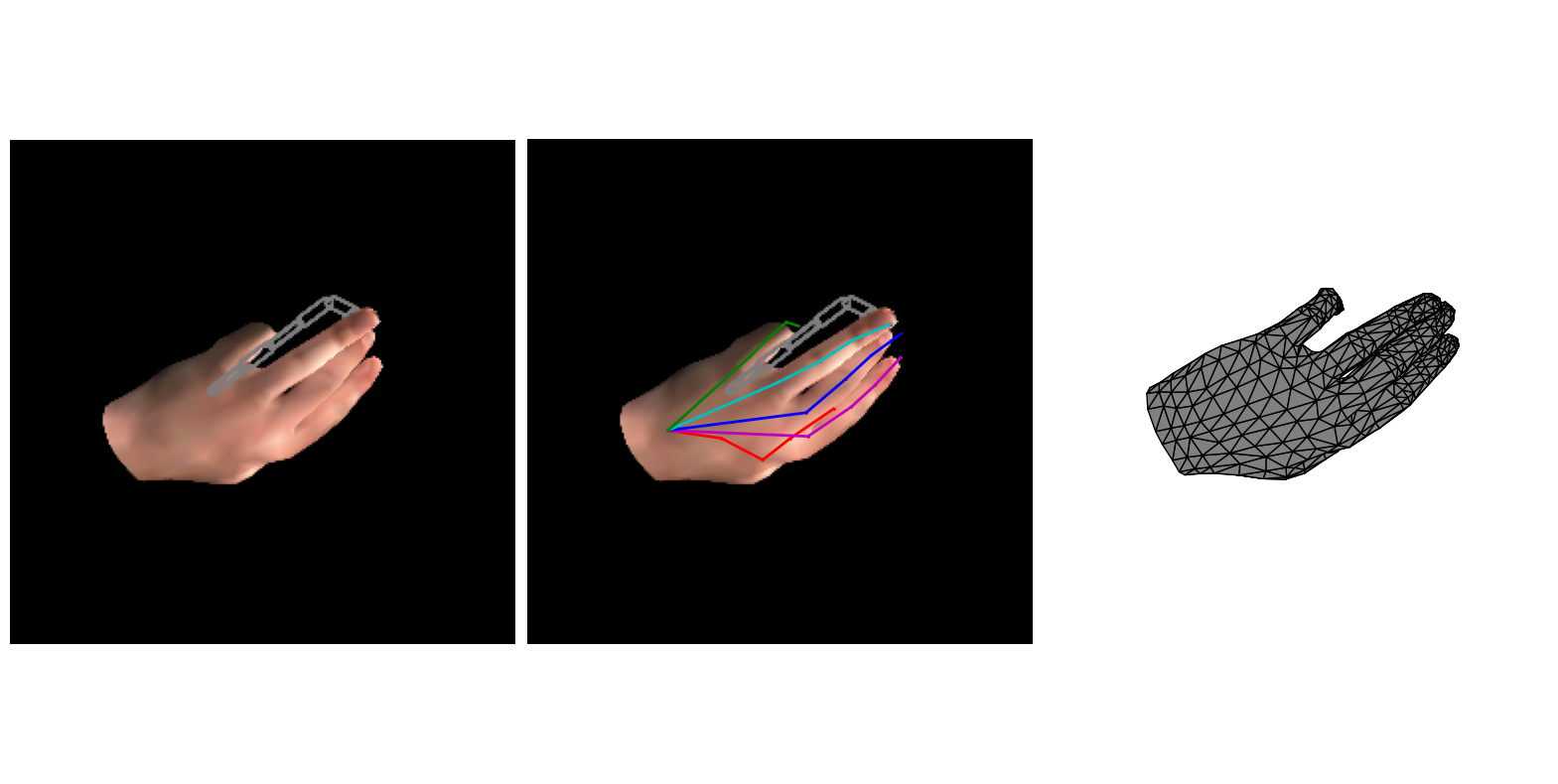}&
			\includegraphics[trim = 0mm 30mm 0mm 30mm, clip, height=3.0cm]{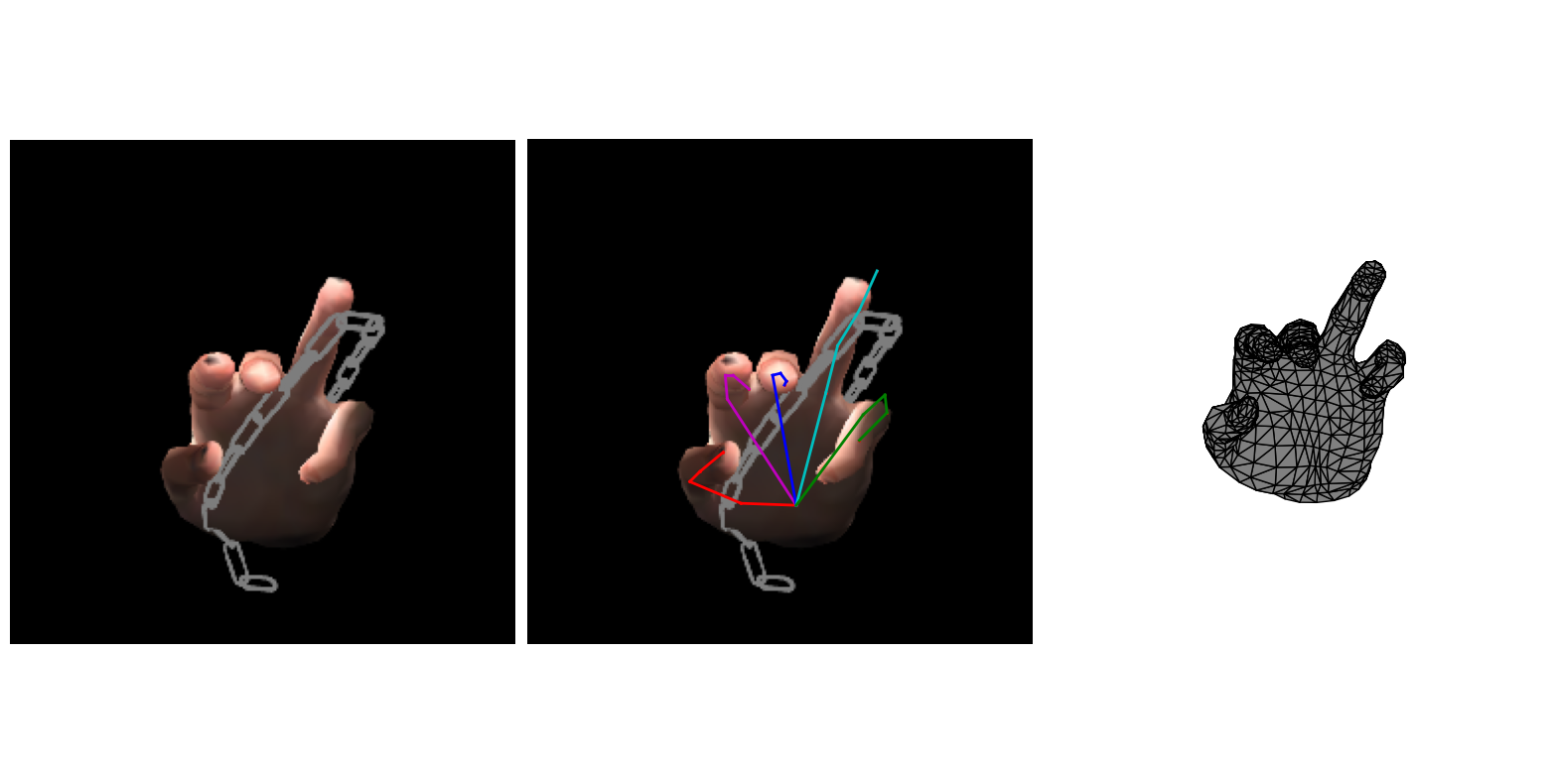} \\
			\includegraphics[trim = 0mm 30mm 0mm 30mm, clip, height=3.0cm]{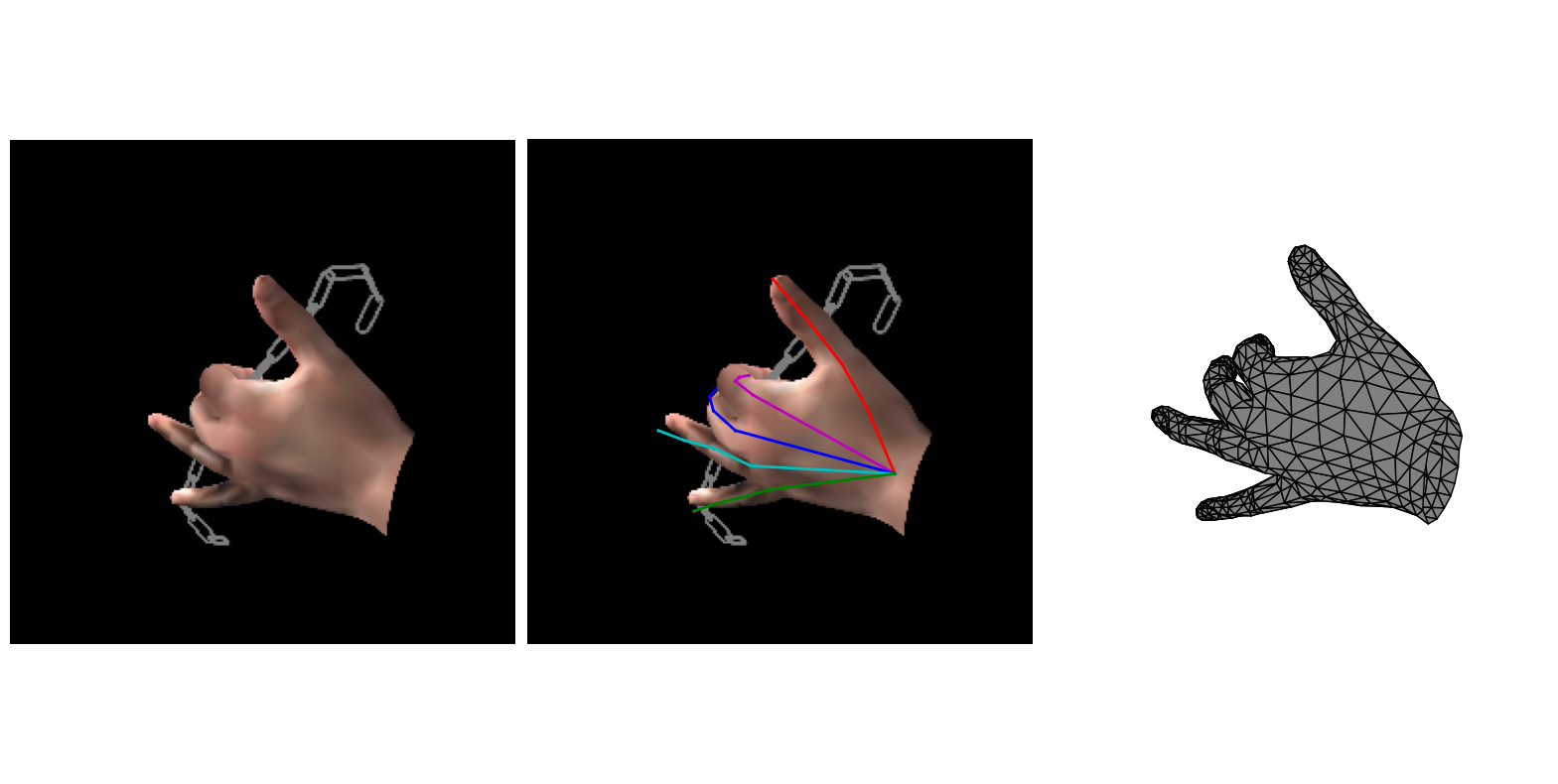} &
			\includegraphics[trim = 0mm 30mm 0mm 30mm, clip, height=3.0cm]{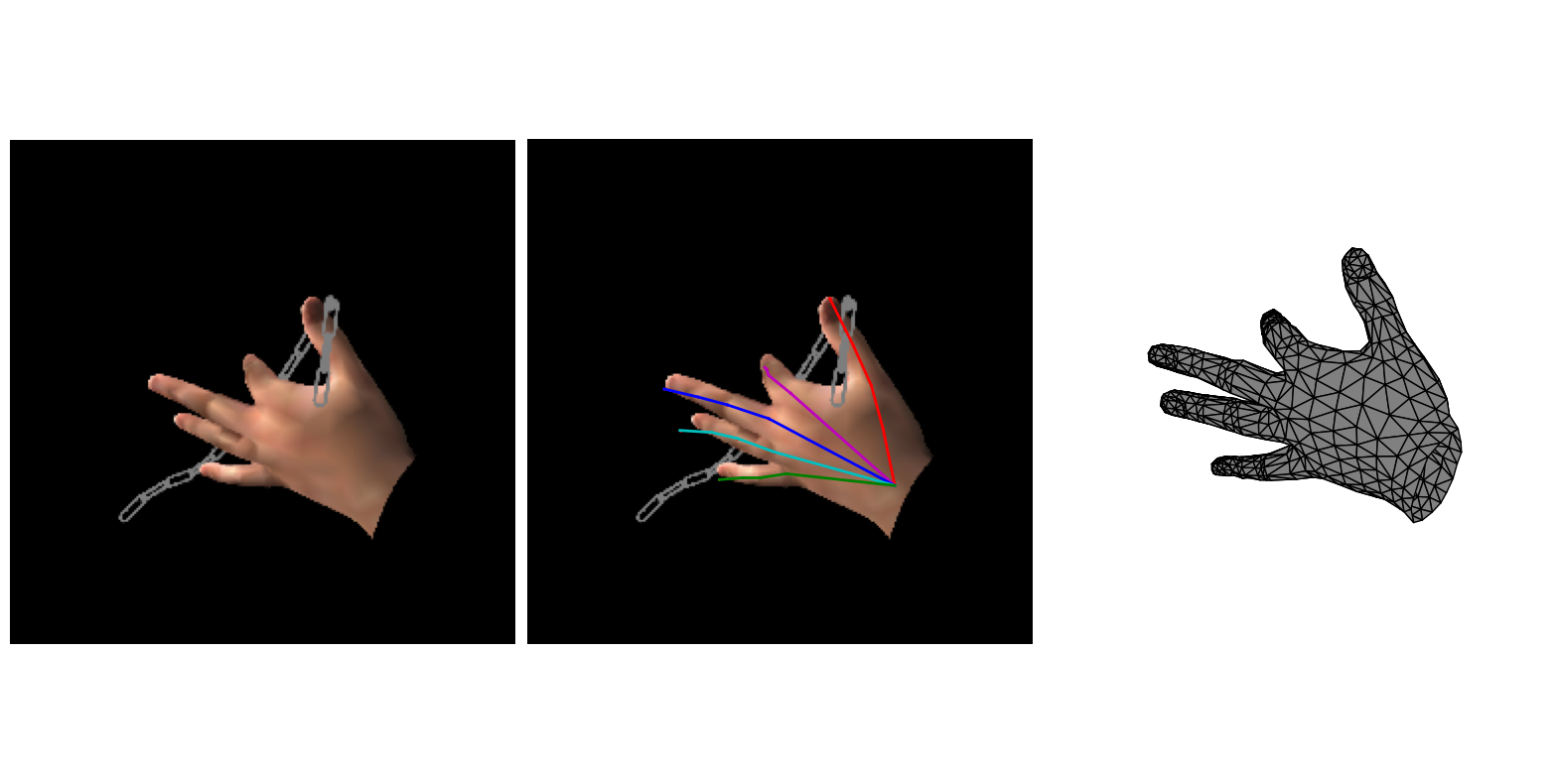} \\ 
			\includegraphics[trim = 0mm 30mm 00mm 30mm, clip, height=3.0cm]{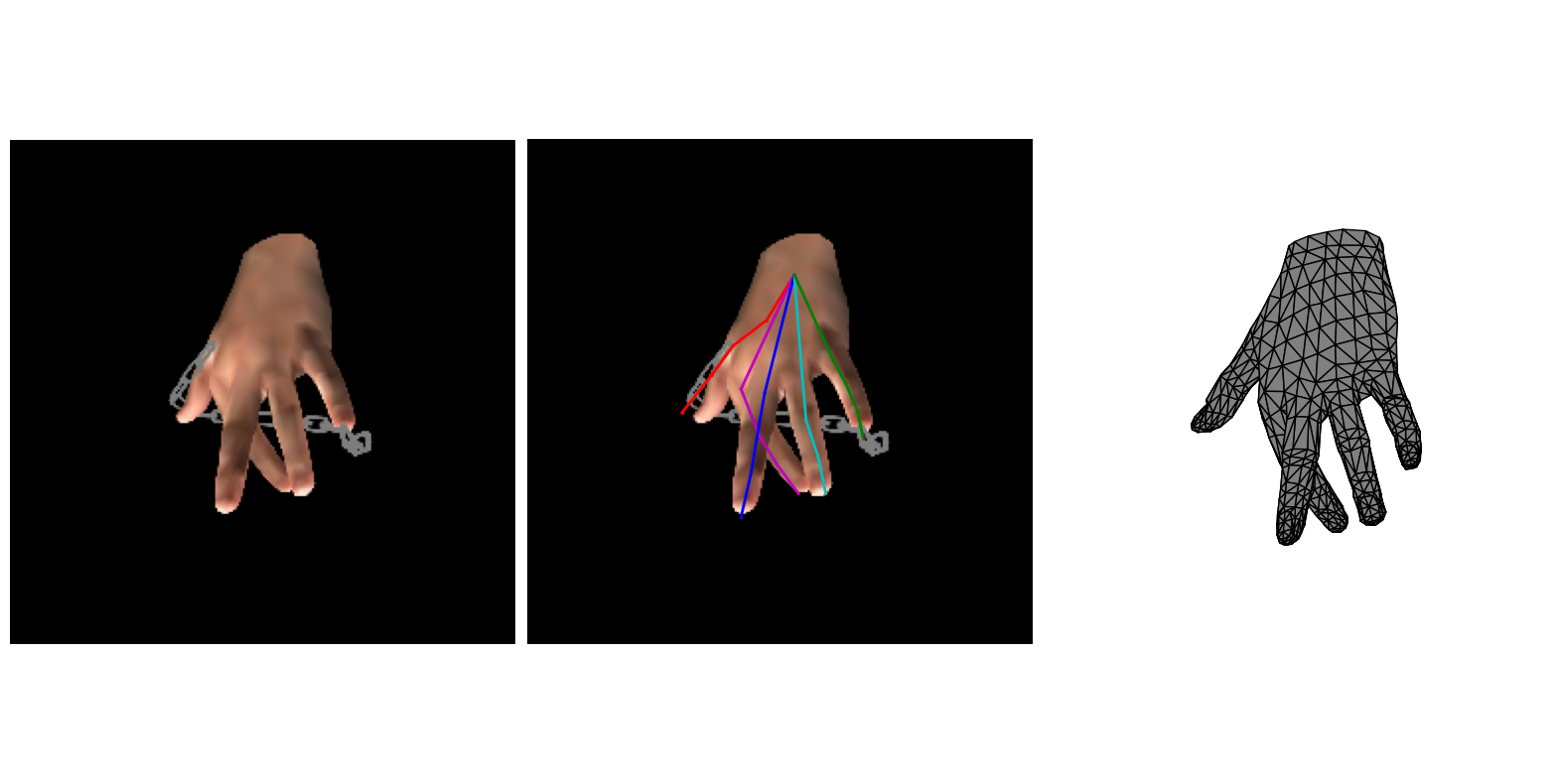}&
			\includegraphics[trim = 0mm 30mm 0mm 30mm, clip, height=3.0cm]{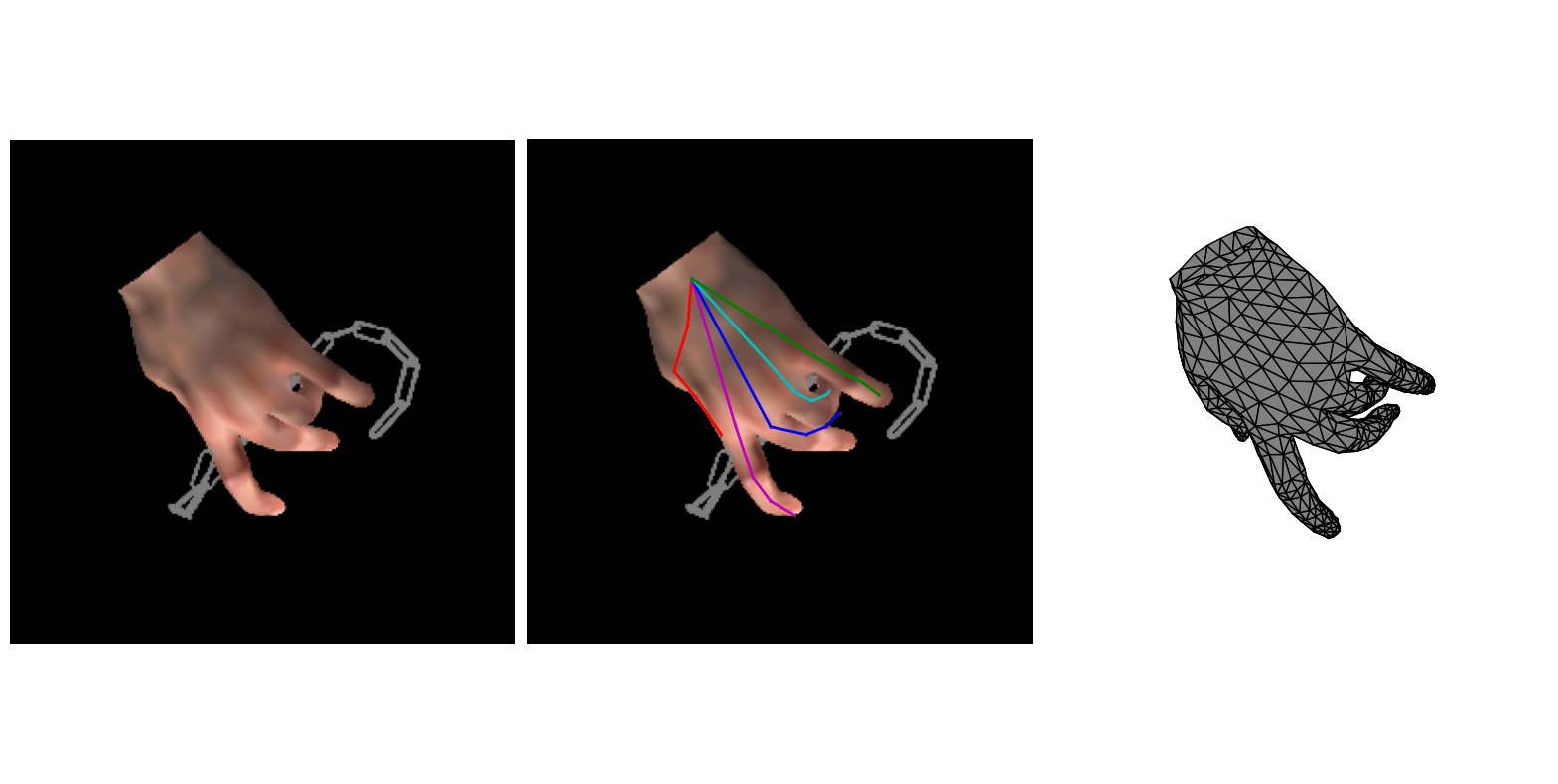} \\
			\includegraphics[trim = 0mm 30mm 0mm 30mm, clip, height=3.0cm]{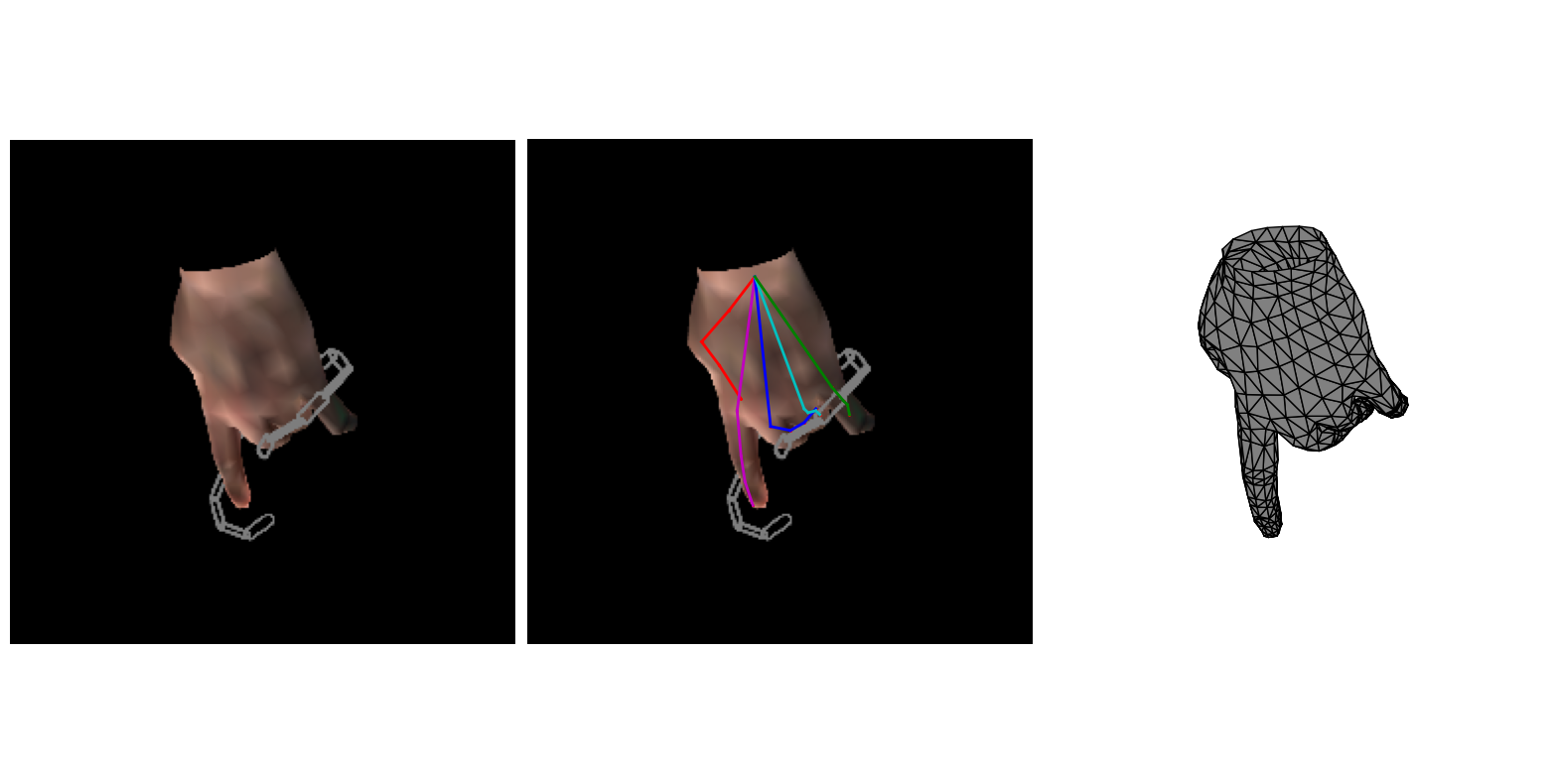} &
			\includegraphics[trim = 0mm 30mm 0mm 30mm, clip, height=3.0cm]{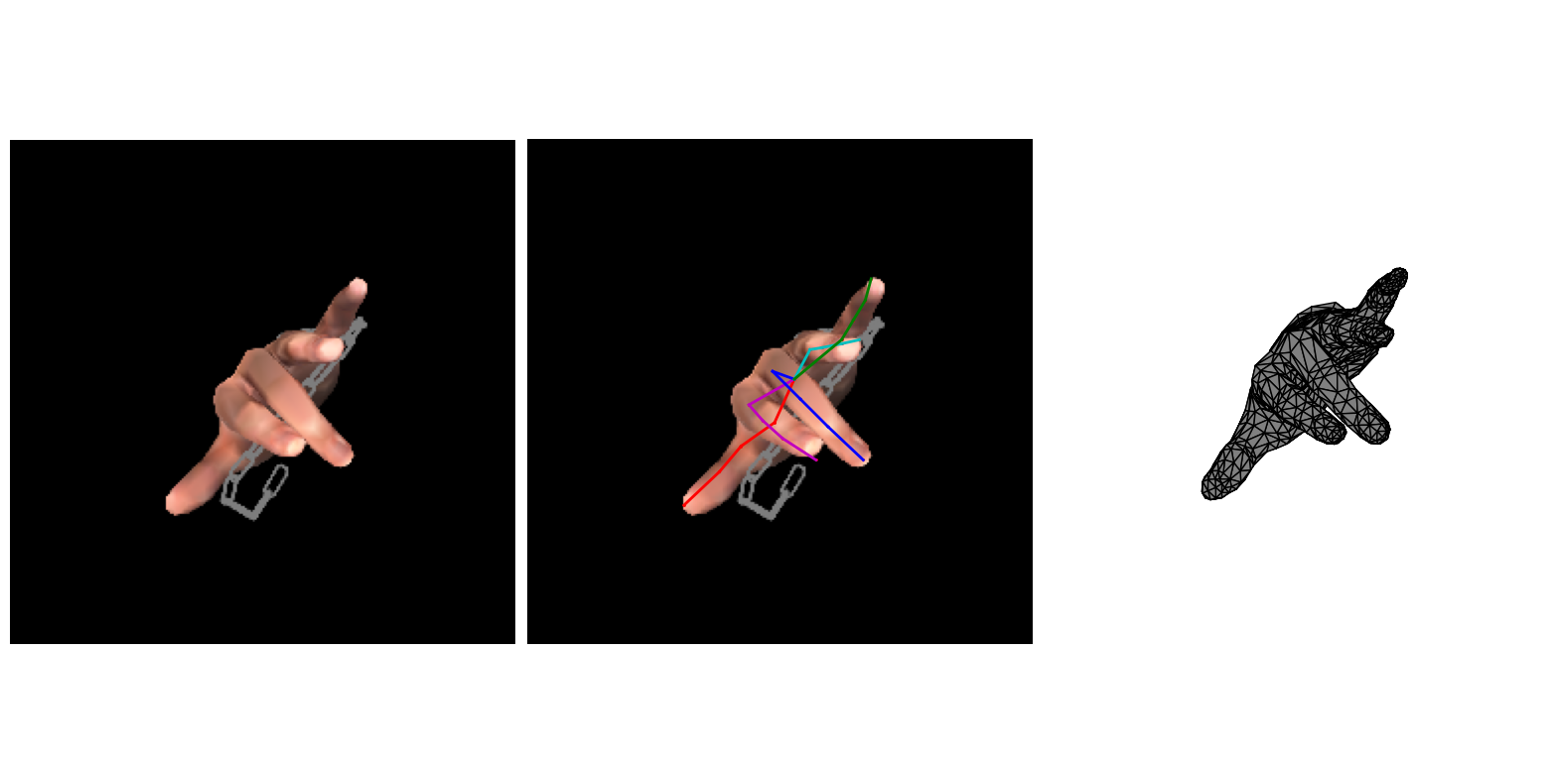} \\ 
		\end{tabular}
		%}
	\end{center}
	\vspace{-0.2cm} 
	\caption{
		Simliar to Fig.~\ref{fig:inter_model_results_hand_hand} only focusing on Hand-Chain interaction images.
	}
	\label{fig:inter_model_results_hand_chain}
\end{figure*}

\begin{figure*}
\begin{center}
		\tabcolsep 0.2cm
 		\renewcommand\arraystretch{0.1}
		%\resizebox{\textwidth}{!}{
		\begin{tabular}{cc}			
			%			\multicolumn{2}{c}{A}  & B \\
			\includegraphics[trim = 0mm 30mm 0mm 30mm, clip,height=3.0cm]{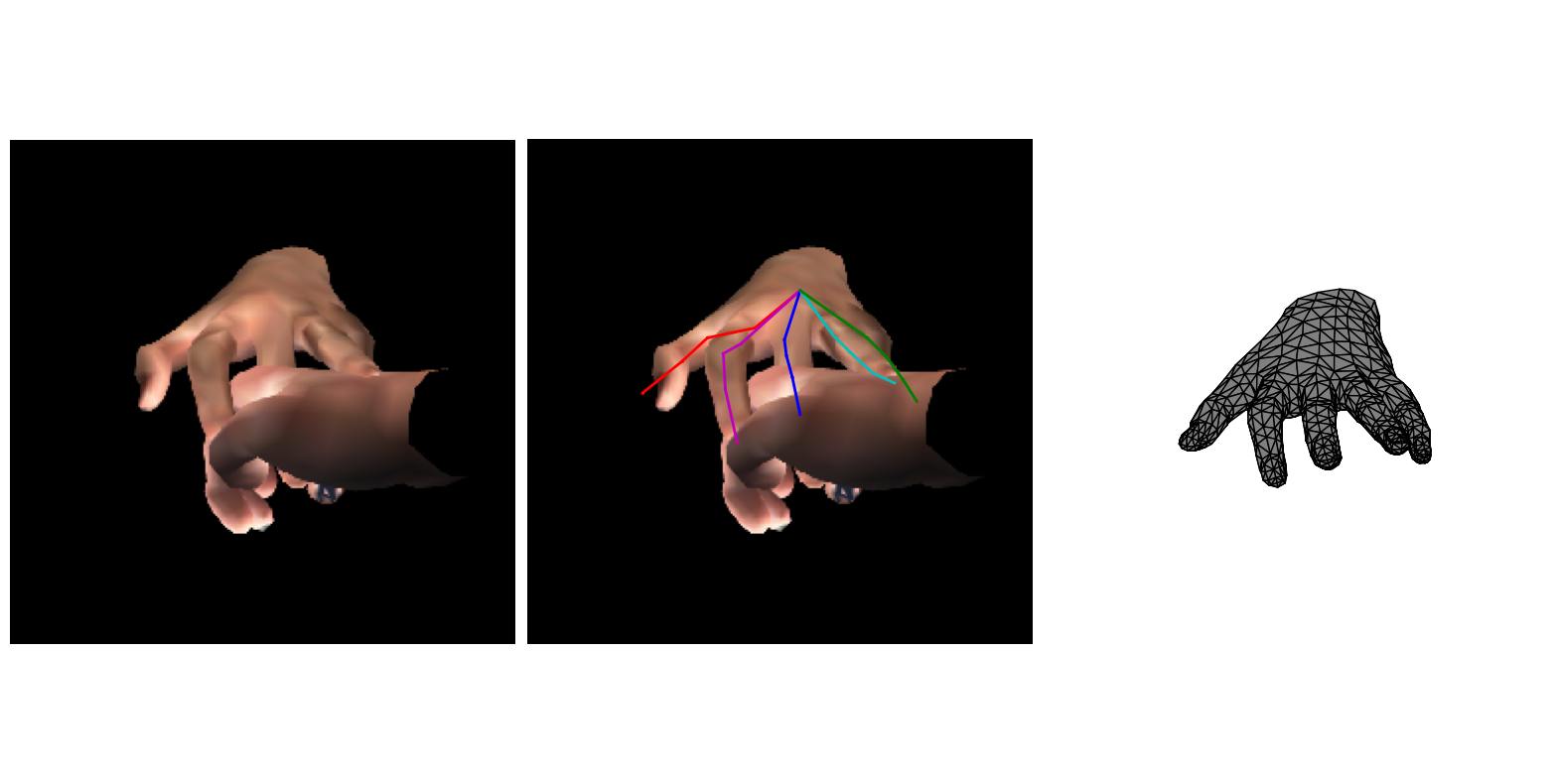} &
			\includegraphics[trim = 0mm 30mm 0mm 30mm, clip, height=3.0cm]{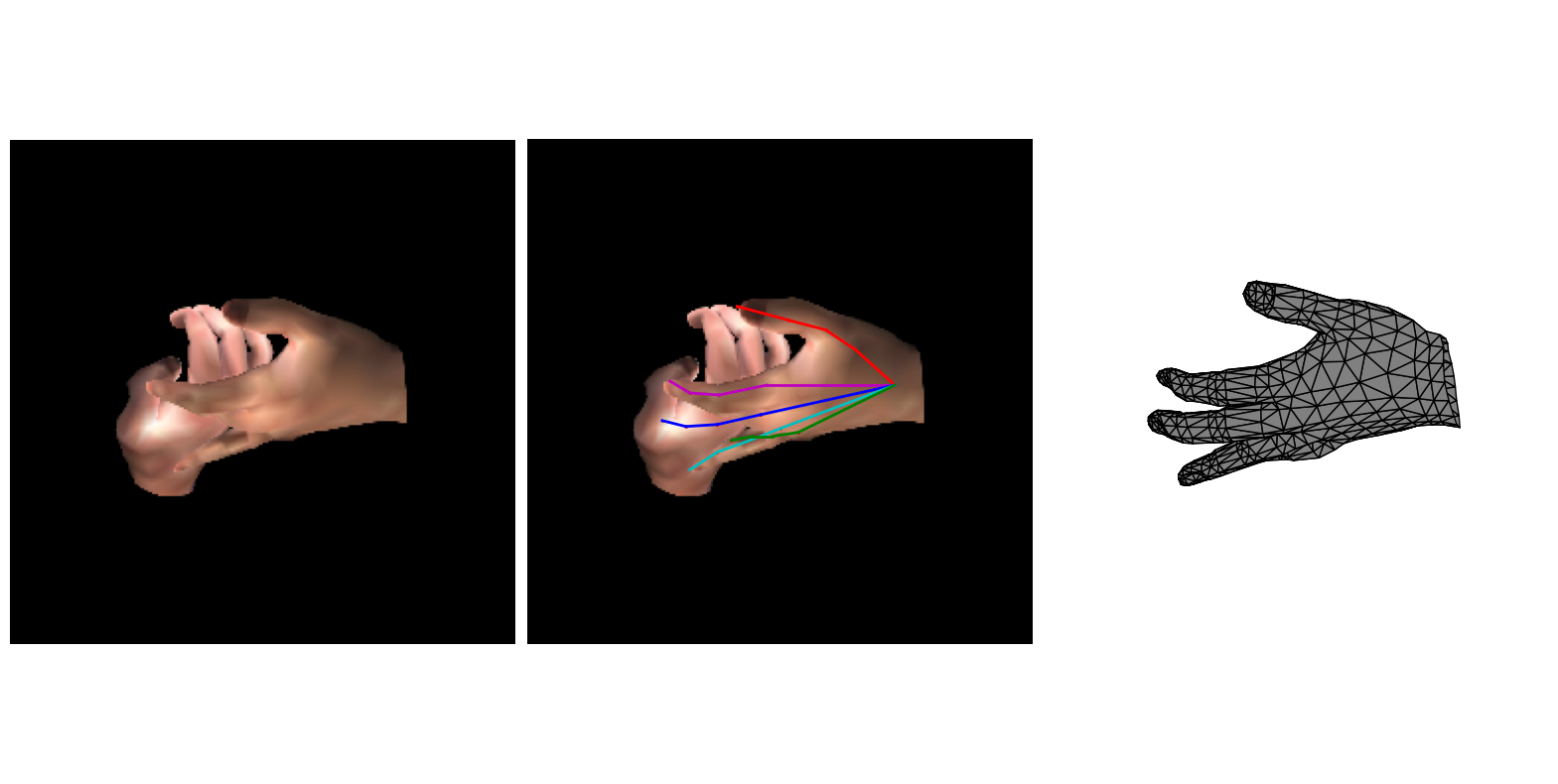} \\
			\includegraphics[trim = 0mm 30mm 0mm 30mm, clip, height=3.0cm]{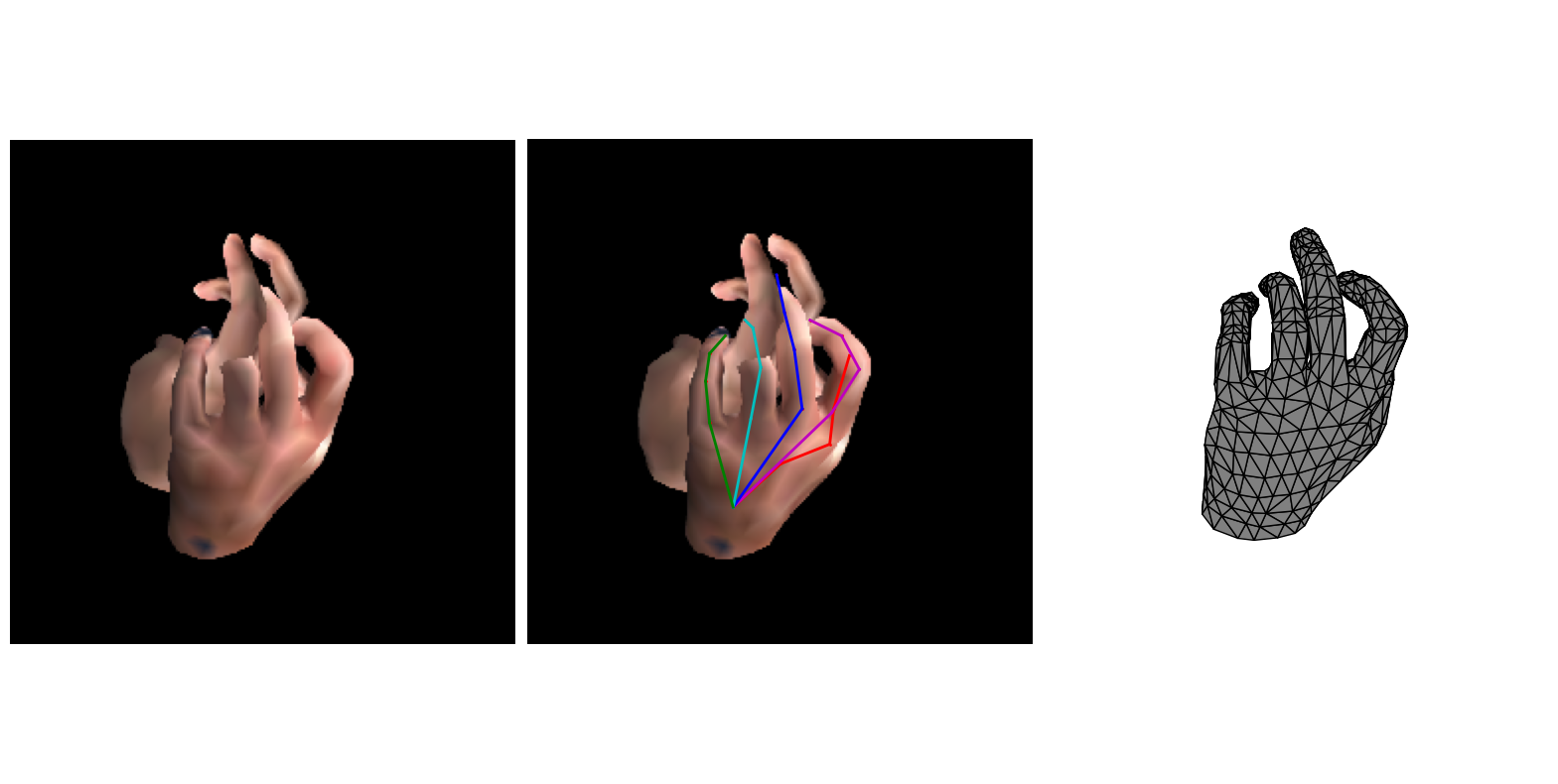} &
			\includegraphics[trim = 0mm 30mm 0mm 30mm, clip, height=3.0cm]{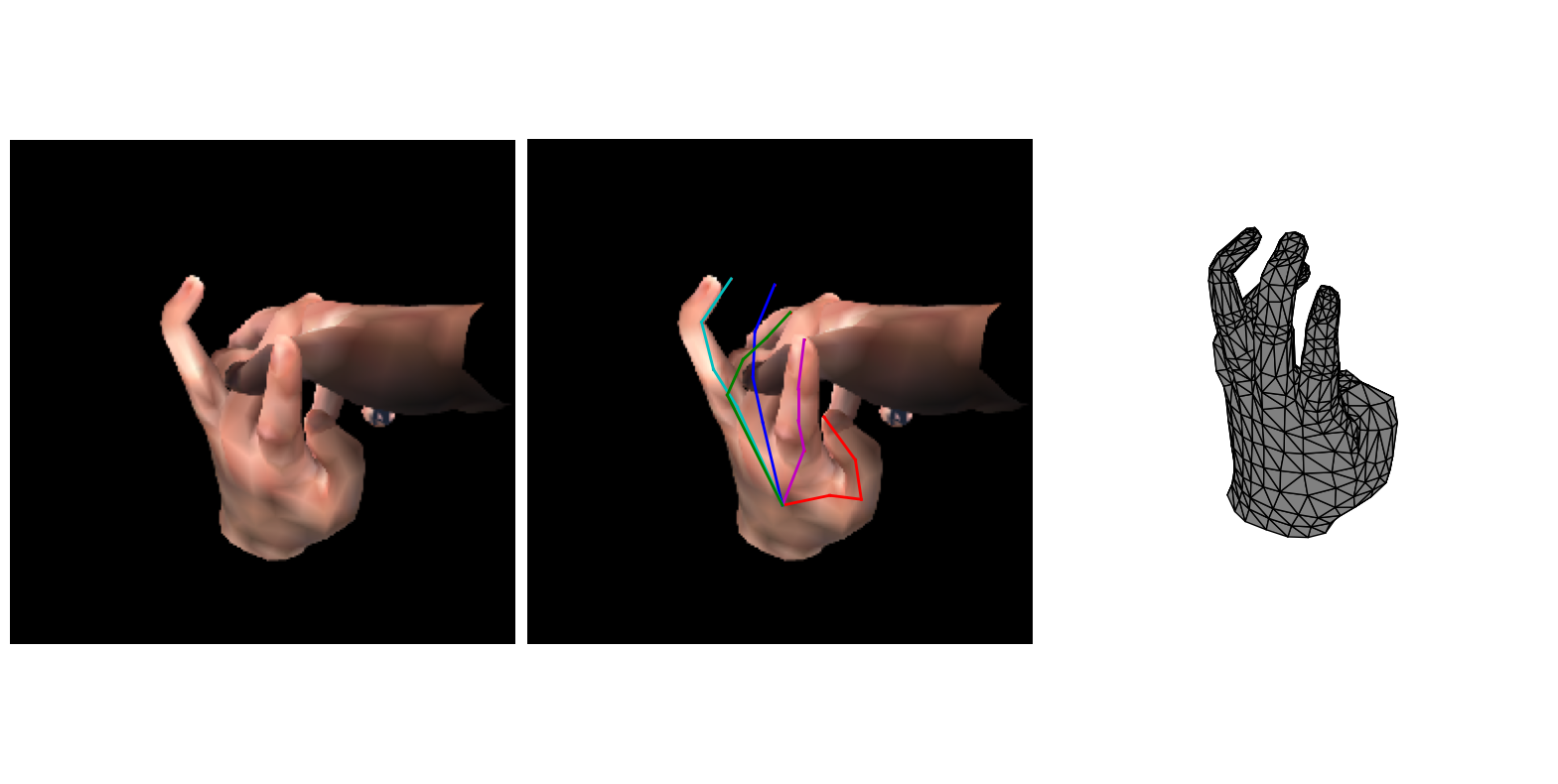} \\ 
			\includegraphics[trim = 0mm 30mm 0mm 30mm, clip,height=3.0cm]{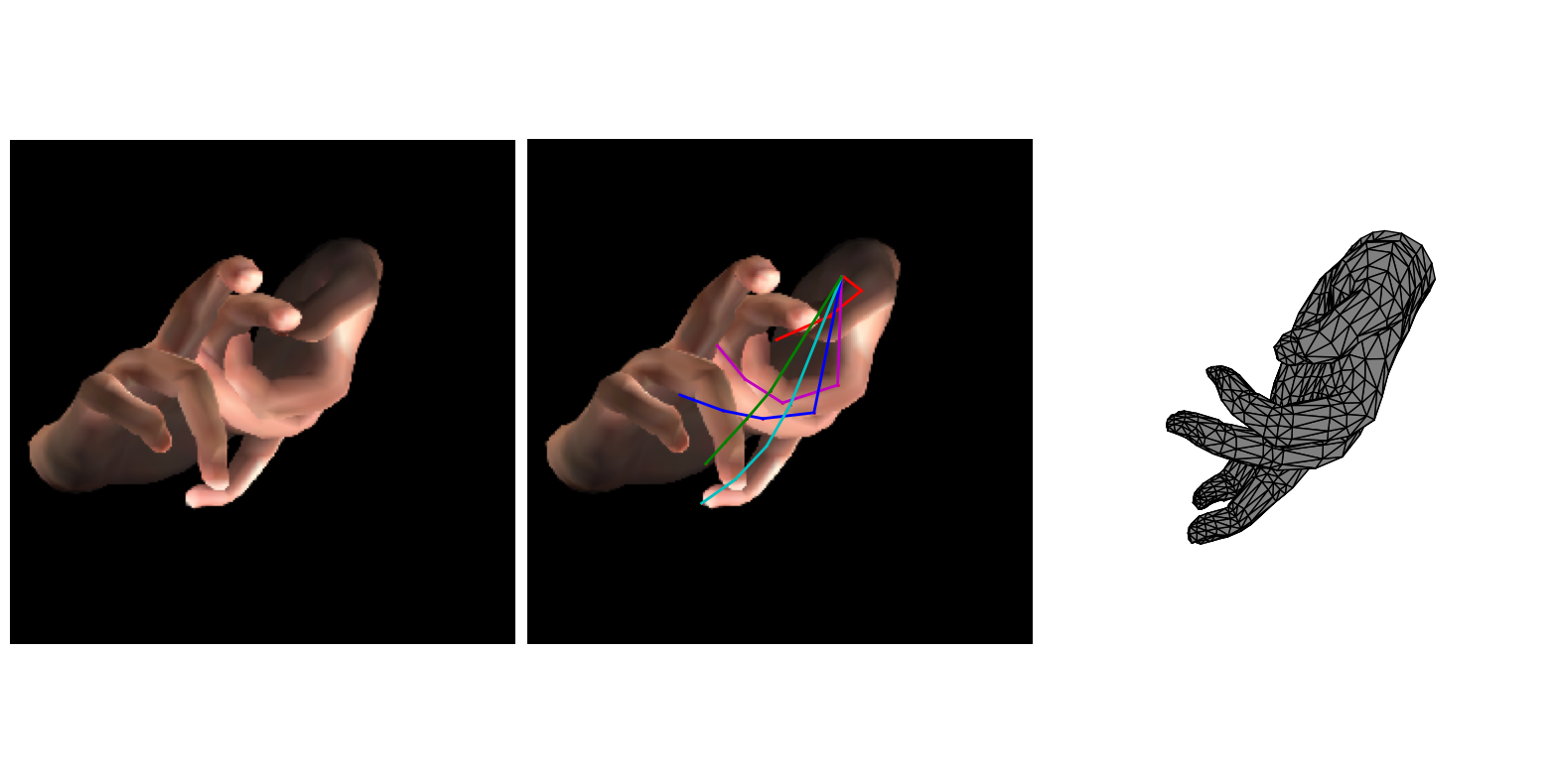} &
			\includegraphics[trim = 0mm 30mm 0mm 30mm, clip, height=3.0cm]{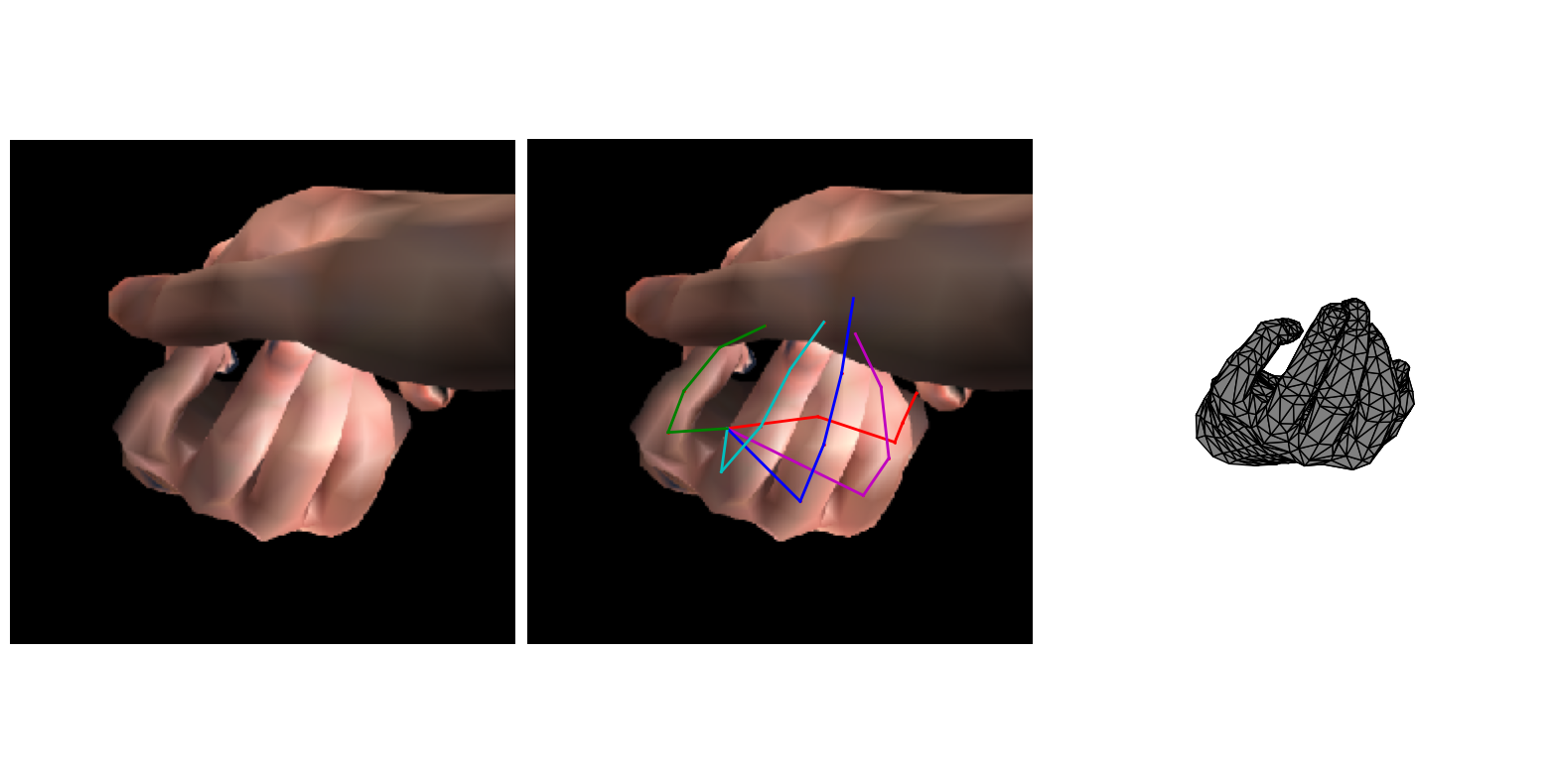} \\
		\end{tabular}
		%}
	\end{center}
	\vspace{-0.2cm} 
	\caption{
		Similar to Fig~\ref{fig:inter_model_results_hand_hand} only showing incorrect predictions for Hand-Hand interaction images.
	}
	\label{fig:inter_model_results_hand_bad}
\end{figure*}

We present qualitative results on Friehand and Inter-SH in Figs~\ref{fig:inter_model_results_hand_hand},~\ref{fig:inter_model_results_hand_chain},~ \ref{fig:friehand_model_results},~\ref{fig:inter_model_results_hand_bad}.

%^^^^^^^^^^^^^^^^^^^^^^^^^^^^^^^^^^^^^^^^^^^^^^^
%^^^^^^^^^^^^^^^^^^^^^

%^^^^^^^^^^^^^^^^^^^^^^^^^^^^^^^^^^^^^^^^^^^^^^^
%^^^^^^^^^^^^^^^^^^^^^

%%%%%%%%%%%%%%%%%%%%%%% RESULTS FIGURE %%%%%%%%%%%%%%%%%%%%%%%%%%%%%%
%^^^^^^^^^^^^^^^^^^^^^^^^^^^^^^^^^^^^^^^^^^^^^^^
\begin{figure*}
\begin{center}
		\tabcolsep 0.2cm
 		\renewcommand\arraystretch{0.1}
		%\resizebox{\textwidth}{!}{
		\begin{tabular}{cccc}			
			%			\multicolumn{2}{c}{A}  & B \\

			\includegraphics[trim = 0mm 0mm 00mm 00mm, clip, height=1.9cm]{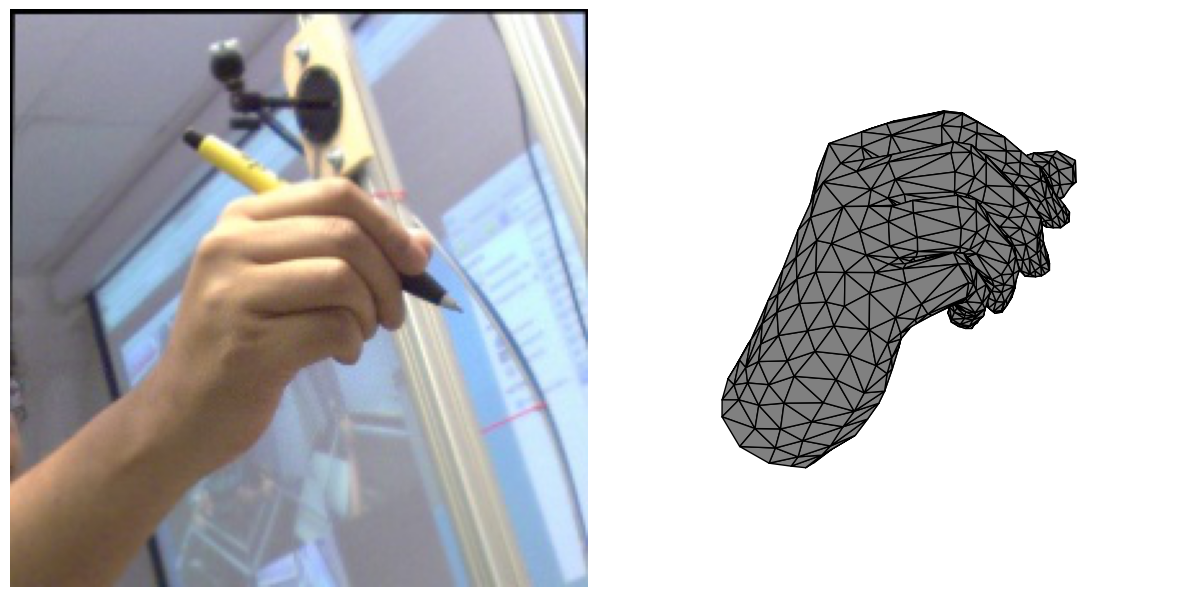} &
			\includegraphics[trim = 0mm 0mm 0mm 00mm, clip, height=1.9cm]{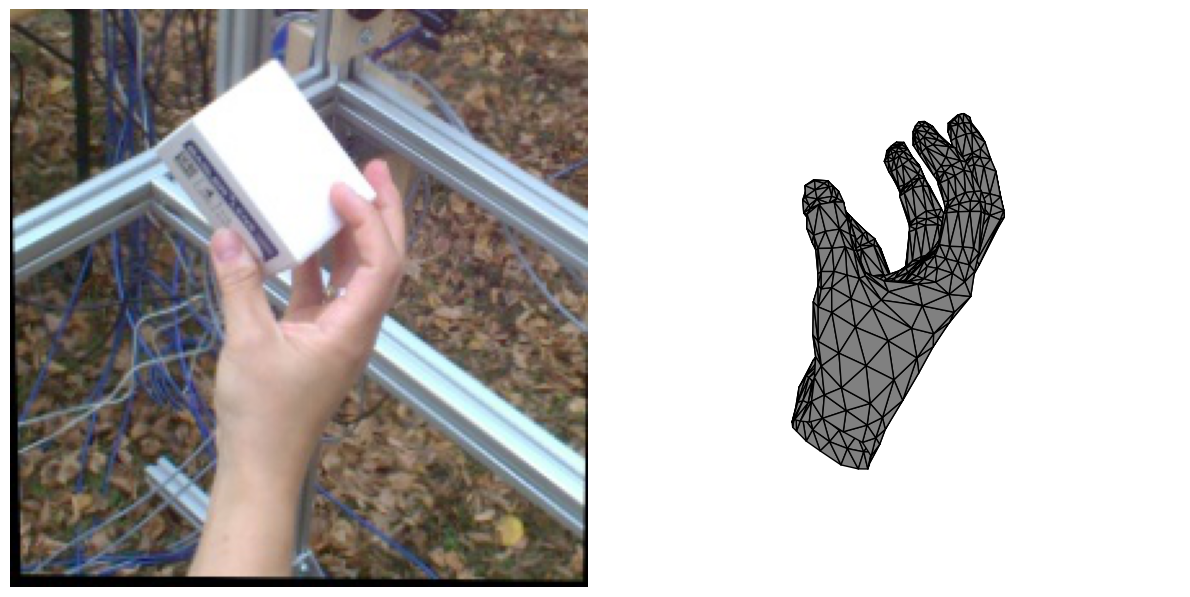} &
			\includegraphics[trim = 0mm 0mm 0mm 00mm, clip, height=1.9cm]{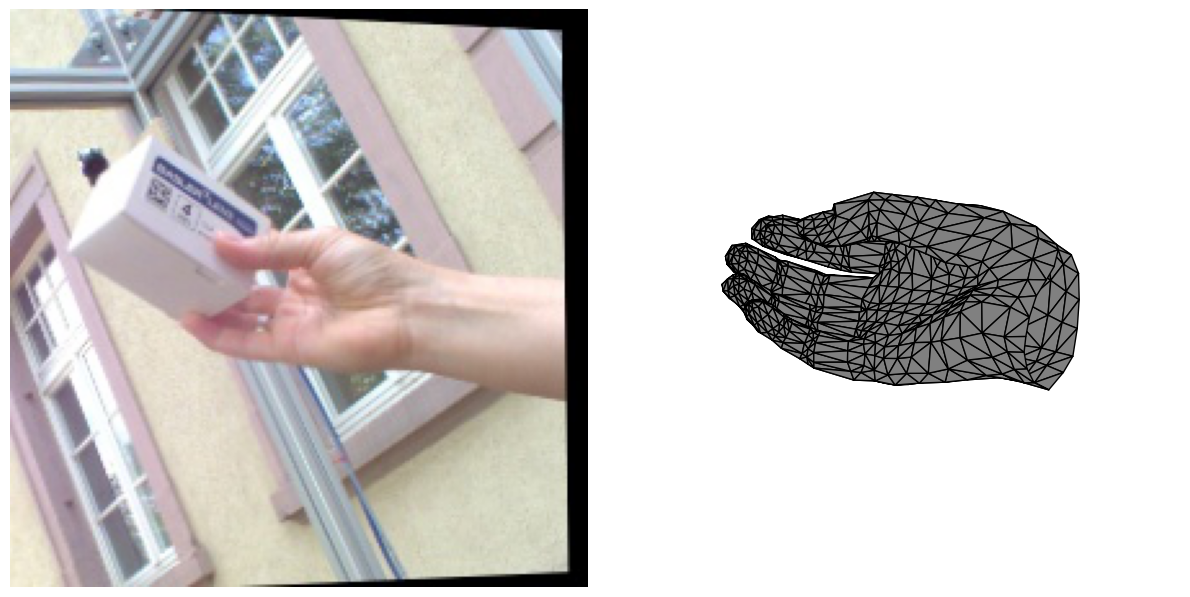} &
			\includegraphics[trim = 0mm 0mm 0mm 00mm, clip, height=1.9cm]{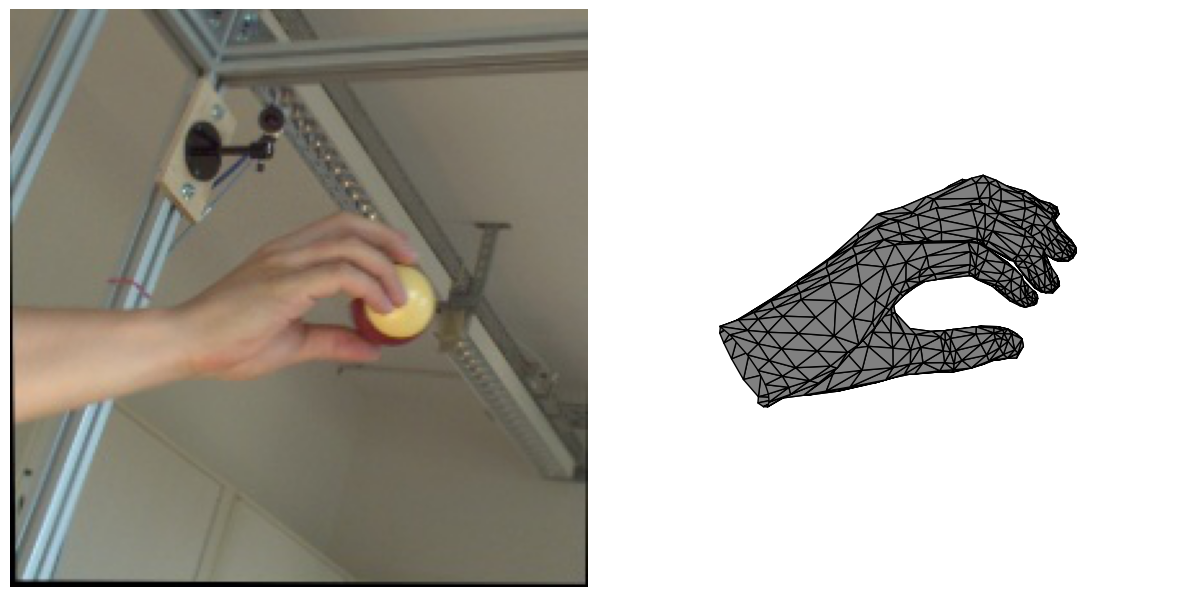} \\ \\
			\includegraphics[trim = 0mm 0mm 00mm 00mm, clip, height=1.9cm]{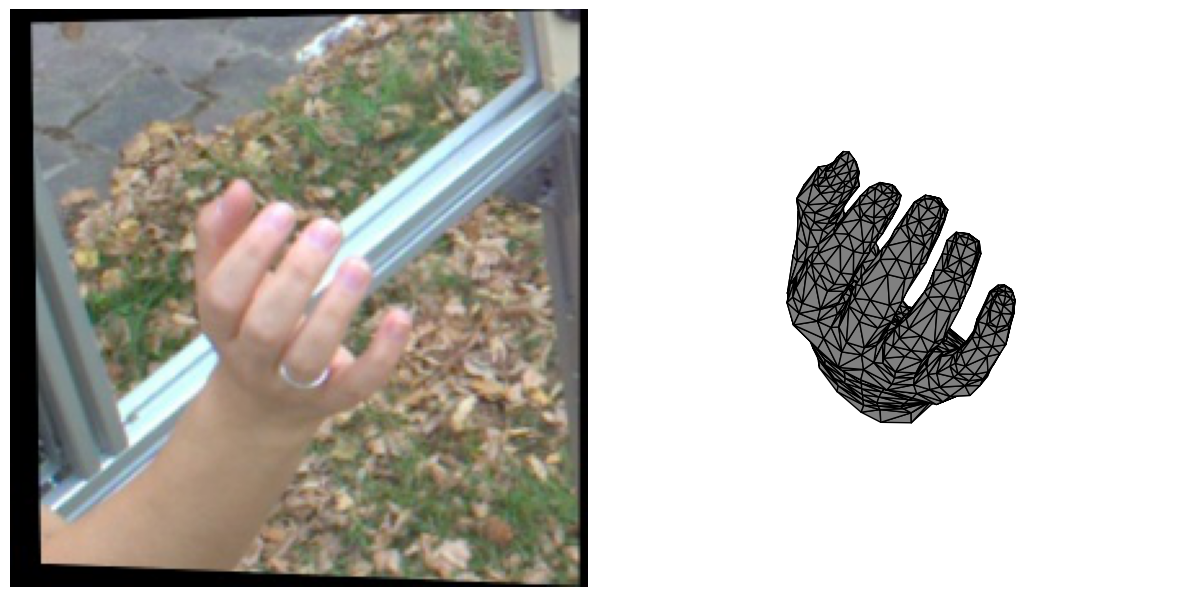} &
			\includegraphics[trim = 0mm 0mm 0mm 00mm, clip, height=1.9cm]{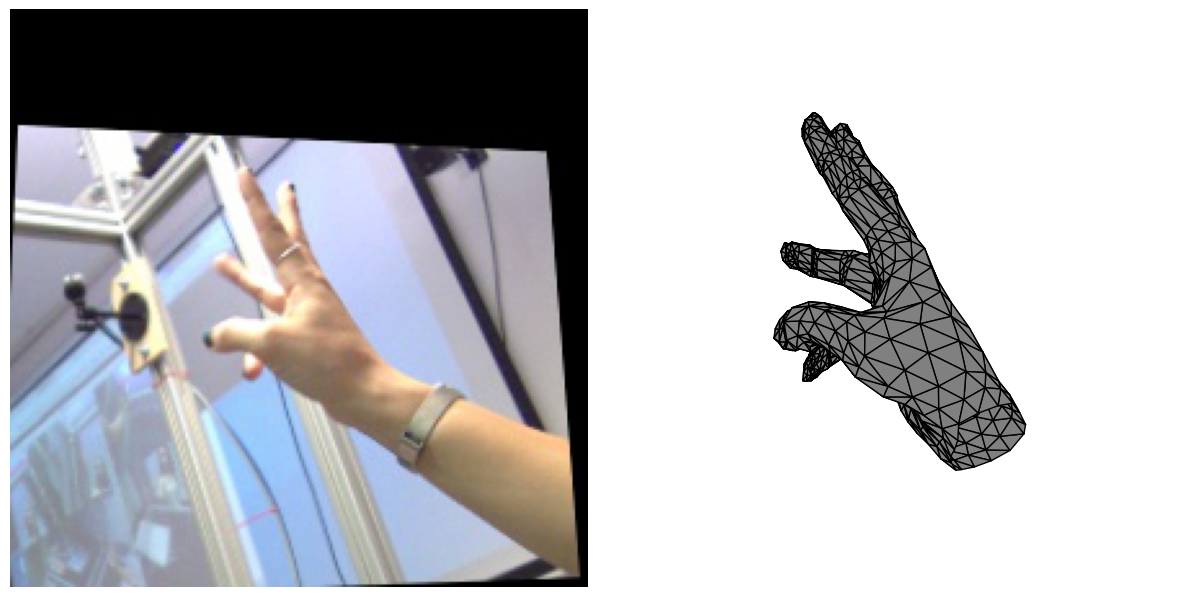} &
			\includegraphics[trim = 0mm 0mm 0mm 00mm, clip, height=1.9cm]{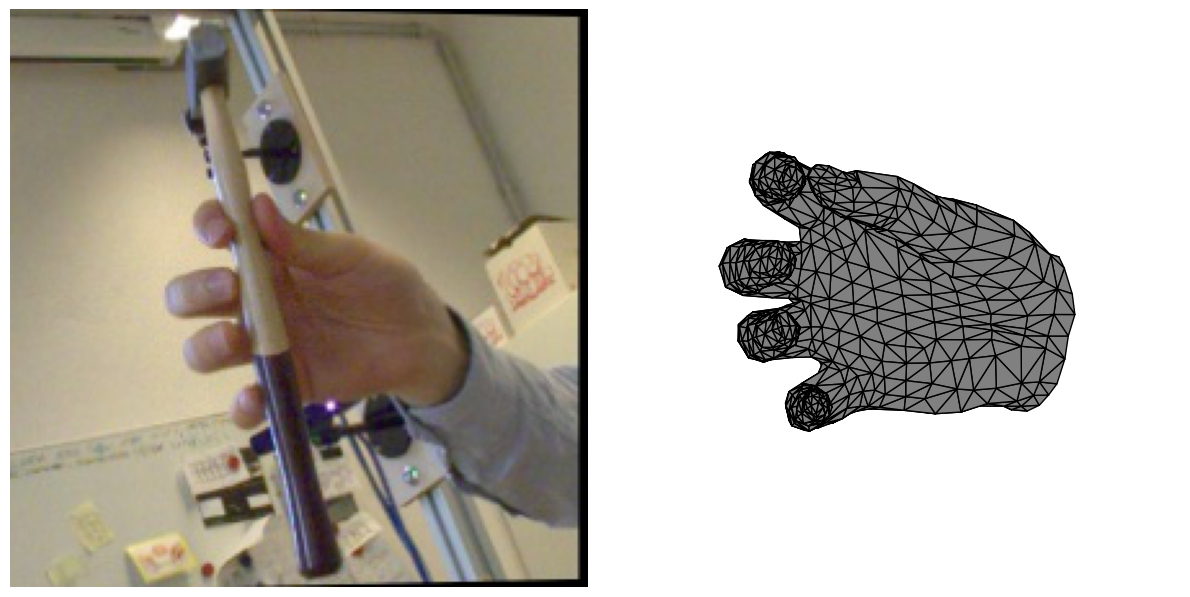} &
			\includegraphics[trim = 0mm 0mm 0mm 00mm, clip, height=1.9cm]{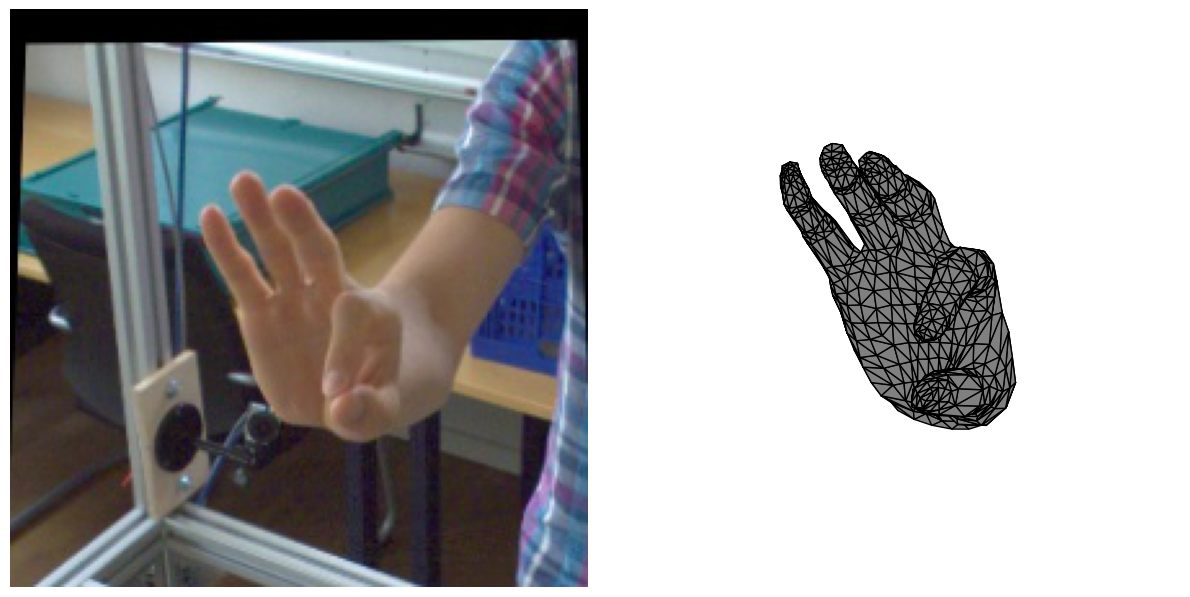} \\ \\
			\includegraphics[trim = 0mm 0mm 00mm 00mm, clip, height=1.9cm]{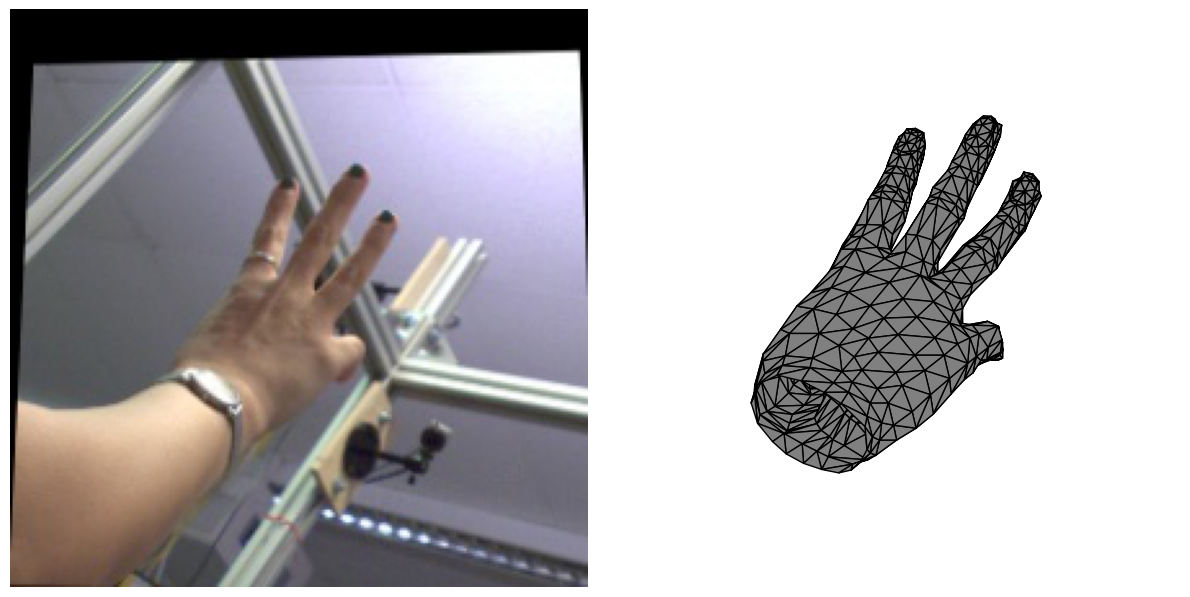} &
			\includegraphics[trim = 0mm 0mm 0mm 00mm, clip, height=1.9cm]{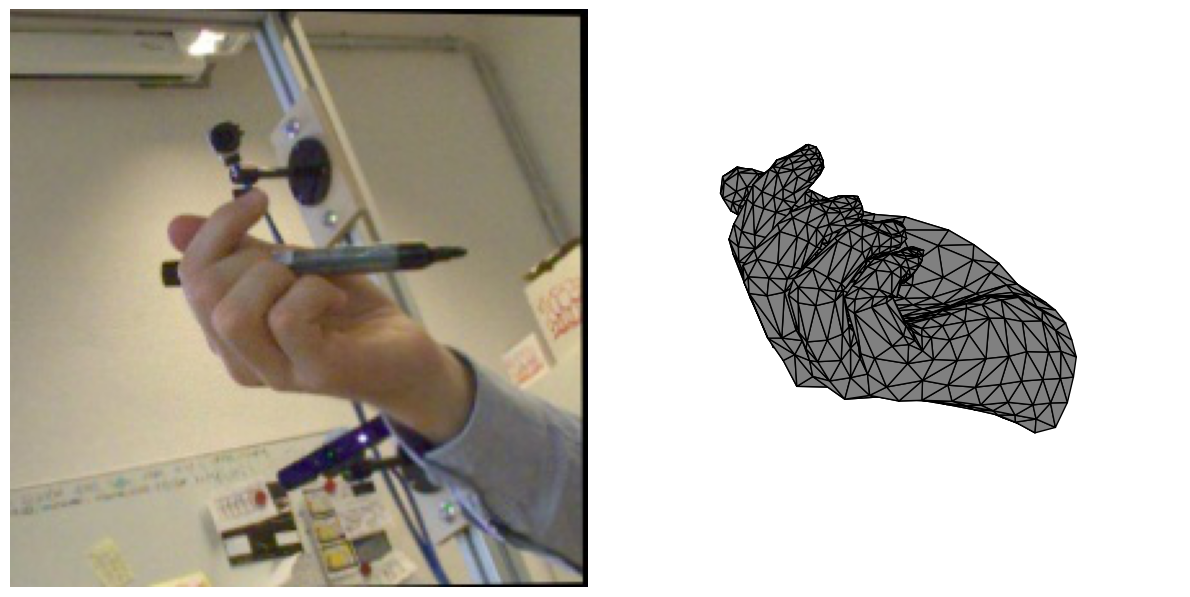} &
			\includegraphics[trim = 0mm 0mm 0mm 00mm, clip, height=1.9cm]{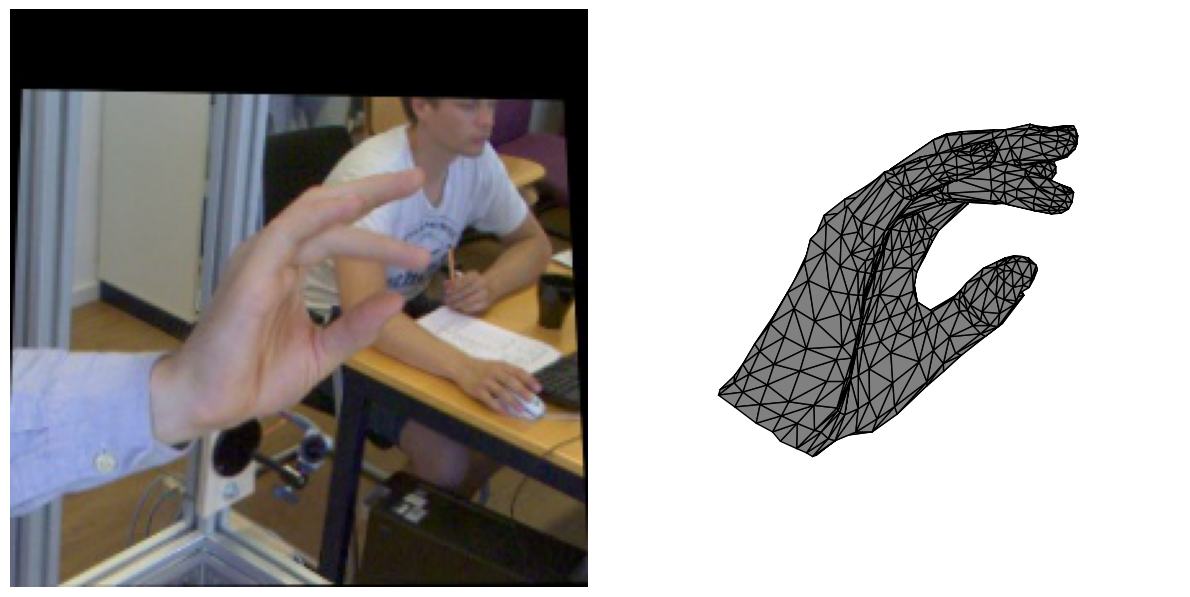} &
			\includegraphics[trim = 0mm 0mm 0mm 00mm, clip, height=1.9cm]{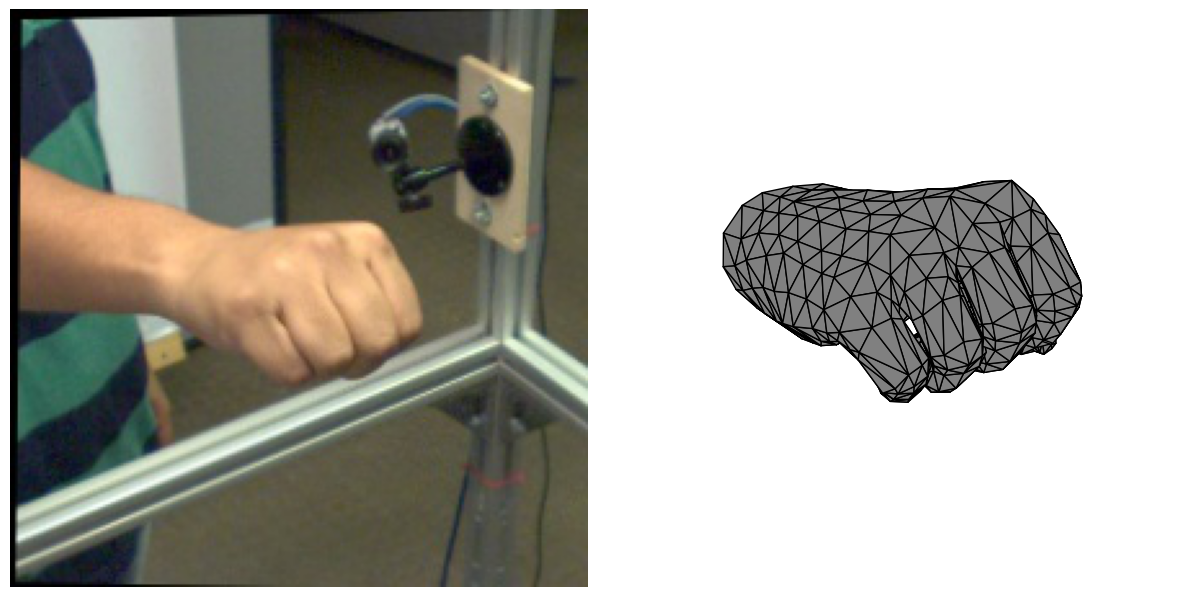} \\ \\
			\includegraphics[trim = 0mm 0mm 00mm 00mm, clip, height=1.9cm]{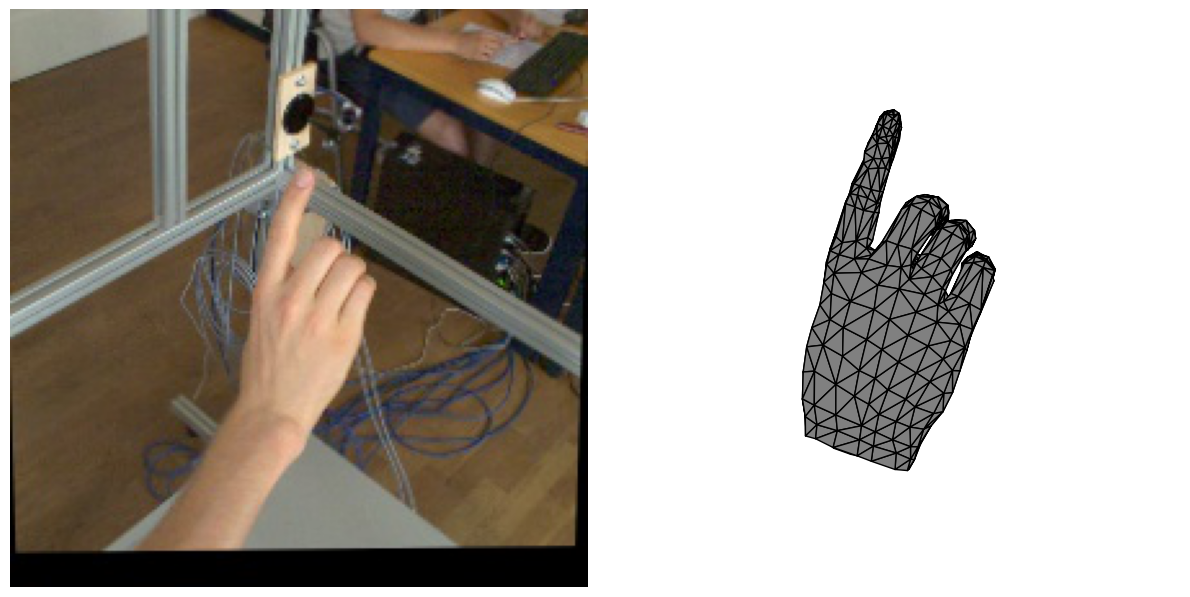} &
			\includegraphics[trim = 0mm 0mm 0mm 00mm, clip, height=1.9cm]{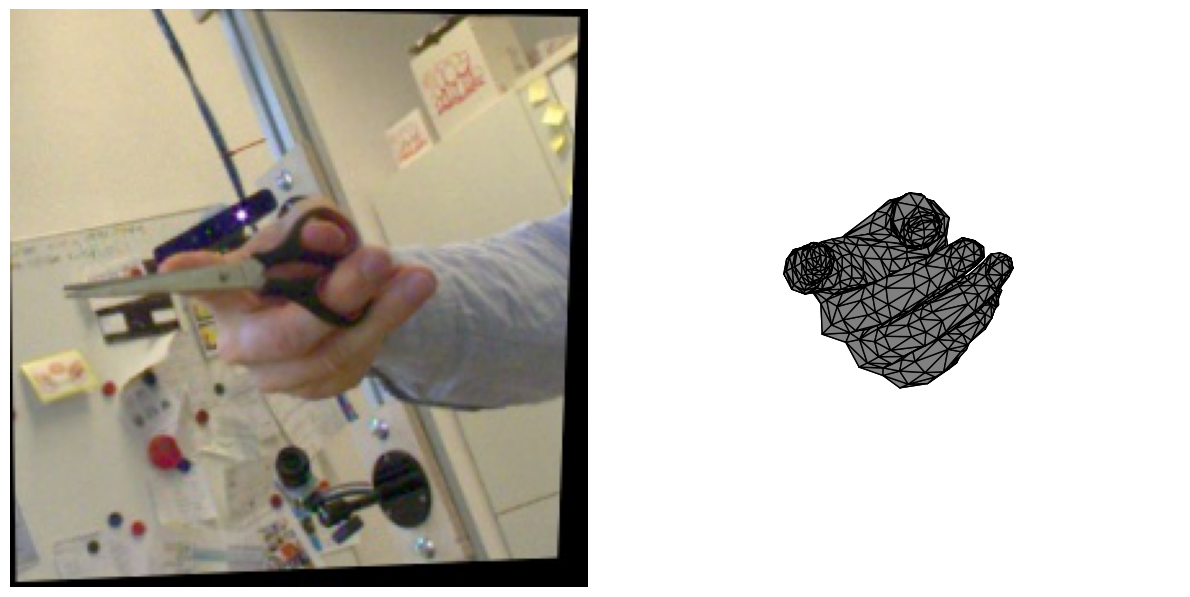} &
			\includegraphics[trim = 0mm 0mm 0mm 00mm, clip, height=1.9cm]{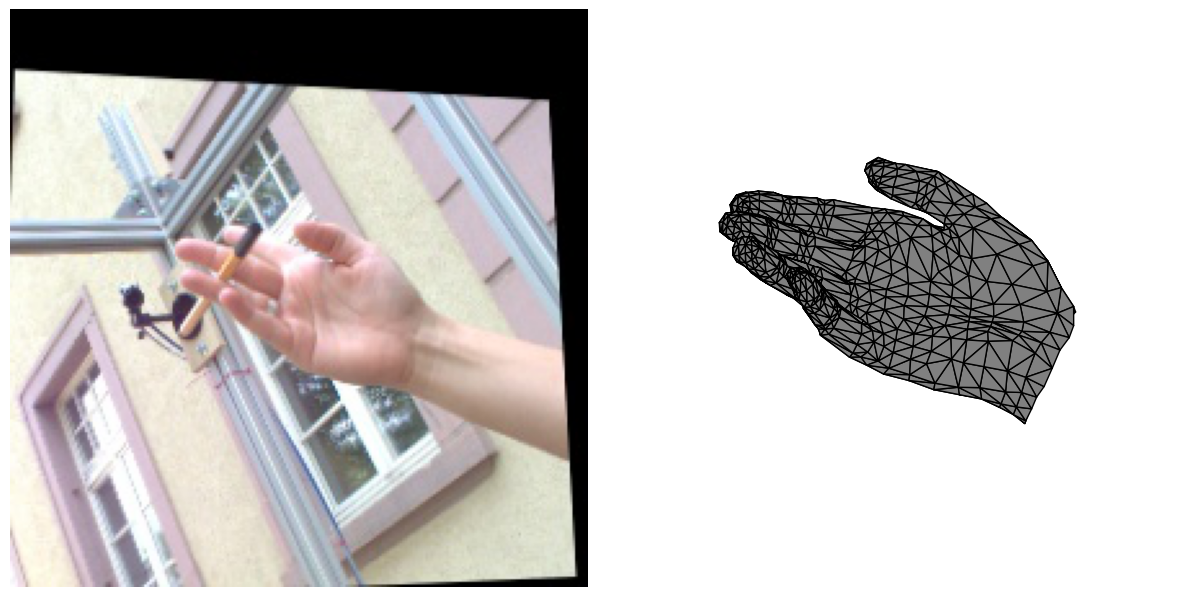} &
			\includegraphics[trim = 0mm 0mm 0mm 00mm, clip, height=1.9cm]{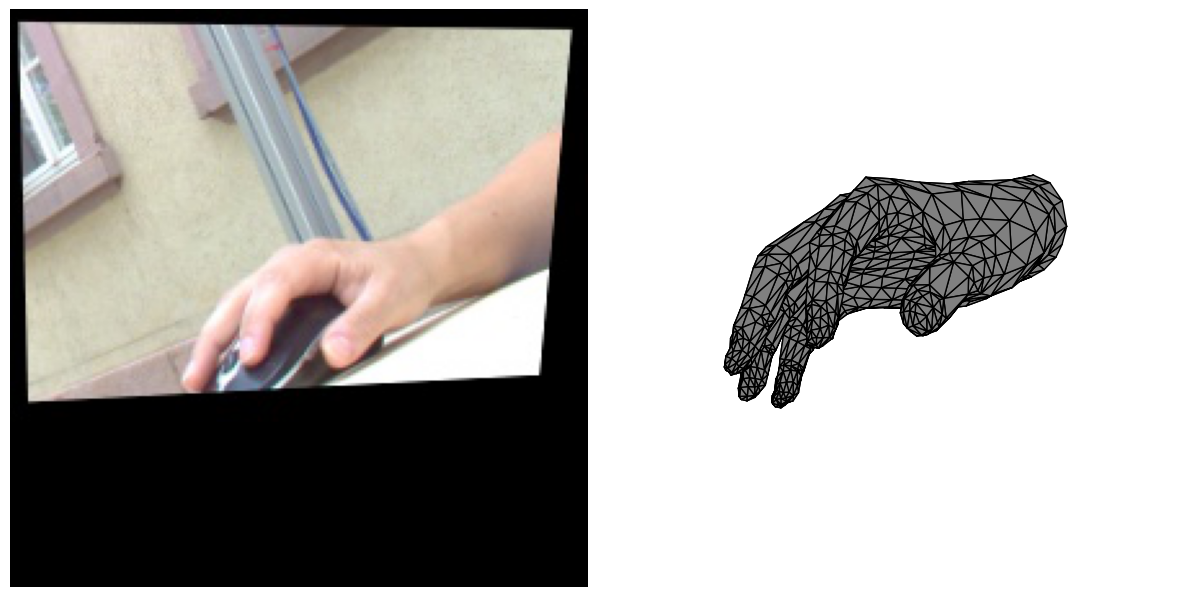} \\ \\
		\end{tabular}
		%}
	\end{center}
	\vspace{-0.2cm} 
	\caption{
		Mesh reconstruction results for our proposed model on the FrieHAND dataset.  
	}
	\label{fig:friehand_model_results}
\end{figure*}
%^^^^^^^^^^^^^^^^^^^^^

%^^^^^^^^^^^^^^^^^^^^^^^^^^^^^^^^^^^^^^^^^^^^^^^
%^^^^^^^^^^^^^^^^^^^^^

\end{document}